\documentclass[runningheads]{llncs}

\usepackage{eccv}

\usepackage{eccvabbrv}

\usepackage{graphicx}
\usepackage{booktabs}

\usepackage[accsupp]{axessibility}  

\usepackage[pagebackref,breaklinks,colorlinks,citecolor=eccvblue]{hyperref}

\usepackage{orcidlink}

\usepackage{enumitem}
\usepackage{balance}

\usepackage{ulem}
\usepackage{todonotes}
\usepackage[T1]{fontenc}
\usepackage{soul}

\usepackage[rightcaption]{sidecap}

\renewcommand{\emph}[1]{\textit{#1}}

\definecolor{cvprblue}{rgb}{0.21,0.49,0.74}
\definecolor{plt:green}{HTML}{2ca02c}
\definecolor{plt:red}{HTML}{d62728}
\definecolor{plt:blue}{HTML}{1f77b4}
\definecolor{plt:orange}{HTML}{ff7f0e}
\usepackage[dvipsnames]{xcolor}
\usepackage{multirow}
\usepackage{makecell}
\usepackage{adjustbox}
\usepackage{tikz}
\usetikzlibrary{tikzmark}
\usepackage{array}
\usepackage{tcolorbox}
\usepackage{booktabs}
\usepackage{wrapfig}
\usepackage{pifont}

\definecolor{mpllightblue}{HTML}{ADD8E6}
\definecolor{mplblue}{HTML}{1F77B4}
\definecolor{mplorange}{HTML}{FF7F0E}
\definecolor{mplgreen}{HTML}{2CA02C}
\definecolor{mplpurple}{HTML}{9467BD}

\begin{document}
	
	\title{Reliability in Semantic Segmentation: \\Can We Use Synthetic Data?} 
	
	\titlerunning{Reliability in Semantic Segmentation: Can We Use Synthetic Data?}
	
	\author{
		Thibaut Loiseau\inst{2}\dag
		\and
		Tuan-Hung Vu\inst{1} \and
		Mickael Chen\inst{1} \and
		Patrick Pérez\inst{3} \and
		Matthieu Cord\inst{1,4} 
	}
	
	\authorrunning{T.Loiseau et al.}
	
	\institute{valeo.ai, Paris, France \and
		LIGM, Ecole des Ponts, Univ Gustave Eiffel, CNRS, Marne-la-Vallee, France \and
		kyutai, Paris, France \and
		Sorbonne Université, Paris, France
	}
	
	\maketitle
	
	\begin{abstract}
Assessing the robustness of perception models to covariate shifts and their ability to detect out-of-distribution (OOD) inputs is crucial for safety-critical applications such as autonomous vehicles. By nature of such applications, however, the relevant data is difficult to collect and annotate. In this paper, we show for the first time how synthetic data can be specifically generated to assess comprehensively the real-world reliability of semantic segmentation models. By fine-tuning Stable Diffusion \cite{rombach2022high} with only in-domain data, we perform \emph{zero-shot generation} of visual scenes in OOD domains or inpainted with OOD objects. This synthetic data is employed to evaluate the robustness of pretrained segmenters, thereby offering insights into their performance when confronted with real edge cases. Through extensive experiments, we demonstrate a high correlation between the performance of models when evaluated on our synthetic OOD data and when evaluated on real OOD inputs, showing the relevance of such virtual testing. Furthermore, we demonstrate how our approach can be utilized to enhance the calibration and OOD detection capabilities of segmenters. \href{https://github.com/valeoai/GenVal}{Code and data} are made public.
\end{abstract}
	\setlength{\skip\footins}{5pt}
\renewcommand{\thefootnote}{}
\footnotetext{\dag~Work done during an internship at valeo.ai} 
\renewcommand{\thefootnote}{\arabic{footnote}}
\section{Introduction}
\label{sec:intro}

Despite the rapid adoption of deep networks in safety-critical applications, reliability~\cite{ovadia2019can,taori2020measuring,tran2022plex} has been an overlooked factor when designing and training these models.
Recent efforts are geared toward enhancing model robustness under covariate shifts in data distributions~\cite{ovadia2019can,recht2019imagenet} and improving the model's ability to detect the \emph{unknown}~\cite{hendrycks2017baseline,yang2022openood,hendrycks2021many}.
For open-world validation, in-domain data is no longer sufficient~\cite{koh2021wilds}; reliable and trustworthy systems demand more rigorous testing on diverse distributions, potentially exhibiting unknown objects.
However, data collection campaigns can be quickly overwhelmed by the growing number of out-of-distribution (OOD) objects and conditions, sometimes extreme.

\begin{figure}[h!]
    \centering
    \includegraphics[trim=1mm 26mm 1mm 14mm, clip, width=0.92\textwidth]{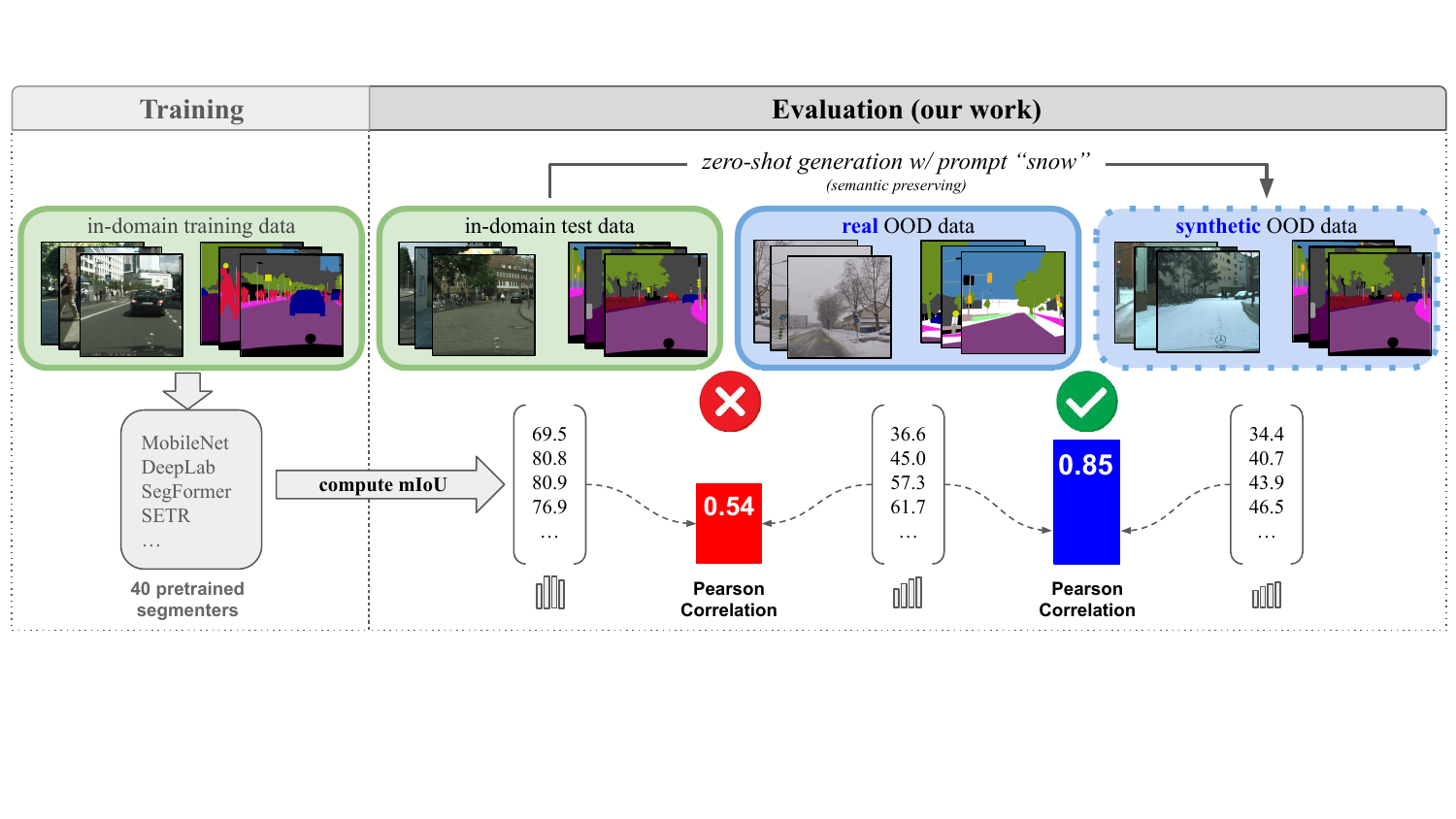}
    \caption{\small  
    \textbf{Assessing 40 pretrained segmenters under covariate shifts.}
    Segmentation models under scrutiny were trained on Cityscapes train set only (in-domain data). They are evaluated on (i) Cityscapes validation set, (ii) real OOD data, and (iii) proposed synthetic data. We observe a strong correlation between results on (ii) and (iii).}
\label{fig:teaser}
\end{figure}

In this paper, we propose to leverage pretrained generative models, \eg Stable Diffusion 1.5 (SD), to alleviate the need for real OOD data. We use SD as a general validator, targeting multiple faces of reliability: \\
- First, we fine-tune SD on in-domain data to enable mask conditioning; in action, zero-shot prompting generates covariate shift images for testing.~\cref{fig:teaser} illustrates the generation of synthetic data in OOD domains and its use for model evaluation. By assessing 40 pretrained segmenters under covariate shifts, our first contribution demonstrates how our generated synthetic OOD benchmark can act as a powerful proxy for the real OOD benchmark ACDC~\cite{sakaridis2021acdc}.\\
- Second, we explore a similar strategy for OOD object detection assessment: we inpaint objects of unknown classes into in-domain data by injecting appropriate prompts during the zero-shot inpainting process. Inpainted images and OOD masks are used to benchmark the 40 models to see how well they can recognize OOD objects. Our results are strongly correlated to the real OOD benchmark~\cite{chan2021segmentmeifyoucan}. Our high-quality synthetic data is featured in the official \href{https://valeoai.github.io/bravo/}{BRAVO} benchmark.\\
- Third, we demonstrate the usefulness of our two synthetic OOD benchmarks for hyperparameter tuning (here model calibration) and training. We calibrate the pretrained models to targeted OOD domains using our synthetic data and validate our strategy by comparing results to using real OOD data. We also train segmenters on inpainted data for OOD detection, obtaining competitive results.

	\section{Related Work}
\label{sec:rw}
\noindent\textbf{Covariate Shifts.} Modern machine learning models, notably deep networks, fall short in preserving their robustness and estimating their prediction confidence in the presence of covariate shifts~\cite{ovadia2019can,recht2019imagenet}. Various benchmarks~\cite{hendrycks2019benchmarking,taori2020measuring,geirhos2021partial,koh2021wilds,singh2023benchmarking} have emerged to address the need for assessing models' reliability under different distributional shifts.
Hendrycks et al.~\cite{hendrycks2019benchmarking} propose a pioneering benchmark featuring data corrupted by various synthetic perturbations such as noise, blur, and brightness. 
Taori et al.~\cite{taori2020measuring} emphasize the importance of realistic shifts in reliability assessment; they highlight the disparity between natural and synthetic shifts, asserting that there is minimal to no robustness transfer from synthetic to natural distribution shifts.
Sign et al.~\cite{singh2023benchmarking} study low-shot robustness to natural distribution shifts, highlighting the robustness properties of advanced architecture and pretraining strategies.
In~\cite{de2023reliability}, de Jorge et al. extend beyond the standard classification setup to address the reliability problem in semantic segmentation, with findings largely aligned with previous works on classification; interestingly, they point to the disconnection between robustness and confidence calibration, urging more attention to calibration during segmenter design and training. 
On a related line, prior works study the connections between in-domain and out-of-domain robustness~\cite{miller2021accuracy,teney2022id,de2023reliability}; they suggest either positive, none, or even negative correlations between ID and OOD robustness, which indeed largely depends on the type of shifts.
In line with~\cite{de2023reliability}, we focus on segmentation as the visual perception task of interest.
Different from previous works, we advocate for the use of advanced generative models to generate realistic synthetic data for testing segmenters under arbitrary covariate shifts.
Taori et al. \cite{taori2020measuring} criticize the synthetic robustness benchmarks and advocate for using real-shift data; while we agree on the importance of realism in testing, we demonstrate that the rapid advancement of generative models now permits very meaningful virtual assessment.
Our goal is to study whether synthetic data can be a superior choice compared to ID data in correlation studies against OOD robustness -- referred to as real-shift robustness here to avoid confusion with OOD detection.
Inspired by~\cite{de2023reliability}, we also calibrate models but using synthetic OOD data instead.

\smallskip\noindent\textbf{OOD Object Detection.}\enspace
In addition to robustness and calibration, the ability to detect the ``unknown'' is  equally important to assess for trustworthy systems~\cite{hendrycks2017baseline,yang2022openood,hendrycks2021many}.
In semantic segmentation, several datasets are available for evaluating  OOD detection on the road, namely LostAndFound~\cite{pinggera2016lost}, StreetHazards~\cite{hendrycks2019scaling} (synthetic), BDD-Anomaly~\cite{hendrycks2019scaling}, Fishyscapes~\cite{blum2021fishyscapes} (synthetic), and SegmentMeIfYouCan~\cite{chan2021segmentmeifyoucan}.
Encountering and capturing images of OOD objects in real-world scenarios, without deliberately placing them on the road, is quite uncommon.
Effectively, existing datasets are limited in scale, both in terms of the number of images and the variety of object classes.
In this work, we propose leveraging advanced zero-shot inpainting techniques to augment an existing segmentation dataset with the insertion of OOD objects; this enables the generation of highly realistic synthetic data for testing and training OOD detection.

\smallskip\noindent\textbf{Synthetic Data for Testing.}\enspace
Generative models have been exploited to create training data for image classification~\cite{besnier2020dataset,he2022synthetic,sariyildiz2023fake}, object detection~\cite{marathe2023wedge}, or semantic segmentation~\cite{le2021semantic,zhang2021datasetgan,li2022bigdatasetgan, wu2023diffumask,hariat2024learning}.
Only recently, a few works have delved into the topic of generative testing data.
Li et al.~\cite{li2023imagenet} exploit diffusion models to realistically edit images, controlling over various object attributes.
This approach enables stress-testing models and understanding their sensitivity to different attributes.
Using SD, LANCE~\cite{prabhu2023lance} generates counterfactual images capable of challenging any given perception model.
To the extent of our knowledge, no existing works have proposed to generate testing data for segmentation reliability.
	\section{Reliability Under Covariate Shifts}
\label{sec:domain}
In this section, we explore whether synthetic data can be used to assess the robustness of pretrained segmenters in the presence of covariate shifts to unseen OOD domains.
We describe the data generation process and present the benchmarking results for a wide range of segmenters on our synthetic data.
We demonstrate the validity of the approach using domains for which a real OOD dataset exists so that we can have access to a gold standard. 
We stress that we do not use the OOD data at any point in the method itself.
This aspect is critical for the method to be applicable to benchmark robustness in the presence of extreme or hazardous conditions.

\subsection{Generating images in arbitrary domains}
\label{sec:gen_img_shift}
\begin{figure}[h]
    \centering
    \resizebox{.48\textwidth}{!}{
    \includegraphics{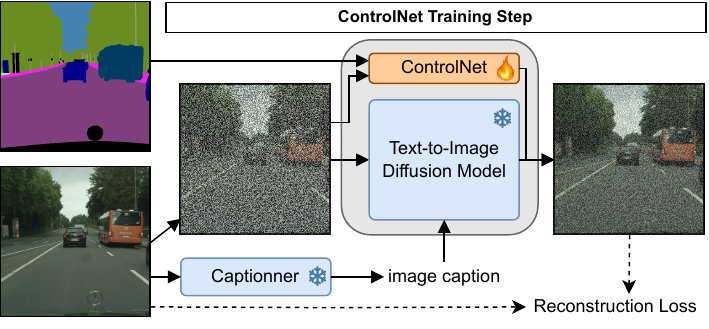}
    }
    \hfill
    \resizebox{.48\textwidth}{!}{\includegraphics{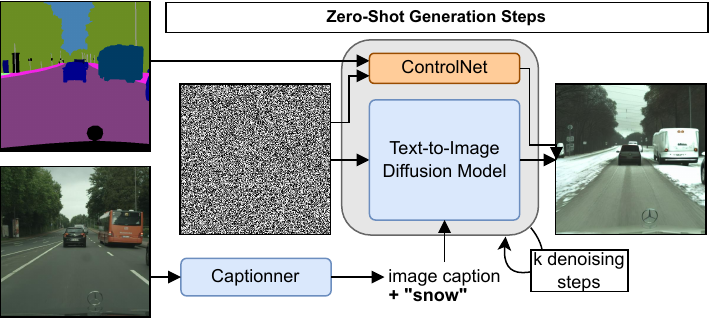}
    }
    \caption{\small \textbf{Generating data with covariate shifts.} Training ({left}) and Sampling ({right}) processes for producing the synthetic data with shifts. {For training, only {in-domain} images and masks are used}. For inference, we use the {in-domain} masks to generate {OOD} images. No real {OOD} data is required in the framework.
    }
    \label{fig:genmodel}
\end{figure}

Our goal here is to obtain pairs of images and semantic masks, with the images belonging to visual domains for which we lack data.
To this end, we leverage a pretrained text-to-image Stable Diffusion 1.5 (SD) model~\cite{rombach2022high}, repurposed as a semantic-conditioned model called ControlNet~\cite{zhang2023adding}.
ControlNet is trained solely on images and segmentation ground truths from the Cityscapes dataset; at train time, the text prompts are captions automatically extracted using CLIP-interrogator~\cite{clipinterrogator}.
As a result, the model is able to perform mask-to-image generation of driving scenes while retaining the ability to steer the generation through text prompting of Stable Diffusion.

To generate synthetic data, we prompt a trained ControlNet by the concatenation of {OOD} domain descriptions and CLIP-interrogator captions obtained from Cityscapes validation images.
Also, segmentation masks in Cityscapes validation set are used to condition the generative process.
Thanks to zero-shot prompting, the synthetic images are aligned with the semantic condition while displaying the visual properties of arbitrary {OOD} domains.
~\cref{fig:genmodel} illustrates the training and generation steps.
Detailed technical descriptions and more visualizations are in~\cref{sec:tech_details} and~\cref{sec:supp_qual}.

\subsection{Robustness Assessment with Synthetic Data}
\label{sec:robustness_assessment}
With the pipeline outlined in~\cref{sec:gen_img_shift}, one can generate synthetic data to assess the robustness of pretrained segmenters in any unseen {OOD} domains through zero-shot prompting.
To quantify robustness, we employ the traditional mean Intersection-over-Union (mIoU) score, measuring the correct overlap between semantic predictions and ground-truth masks.
Given that our synthetic dataset comprises pairs of segmentation masks and synthetic images, one can straightforwardly derive synthetic scores for any pretrained segmenters.
We here investigate whether synthetic performance can {faithfully} 
reflect the performance on real data in {OOD} domains under covariate shifts.

\begin{figure*}[t!]
    \centering
    \includegraphics[trim=0 1.2cm 0 0, clip,width=\textwidth]{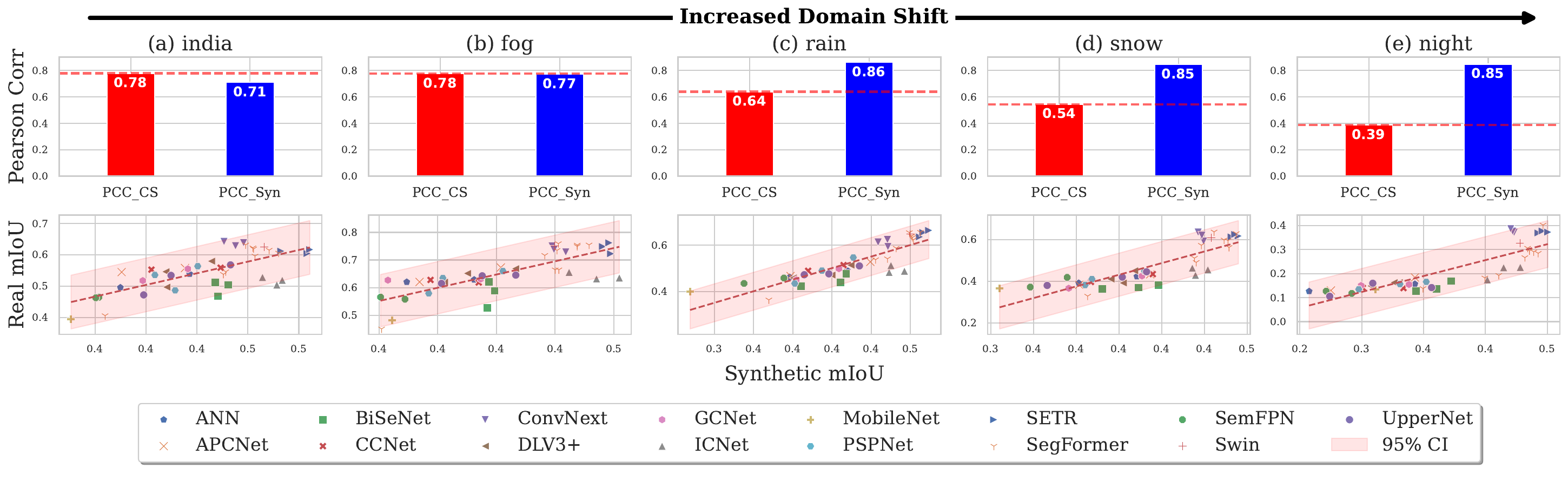}
    \caption{\small \textbf{Robustness correlation between real and synthetic covariate shifts across $40$ pretrained segmenters.} The tested models, see families in bottom legend, cover different architectures and sizes.
    (\textit{top}) Pearson Correlation Coefficients of mIoUs between Cityscapes and real-shifts (`PCC\_CS'~\textcolor{red}{\rule{0.5em}{2ex}}), and between synthetic shifts and real ones (`PCC\_Syn'~\textcolor{blue}{\rule{0.5em}{2ex}}). 
    (\textit{bottom}) Scatter plots of synthetic \vs real mIoUs along with the linear regression line accompanied by $95\%$ confidence intervals (`CI').
    (\textit{a-e}) Five types of domain shifts from Cityscapes in-domain distribution, sorted by increasing gap as assessed by decreasing PCC\_CS.
    The robustness results on synthetic data exhibit a strong correlation with those on real data, particularly in the case of the most distant shifts like `snow' and `night'. More details are provided in~\cref{sec:class_wise}.} 
    \label{fig:mIoU_covariate_shift}
\end{figure*}

\smallskip\noindent\textbf{Experimental Setups.} 
We address weather shifts and geographical shifts, which are often encountered in autonomous driving.
These covariate shifts are exhibited in two existing real datasets: the Adverse Conditions Dataset (ACDC)~\cite{sakaridis2021acdc} and the Indian Driving Dataset (IDD)~\cite{varma2019idd}.
We utilize those real data to quantify the quality of our synthetic data and to validate its usefulness.

Synthetic data are generated by conditioning on semantic masks from the Cityscapes validation set.
To prompt ControlNet, we concatenate CLIP-interrogator's caption with a domain description following a simple template \texttt{[$<$caption$>$, in $<$domain$>$]} where \texttt{domain} is either `india', `fog', `rain', `snow' or `night'.

For testing, we gather a collection of 40 publicly available segmenters \emph{trained only on Cityscapes}, representative of different backbones, segmentation architectures, and sizes.
The full list of models is in~\cref{sec:tech_details}. 

\smallskip\noindent\textbf{Results.}
In~\cref{fig:mIoU_covariate_shift}, we present our main results.
Our primary metric is the Pearson Correlation Coefficient (PCC) between the mIoUs on testing data and on real-shift data from ACDC's splits or from IDD.
The testing data can be either the Cityscapes validation set (CS) or our synthetic data (syn); the idea is to see which testing data --whether real CS or our synthetic one-- correlates more with the real-shift data.
{Note that in the absence of OOD data, the Cityscapes validation set acts as the closest easily available proxy and is a reasonable predictor for OOD performance as pointed out by Jorge et al.~\cite{de2023reliability}.}

We organize our results based on increasing domain gaps relative to the Cityscapes domain.
The domain gaps are quantified by the Pearson correlation between Cityscapes mIoUs and real-shift mIoUs, annotated as PCC\_CS and visualized as red bars \textcolor{red}{\rule{0.5em}{2ex}} in the subplots of~\cref{fig:mIoU_covariate_shift}.
Moving from left to right, \ie with growing domain gaps, we observe a widening discrepancy between PCC\_CS and PCC\_Syn.
Here, PCC\_Syn (\textcolor{blue}{\rule{0.5em}{2ex}}) represents the Pearson correlation between synthetic mIoUs and real-shift mIoUs.
In domains with small gaps, PCC\_CS and PCC\_Syn are relatively comparable. However, in domains with more adverse shifts, such as `snow' and `night', PCC\_Syn outperforms PCC\_CS significantly, exceeding PCC\_CS in `night' by more than double.

\begin{figure}[h!]
    \centering
    \includegraphics[width=0.8\linewidth]{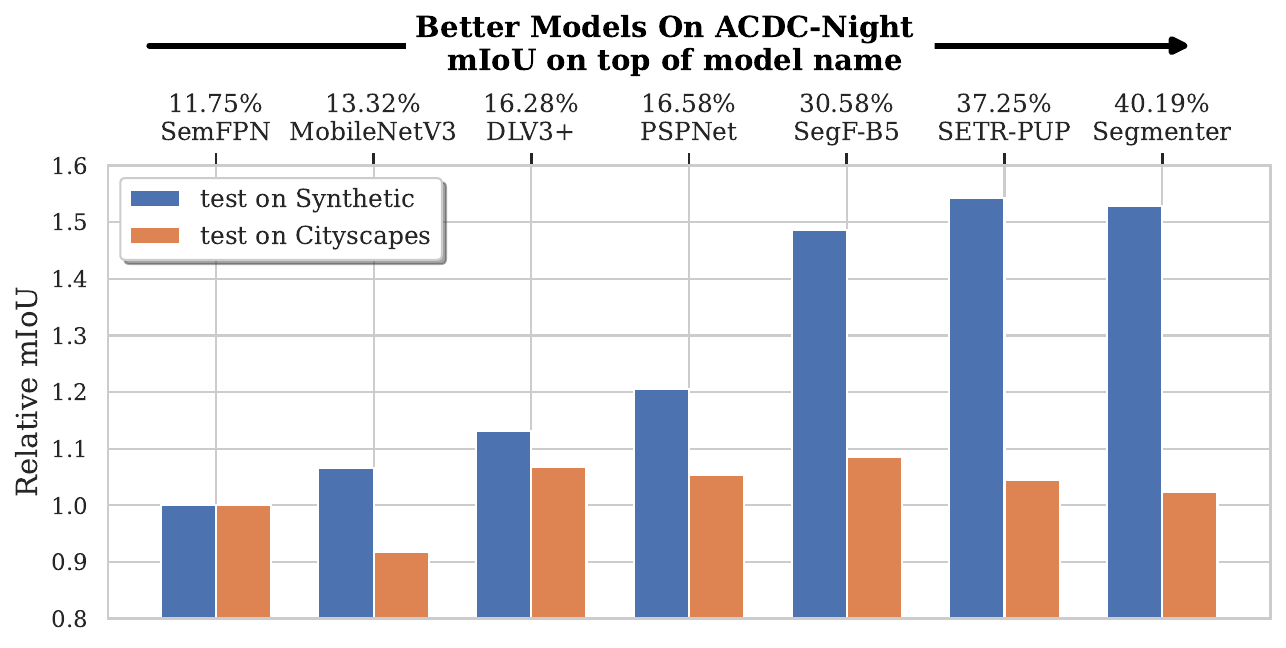}
    \caption{\small \textbf{Day-night shift.}
    Models are ranked from left to right by their robustness on real night data -- ACDC-Night mIoUs are shown on top of model names.
    For each presented architecture, the most robust model on Cityscapes is tested; the Semantic-FPN, DeeplabV3+, and PSPNet models have ResNet-101 as backbone. 
    The Semantic-FPN model (lowest mIoU on ACDC-Night) serves as the reference for computing the relative mIoUs. 
    Blue bars or orange bars show the relative mIoUs when testing on our synthetic data (\textcolor{mplblue}{\rule{0.5em}{2ex}}) or testing on Cityscapes validation data (\textcolor{mplorange}{\rule{0.5em}{2ex}}).
    Cityscapes scores are not reliable for ranking models in the night domain.
    Synthetic scores exhibit a stronger correlation with real night scores, as evidenced by the more consistently increasing trend in the blue bars from left to right.
    }
    \label{fig:mioucorr_day_night}
\end{figure}

In~\cref{fig:mioucorr_day_night}, we analyze the results for the `night' condition using the most robust models across different architectures, ranging from 
ConvNets to recent transformer networks.
We use the Semantic-FPN score as the reference to normalize the scores of other models.
This normalization aims to illustrate the relative improvement in robustness in terms of architecture.
We rank the models from left to right based on their performance on real night data from the ACDC-night split.
The consistently increasing trend of synthetic scores (\textcolor{mplblue}{\rule{0.5em}{.5ex}}\textcolor{mplblue}{\rule{0.5em}{1ex}}\textcolor{mplblue}{\rule{0.5em}{2ex}}) from left to right demonstrates a strong correlation with the ranking based on real scores.
In contrast, Cityscapes scores (\textcolor{mplorange}{\rule{0.5em}{1.5ex}}\textcolor{mplorange}{\rule{0.5em}{2ex}}\textcolor{mplorange}{\rule{0.5em}{1.75ex}}) are not indicative of night performance: a higher mIoU obtained on Cityscapes does not immediately translate into a higher mIoU at night.

\begin{SCfigure}[][t!]
    \centering
    \includegraphics[width=0.5\linewidth]{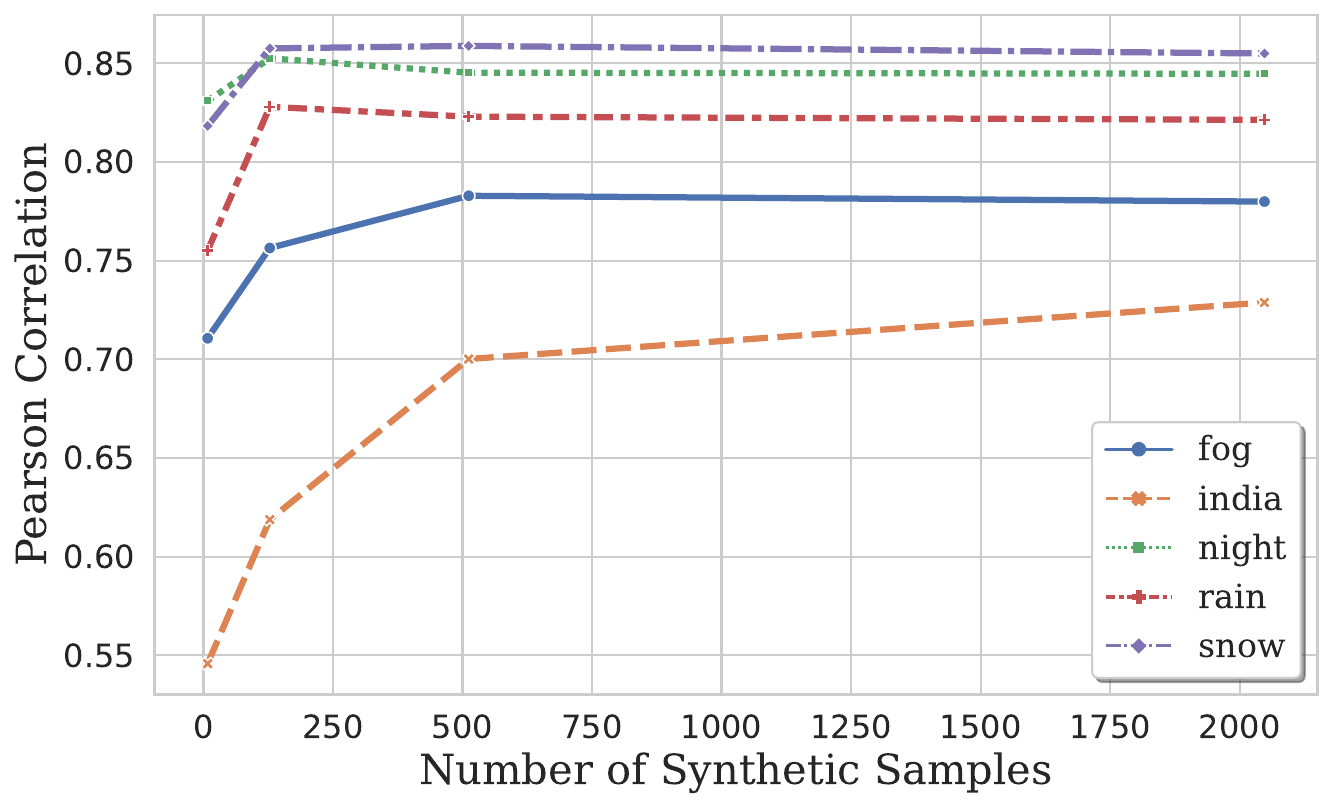}
    \caption{\small \textbf{Pearson Correlation \vs \# Synthetic Samples.} Using more synthetic samples contributes to increased stability in the results. Empirical plots demonstrate that approximately $500$ samples are sufficient for a stable correlation assessment.}
    \label{fig:mioucorr_vs_numsamples}
\end{SCfigure}

Since synthetic data can be generated in any desired quantity, a natural question arises: how many images are sufficient?
In addressing this question, we conducted experiments and presented the results in~\cref{fig:mioucorr_vs_numsamples}.
Our empirical finding suggests that $\sim 500$ synthetic images are adequate for a stable and reliable assessment of robustness.

\noindent\textbf{Discussion.} In their recent work on reliability in semantic segmentation, Jorge et al. \cite{de2023reliability} systematically quantified the robustness of segmenters on real-shift data; similarly to ours, they draw comparisons from ACDC and IDD datasets.
One intriguing finding in this paper is that ``[...] the larger the domain shift, the larger the improvement brought by more recent segmentation models'', hinting at a correlation between model robustness on in-domain data and covariate-shift data; that corresponds to the CS baseline we consider here.
In our study, we delve deeper into this correlation, choosing to separately address different weather types instead of grouping them all together as done in~\cite{de2023reliability}.
In domains exhibiting small gaps to Cityscapes, such as IDD or ACDC-Fog, our conclusion aligns with~\cite{de2023reliability}.
However, as domain gaps increase, the discrepancy between Cityscapes mIoUs and real-shift mIoUs becomes more pronounced, resulting in poor PCC\_CS scores.
On the contrary, synthetic mIoUs and real-shift mIoUs exhibit a strong correlation across shifts.
\textit{Our empirical study has validated that synthetic performance is a reliable indicator of model robustness in the presence of covariate shifts.}\\
{With synthetic data, we are able to confirm observations from the literature on real data, such as (i) the robustness of transformer and ConvNext backbones and that (ii) within an architecture family, robustness correlates with the number of parameters and the robustness of the backbone.}

\noindent\textbf{{Different generative models?}} The important ingredient in our pipeline is the generative model. In~\cref{tbl:other_gans}, we ablate by replacing the default SD1.5 model with other available ones: image-2-image GAN called TSIT~\cite{jiang2020tsit}, physics-based fog simulator~\cite{sakaridis2018semantic}, and a bigger SD variant called SDXL~\cite{podell2023sdxl}.
On `night', `rain', and `snow', SD variants perform better than TSIT, while of note, TSIT was trained on real OOD data.
The dedicated fog simulator performs very well, on par with our results using the larger model SDXL.
Physics-based simulators are definitely valuable and this direction should be investigated further; however, such simulators require in-depth knowledge of OOD domains and hence are very difficult to design. 
Both GAN-based and physics-based approaches are limited in their scalability to many more OOD domains.
Comparing SD1.5 \vs SDXL, we observe comparable results on challenging OOD domains while SDXL performs much better on `fog' and `india'; such results hint at the future potential of stronger and better generative models in further advancing virtual testing.
Unfortunately, as SDXL is much more memory-demanding, we only limit our experiments with SDXL in this particular study.\\
In~\cref{sec:fid_score}, we report the FID scores, which measure the direct distance between synthetic and real distributions; we also extend a discussion on some limitations of our framework.

\begin{table}[t!]
    \begin{tabular}{l|c|c|ccccc}
        \toprule
        & OOD expertise? & OOD data? & Night & Rain & Snow & Fog & India \\
        \midrule
        GAN-based TSIT~\cite{jiang2020tsit}& \textcolor{plt:green}{no} & \textcolor{plt:red}{required} & 0.83 & 0.84 & 0.81 & - & - \\
        Physics-based Fog Sim.~\cite{sakaridis2018semantic}& \textcolor{plt:red}{required}& \textcolor{plt:green}{no} & - & - & - & 0.82 & - \\
        Ours w/ SD1.5 (default)& \textcolor{plt:green}{no}& \textcolor{plt:green}{no}  & \textbf{0.85} & {0.86} & \textbf{0.85} & {0.77} & 0.71 \\
        Ours w/ SDXL~\cite{podell2023sdxl}& \textcolor{plt:green}{no}& \textcolor{plt:green}{no} & {0.84} & \textbf{0.90} & {0.82} & \textbf{0.89} & \textbf{0.93}\\
        \bottomrule
    \end{tabular}
    \caption{\small \textbf{Different generative models.} Our pipeline achieves better performance (PCC with real-shifts) while does not require any OOD knowledge. Of note, it's very easy to {apply to new unseen}
    OOD domains using our pipeline, which remains challenging for GAN-based and physics-based models. }
    \label{tbl:other_gans}
\end{table}

\subsection{Confidence Calibration with Synthetic Data}
\label{sec:calibration}

Confidence calibration is a crucial aspect of deep networks, particularly when employed in safety-critical applications such as autonomous driving.
Jorge et al.~\cite{de2023reliability} highlighted a disconnection between model robustness and calibration, asserting that ``... despite the remarkable improvements in terms of robustness, recent models are not significantly better calibrated''.
Therefore, it is essential to devise techniques and protocols for recalibrating data, particularly in domains exhibiting covariate shifts.
Drawing inspiration from this, we explore the feasibility of using synthetic data to recalibrate pretrained segmenters.

We perform temperature scaling using our synthetic data. Temperature scaling~\cite{guo2017calibration} is a well-established technique for calibrating pretrained models, typically conducted on a small validation set within the {OOD} domain.
In our study, for each segmenter, we utilized the same sets of synthetic data generated in~\cref{sec:robustness_assessment} to optimize temperature scaling factors, with one adjustment made for each covariate shift.
For comparison, we replicate the process using real-shift data from ACDC and IDD.

\begin{SCfigure}[][t!]
    \centering
    \includegraphics[width=0.45\linewidth]{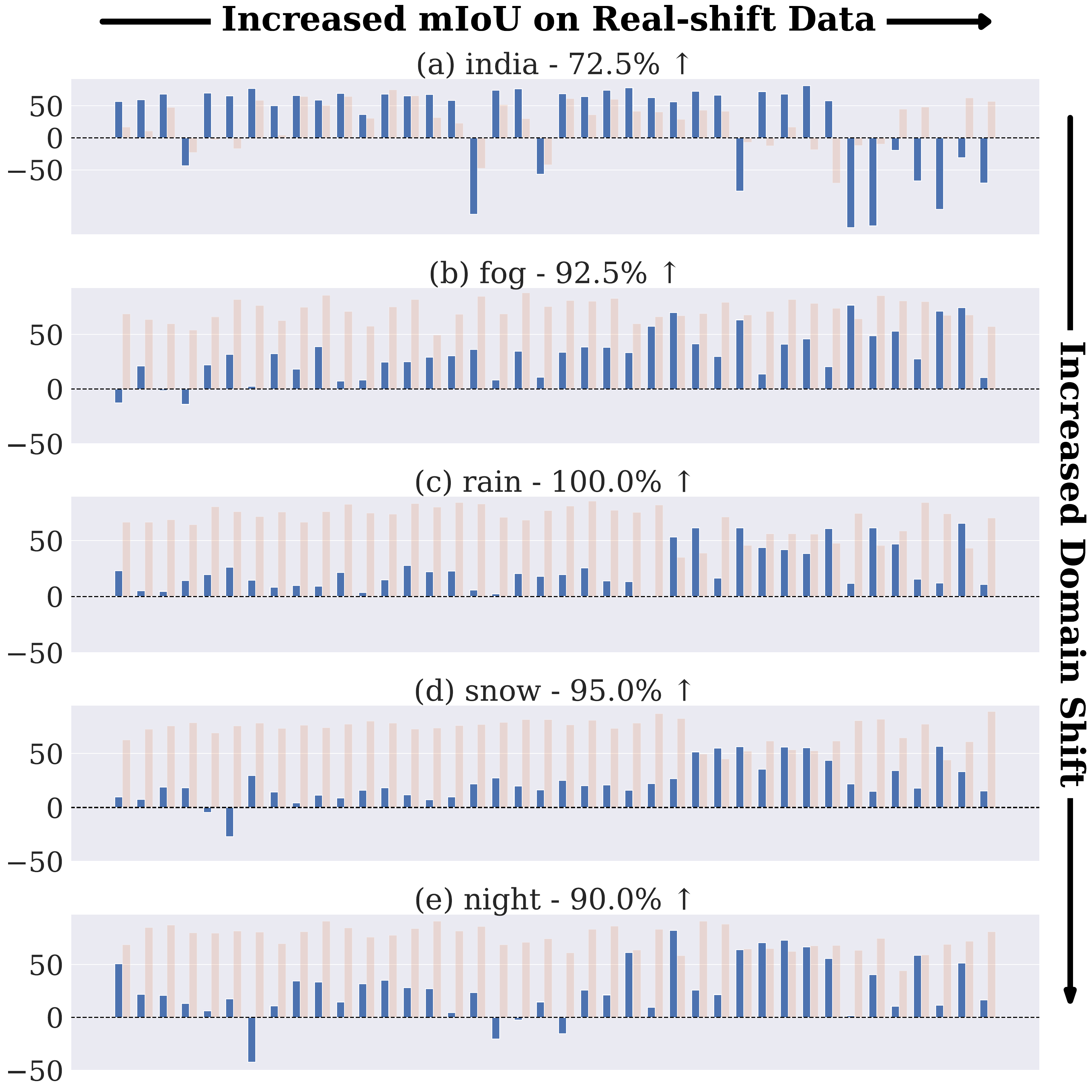}
    \caption{\small \textbf{ECE improvement using synthetic data.} The 40 segmenters are calibrated using either real-shift data or synthetic data. For each model, the relative ECE improvement (\%) over its non-calibrated version is computed, visualized by \textcolor{mplblue}{\rule{0.5em}{2ex}} (synthetic shift) and \textcolor{mplorange}{\rule{0.5em}{2ex}} (real shift). The subplots are ranked by model robustness on real-shift data (left-to-right) and by the increased domain shift (top-to-bottom). 
    The titles of the subplots indicate the percentage of models that showed improvement with synthetic data.}
    \label{fig:ece_improvement}
\end{SCfigure}

~\cref{fig:ece_improvement} reports the calibration improvement for the $40$ pretrained segmenters using either real-shift data (\textcolor{mplorange}{\rule{0.5em}{2ex}}) or synthetic data (\textcolor{mplblue}{\rule{0.5em}{2ex}}).
The subplots are arranged in increasing domain gap order from top to bottom, with the segmenters ranked from left to right based on increasing robustness on real-shift data.
The Expected Calibration Error (ECE)~\cite{naeini2015obtaining} quantifies the calibration results, with a lower ECE indicating a better-calibrated model.
For better interpretation, we present the relative ECE improvement, computed as the percentage decrease in ECE after calibration compared to the original ECE without calibration.
For example, a model with an ECE of 0.4 before calibration and an ECE of 0.2 after calibration will achieve a $(0.4-0.2)/0.4 = 50\%$ relative improvement.

\textit{We observe promising calibration results when employing our synthetic data.} While not as good as real-shift data, synthetic data achieves a promising success rate of $72.5\%$ on IDD and exceeds $90\%$ on the four ACDC shifts.
Interestingly, in weather shifts, we empirically observe that more robust models derive greater benefits when calibrated using our synthetic data; the reverse is observed for `europe-india' geographical shift.
While with real-shift data, robustness and calibration are not well correlated~\cite{de2023reliability}, our results suggest that a potential correlation might exist between the two factors when using synthetic data.
{We note that calibration with temperature scaling does not always guarantee ECE improvement. 
Such phenomenon may happen even using real data, especially under domain shifts as explored in prior work}~\cite{ovadia2019can}.
In~\cref{sec:calib_details_results}, we provide more technical details and results.

\begin{SCfigure}[][h!]
	\centering
    \resizebox{.5\textwidth}{!}{
        \footnotesize
        \setlength{\tabcolsep}{0.001\linewidth}
        \renewcommand{\arraystretch}{0.29}
	\begin{tabular}{ccccc}
        &Syn. Img& Sem.FPN & MobileV3 & SegF-B5 \\
        \rotatebox{90}{\,\,\,\,\,\,Flood}&\includegraphics[width=0.12\textwidth]{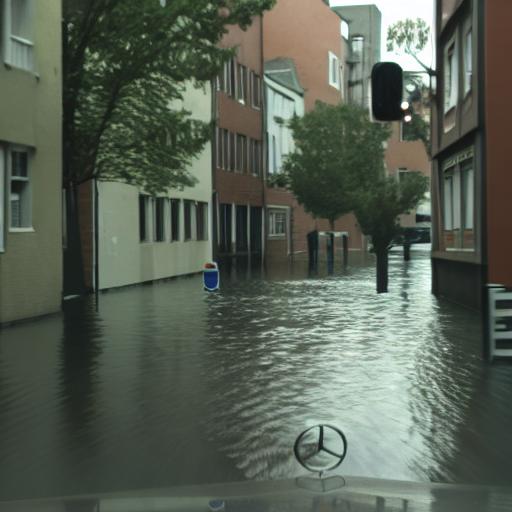} & \includegraphics[width=0.12\textwidth]{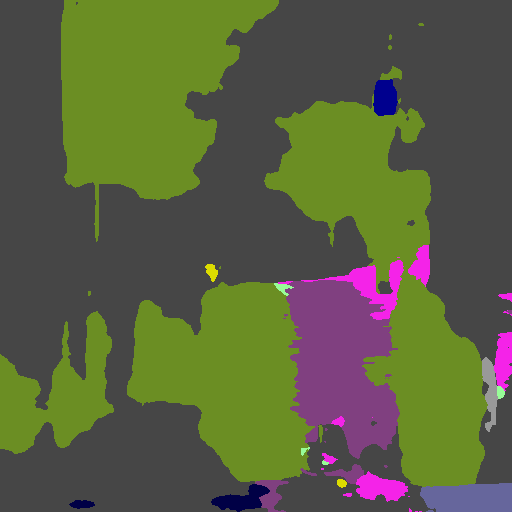} & \includegraphics[width=0.12\textwidth]{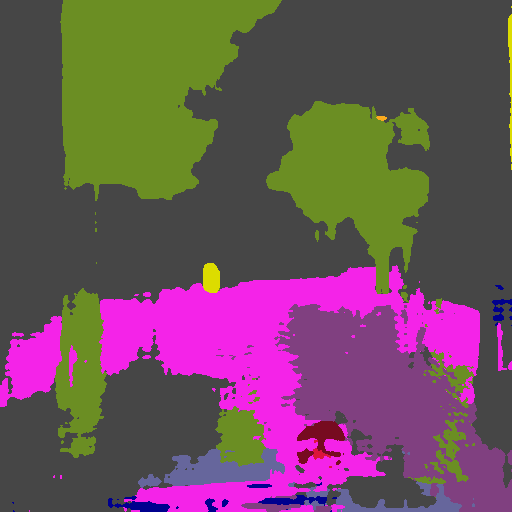} & \includegraphics[width=0.12\textwidth]{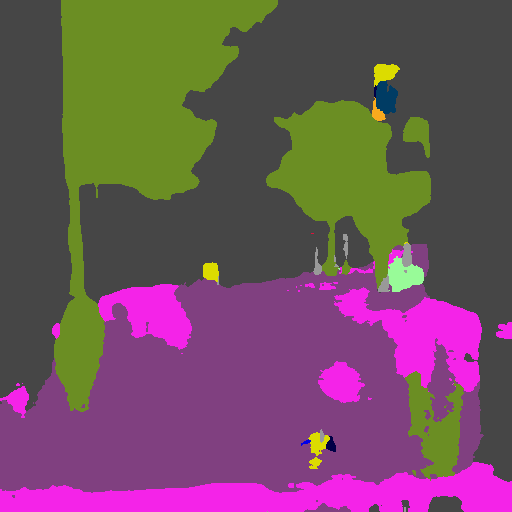} \\
        \rotatebox{90}{\,\,\,Autumn}&\includegraphics[width=0.12\textwidth]{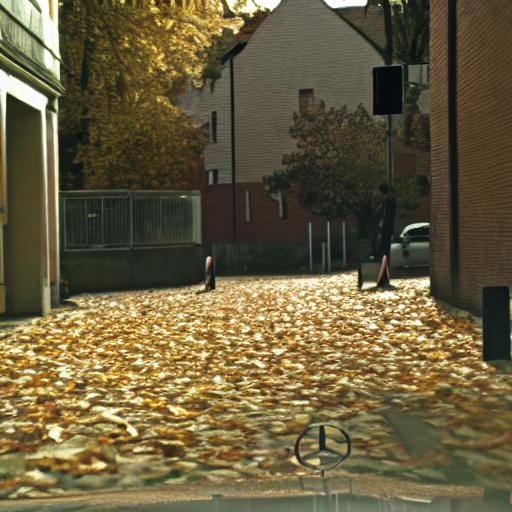} & \includegraphics[width=0.12\textwidth]{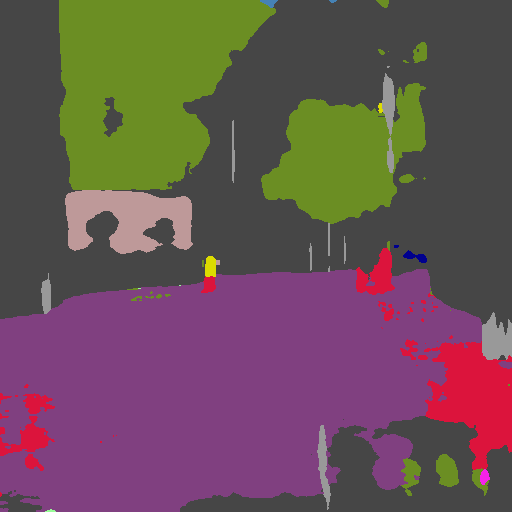} & \includegraphics[width=0.12\textwidth]{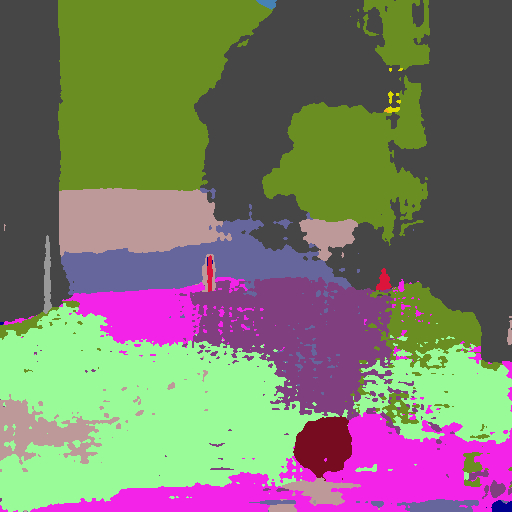} & \includegraphics[width=0.12\textwidth]{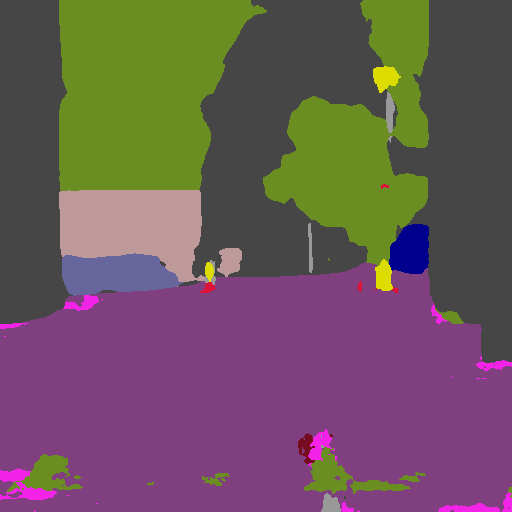} \\
        \rotatebox{90}{\,\,\,\,\,\,\,\,\,Fire}&\includegraphics[width=0.12\textwidth]{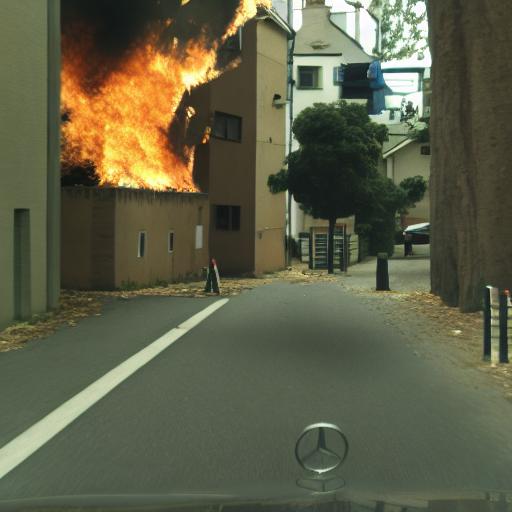} & \includegraphics[width=0.12\textwidth]{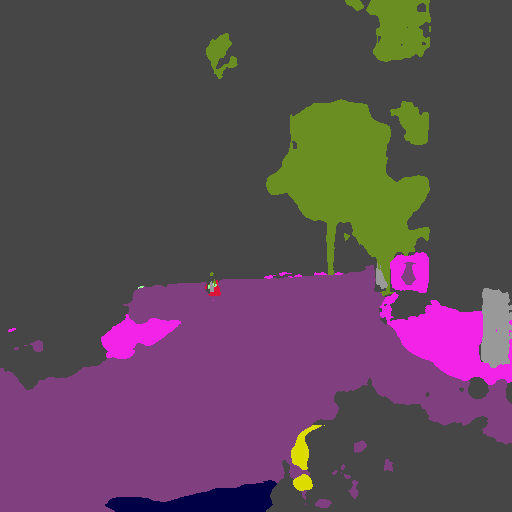} & \includegraphics[width=0.12\textwidth]{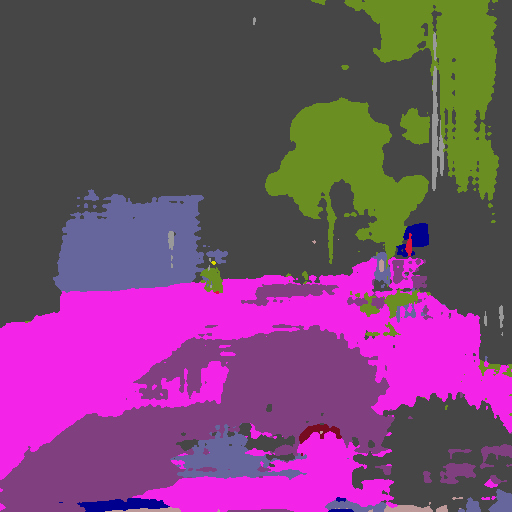} & \includegraphics[width=0.12\textwidth]{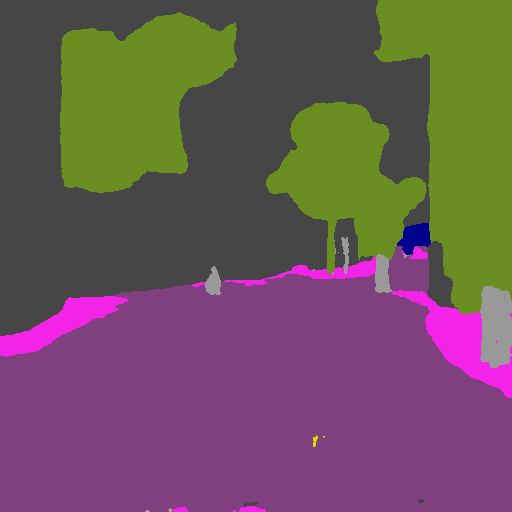} \\
	\end{tabular}
    }
    
    \caption{\small \textbf{Qualitative results.} Examples of rare conditions generated for testing and predictions from different models. Results of the strong model like SegFormer-B5 are visibly better than the Semantic-FPN and MobileNetV3. \bigskip\bigskip\bigskip\medskip}
    \label{fig:qualres_sec3}
\end{SCfigure}

\subsection{On Practical Applicability}
One significant advantage of our framework lies in its potential to address rare conditions simply through prompting.
The practical applicability of generative testing is tremendous.
Our results demonstrate promising signals; practitioners can begin assessing and ranking their pretrained models for a new, unseen {OOD} domain of interest \emph{without the need for real data collection}.
In practice, our proposed generative benchmarking can serve as the initial step in a full validation pipeline, helping filter out non-robust prototypes and thereby saving on total operational costs.
Starting from complementing real-data validation, one can envision a future where generative techniques mature to the point of fully replacing real-data validation.
In~\cref{fig:qualres_sec3}, we visualize some synthetic images and model predictions under rare conditions, such as being flooded with water, having autumn leaves scattered across the road, or having a building on fire.
We observe clear visual distinctions between weaker (Semantic-FPN and MobileNetV3) and stronger (SegF-B5) models, knowing that their Cityscapes scores do not differ significantly.
More examples are provided in~\cref{sec:supp_qual}.

	\section{Reliability Against OOD Objects}
\label{sec:ood}

We now address the reliability of segmentation models in the presence of Out-of-Distribution (OOD) objects.
To begin, we explain our pipeline for inpainting random OOD objects into existing Cityscapes images.
Following that, we demonstrate how one can utilize inpainted images for OOD detection assessment and for enhancing OOD detection.

\subsection{Inpainting Anomaly Objects}
\label{sec:ood:gen}
We inpaint random objects into Cityscapes images.
To this end, we initially sample a location — a square box to which we inpaint the new object.
We crop the box, upsample its content to match the preferred output size of the generative model, and inpaint an object guided by a text prompt.
In this step, we leverage Stable Diffusion inpainting capabilities, obtaining high-definition square images of the desired object.
This image is then resized and pasted back into the original image, creating a final high-definition synthetic image.
To ensure compositional consistency, we employ two techniques:
Firstly, we divide the cropped box into two regions by center cropping it again.
We inpaint only the inner region, leaving the outer region untouched, similar to the approach in RePaint~\cite{LugmayrDRYTG22}.
Secondly, after composing the final image, we address any remaining inconsistencies by applying a light noise over the entire picture and performing reverse diffusion again.
Details and visualizations are provided in~\cref{sec:tech_details}.

After inpainting, it is necessary to extract the mask corresponding to the new object.
To achieve this, we begin with a high-definition square image and apply the Grounded Segment Anything~\cite{kirillov2023segany,liu2023grounding}, prompted with the name of the object.
This process yields a mask within the square image, which can then be repositioned in the full image.

\begin{SCfigure}[][t!]
    \centering
    \resizebox{.45\textwidth}{!}{\includegraphics[width=\linewidth]{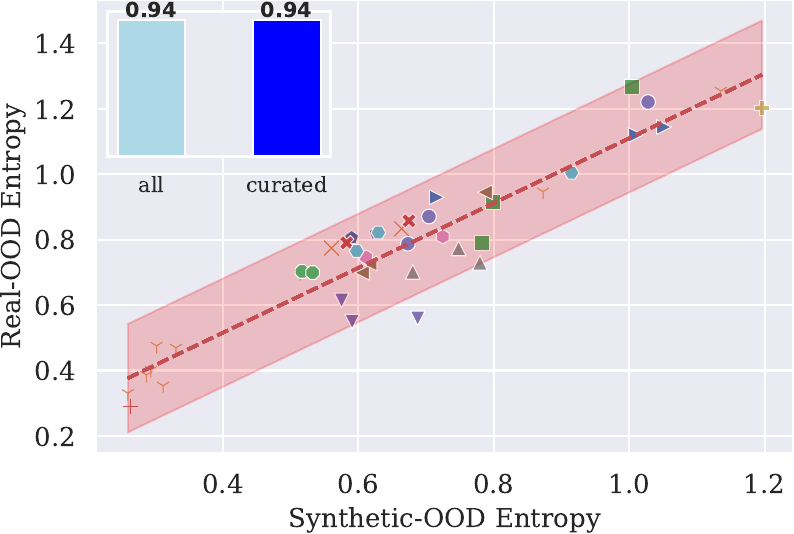}}
    \caption{\small \textbf{Entropy Correlation.} The top-left inset reports the Pearson correlations between real-OOD entropy \vs synthetic-OOD entropy, computed either on `curated` (\textcolor{blue}{\rule{0.5em}{2ex}}) or all synthetic inpainted images (\textcolor{mpllightblue}{\rule{0.5em}{2ex}}). Evaluations are performed on the same model set used in~\cref{fig:mIoU_covariate_shift}, with similar markers.\bigskip}
    \label{fig:ood_ent}
\end{SCfigure}
\begin{figure*}[ht!]
    \centering
    \includegraphics[width=\linewidth]{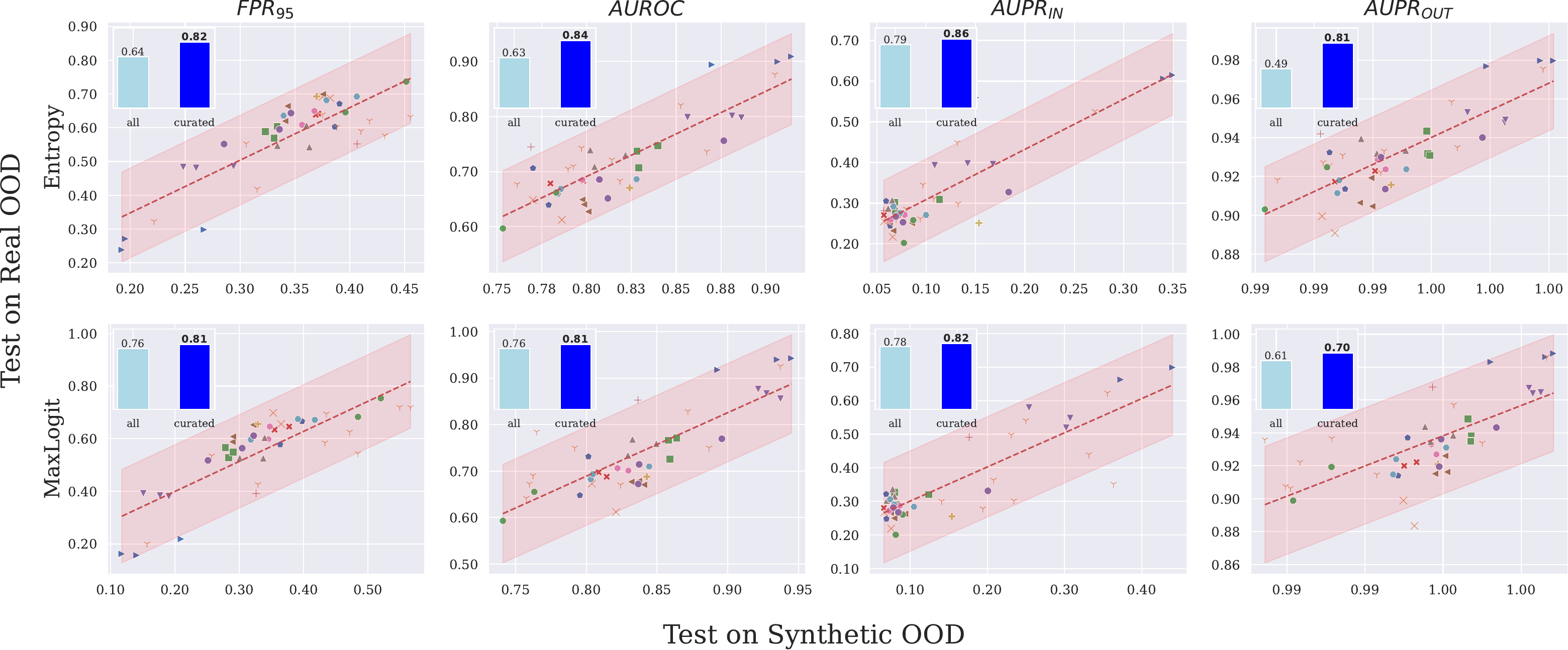}
    \caption{\small \textbf{Correlation in OOD Detection.} Each subplot scatters computed anomaly scores of segmenters on real OODs (y-axis) and on synthetic OODs (x-axis). The top row shows the four anomaly metrics utilized: FPR$_{95}$, AUROC, AUPR$_{IN}$, and AUPR$_{OUT}$. The results are organized into two rows corresponding to two different confidence measures (i) Entropy and (ii) MaxLogit.
    In the top-left corner of each subplot, an inset plots Pearson correlations to real OOD for `curated` (\textcolor{blue}{\rule{0.5em}{2ex}}) and `all` (\textcolor{mpllightblue}{\rule{0.5em}{2ex}}) synthetic sets.
    Evaluations are performed on the same model set used in~\cref{fig:mIoU_covariate_shift}, with similar markers. }
    \label{fig:ood}
\end{figure*}
Our end-to-end generation pipeline is fully automatic.
Through qualitative assessment, we achieve a satisfactory success rate in terms of generation realism; some inpainted images are illustrated in~\cref{fig:qualres_sec4} and much more in~\cref{fig:supp_sec4_quali}.
However, this still leaves a few generations with artifacts, characterized by either unusual compositions or unrealistic details.
\textit{We here question the criticality of realism in assessing OOD detection and, furthermore, in improving OOD detection.}
To this end, we construct two different sets: (i) all 23,040 images generated automatically and (ii) 656 curated images where we manually select the best images in terms of visual quality and realism; {specifically, curators filter out strong color saturation differences or partial objects, \eg animal heads.}
We note that the manual selection process for the curated set is not exhaustive and is constrained by our allocated resources; there are many more high-quality images in the `all' set.
In what follows, we present results using both curated and uncurated sets.

\subsection{Assess OOD Detection}
\noindent\textbf{Experimental setup.} To measure how the segmenters react to unseen OOD objects, we use standard anomaly detection metrics~\cite{yang2022openood}, which are False Positive Rate at 95\% true positives ($\text{FPR}_\text{95}$), Area Under ROC curve ($\text{AUROC}$), and Area Under Precision-Recall curve ($\text{AUPR}$). AUPR are declined into $\text{AUPR}_{\text{IN}}$ and $\text{AUPR}_{\text{OUT}}$, which consider the in-distribution regions, respectively the out-of-distribution regions (the inpainted object), as positive regions to compute the Precision-Recall curves.

All segmenters in our study are not designed to produce confidence scores.
We thus seek various techniques to derive confidence scores from pretrained models~\cite{hendrycks2019scaling} and eventually narrow down the options to two measures: (i) Entropy of soft-probability predictions, and (ii) MaxLogit as the maximum logit value (before softmax) among the classes.
While Entropy is the traditional measure of uncertainty, MaxLogit is a recent and surprising finding that has been proven to be more effective in estimating OOD confidence~\cite{hendrycks2019scaling}.

\noindent\textbf{Results.}
For quantitative comparison, we leverage the SegmentMeIfYouCan (SMIYC) dataset~\cite{chan2021segmentmeifyoucan}, a recent dataset for OOD detection.
We resort to the RoadAnomaly21 split in SMIYC, due to similarity in object scales to our synthetic data.
We analyze the correlation between the OOD scores obtained on RoadAnomaly21 and one using our synthetic inpainted data.
~\cref{fig:ood_ent} reports our first analysis on the entropy in the OOD areas, either real or generated.
For each model, we compute the Pearson Correlation (PCC) between real-OOD entropy and synthetic-OOD entropy; the computation is done on both `curated` and `all` sets.
We observe a very high entropy correlation between real- and synthetic-OOD, reaching $0.94$ PCC using both `curated` and `all` sets.
In~\cref{fig:id_vs_ood}, we show a control experiment in which we inpaint in-domain object class `car` into Cityscapes scenes, and we analyze models' responses to synthetic cars, real cars, and OOD objects. 

\begin{figure*}[t!]
	\setlength{\tabcolsep}{0.002\linewidth}
    \setlength{\fboxsep}{0pt}
    \setlength{\fboxrule}{1.5pt}
	\centering
    \scriptsize
	\begin{tabular}{cccccc}
        &&\multicolumn{4}{c}{
                  \begin{tikzpicture}
                    \draw[->,>=latex, line width=2.5pt] (-6.6,0) -- (2,0) node[midway,fill=white] {\textbf{Better Models in AUROC on Synthetic}};
                  \end{tikzpicture}
                } \\
        &Images& Semantic-FPN & MobileNetV3 & SegFormer-B5 & SETR-PUP \\
        \rotatebox{90}{\,\,\,\,\textcolor{mplorange}{Real}}&\fcolorbox{mplorange}{white}{\includegraphics[width=0.18\textwidth]{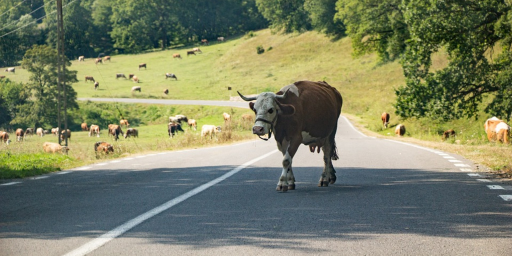}} & \includegraphics[width=0.18\textwidth]{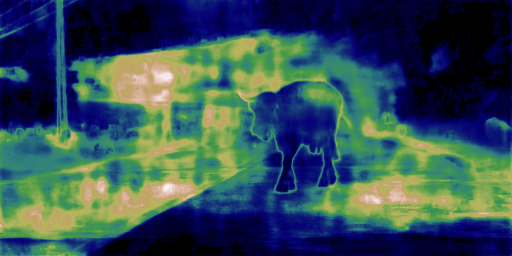} & \includegraphics[width=0.18\textwidth]{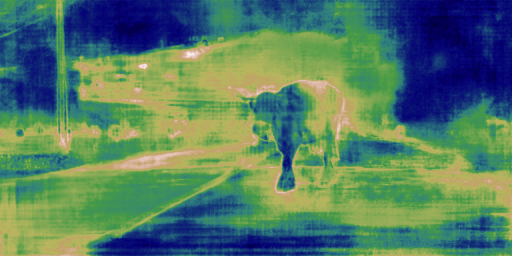} & \includegraphics[width=0.18\textwidth]{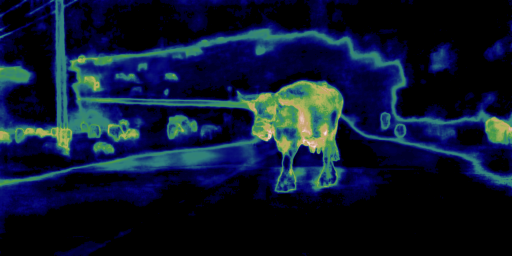} & \includegraphics[width=0.18\textwidth]{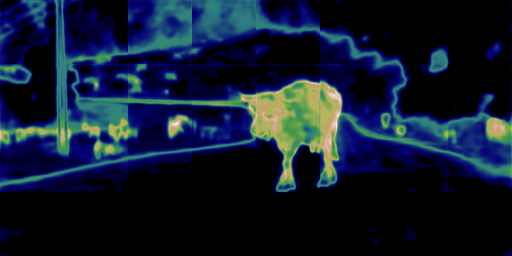} \\
        \rotatebox{90}{\,\,\,\,\,\textcolor{mplblue}{Syn.}}&\fcolorbox{mplblue}{white}{\includegraphics[width=0.18\textwidth]{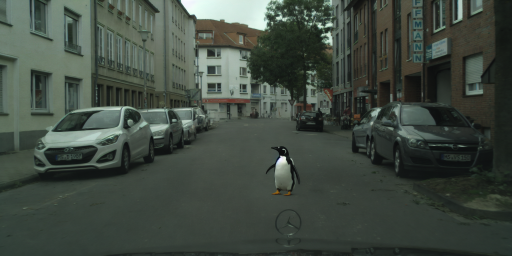}} & \includegraphics[width=0.18\textwidth]{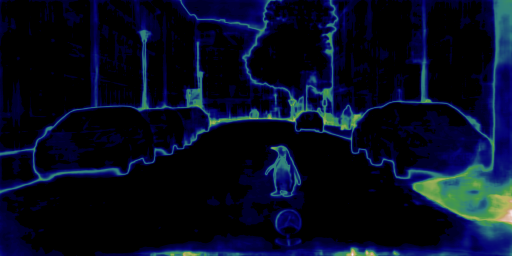} & \includegraphics[width=0.18\textwidth]{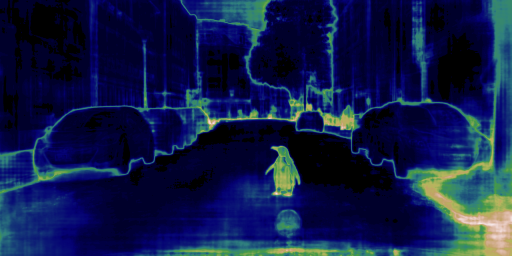} & \includegraphics[width=0.18\textwidth]{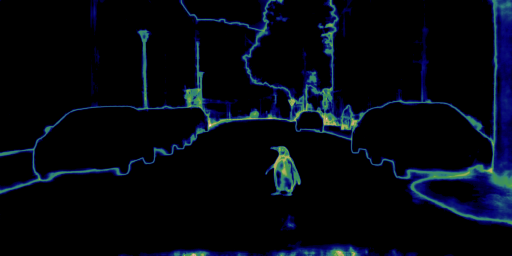} & \includegraphics[width=0.18\textwidth]{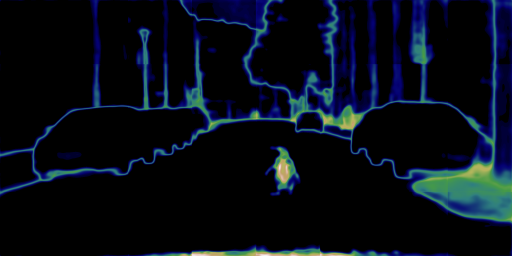} \\
        \rotatebox{90}{\,\,\,\,\,\textcolor{mplblue}{Syn.}}&\fcolorbox{mplblue}{white}{\includegraphics[width=0.18\textwidth]{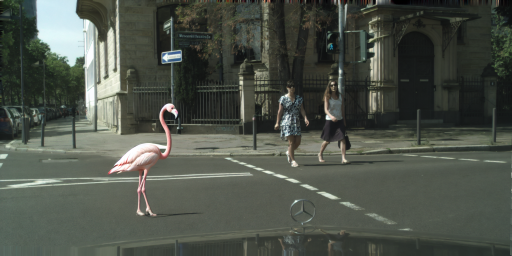}} & \includegraphics[width=0.18\textwidth]{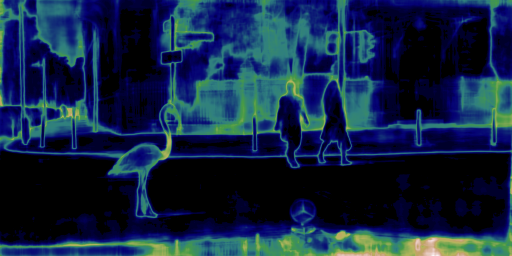} & \includegraphics[width=0.18\textwidth]{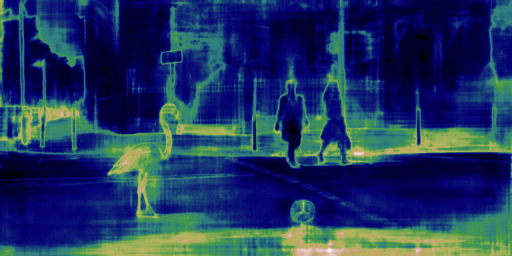} & \includegraphics[width=0.18\textwidth]{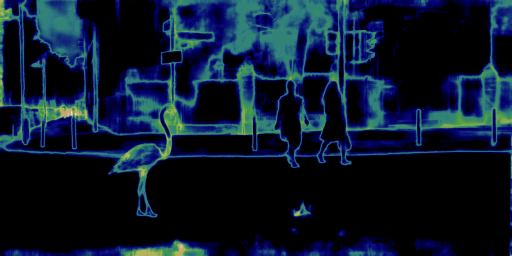} & \includegraphics[width=0.18\textwidth]{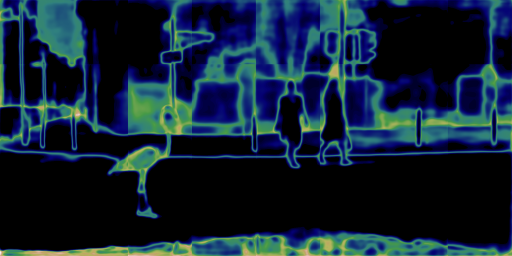} \\
        
	\end{tabular}
    \smallskip
    \caption{\small \textbf{Qualitative results.} Confidence maps are visualized for the four exemplified models on real data (first row) and synthetic inpainted data (second and third rows). Hotter colors correspond to higher OOD likelihood. Ideally, results should exhibit hot colors in OOD areas and cold colors everywhere else. We observe a strong correlation in model reactions to real and synthetic OOD regions, particularly for more recent and robust models. {We note that the real data in SMIYC also exhibit distributional shifts against Cityscapes; which already causes confusion to weak models in the background}.}
    \label{fig:qualres_sec4}
\end{figure*}
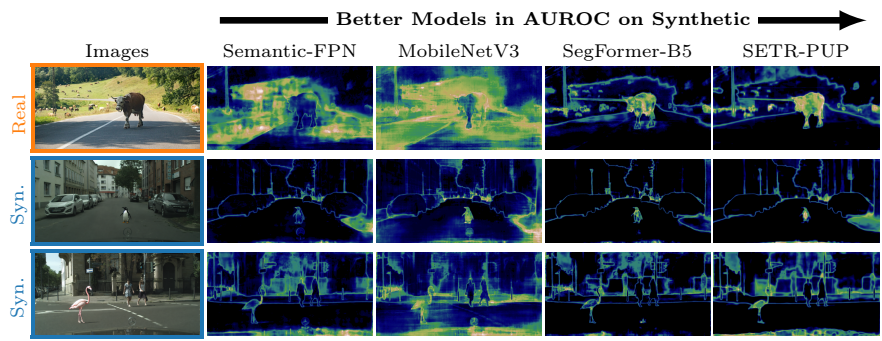

\begin{SCfigure}[][h!]
    \centering
    \includegraphics[width=0.5\linewidth]{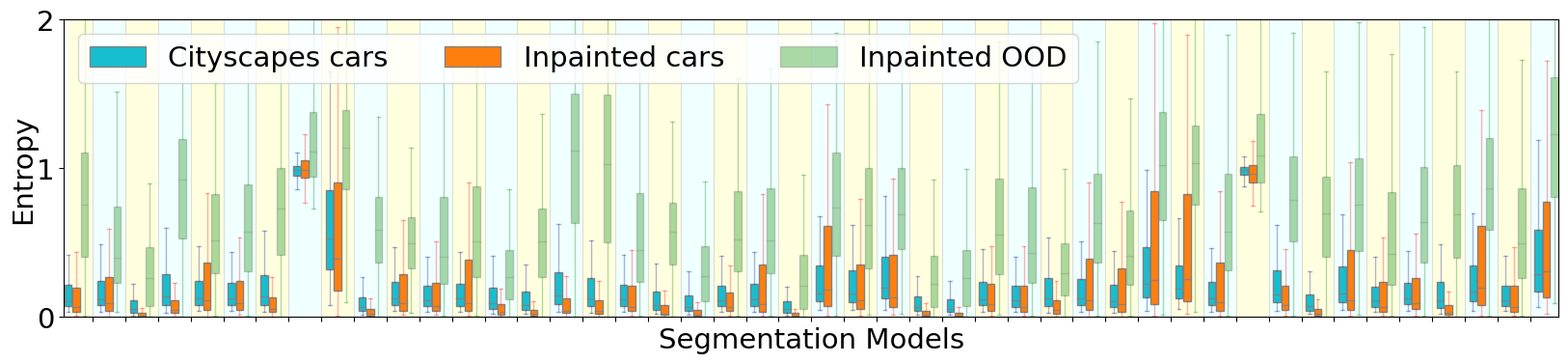}
    \caption{\small \textbf{In-domain vs OOD inpainted objects.} Models' responses to synthetic cars are close to real cars, and far from synthetic OOD objects}
    \label{fig:id_vs_ood}
\end{SCfigure}

We then analyze the correlation between real and synthetic anomaly metrics.
~\cref{fig:ood} presents our primary findings and~\cref{fig:qualres_sec4} illustrates some qualitative results.
We observe a strong correlation with real scores when utilizing the 'curated' set for computing synthetic scores; the curated PCCs (\textcolor{blue}{\rule{0.5em}{2ex}}) are consistently around 0.8 across multiple metrics, irrespective of the two confidence measures.
Although the correlations are somewhat weaker when using all uncurated synthetic data (\textcolor{mpllightblue}{\rule{0.5em}{2ex}}), such results remain acceptable, particularly when no effort is dedicated to curation.

\textit{Our results validate the potential of utilizing realistic synthetic data, inpainted with anomaly objects, for assessing OOD detection.
In OOD testing, it is quite important to use high-quality synthetic inpainted data. Nonetheless, even non-curated synthetic data can offer an acceptable estimation of real performance with minimal curation costs.}

\begin{SCtable}[][ht!]
    \centering
    \small
    \begin{tabular}{lccc}
        \toprule
        \textbf{Method} & \textbf{AUROC} ($\uparrow$) & \textbf{AUPR} ($\uparrow$) & \textbf{FPR95} ($\downarrow$) \\
        \midrule
        RbA~\cite{nayal2023rba} Swin-B & 95.6 & 78.4 & 11.8\\
        \,\,$+$ COCO~\cite{nayal2023rba} & \textbf{97.8} & \textbf{85.3} & {8.5} \\
        \,\,$+$ Ours (curated)& {97.2} & \uline{84.9} & \textbf{8.1} \\
        \,\,$+$ Ours (all)& \uline{97.3} & {84.8} & \uline{8.2} \\
        \midrule
        RbA~\cite{nayal2023rba} Swin-L & 96.4 & 79.6 & 15.0 \\
        \,\,$+$ COCO~\cite{nayal2023rba} & \textbf{98.2} & \textbf{88.7} & \uline{8.2} \\
        \,\,$+$ Ours (curated)& {97.2} & 88.0 & \textbf{7.9} \\
        \,\,$+$ Ours (all)& \uline{98.1} & \uline{88.6} & 8.3 \\
        \bottomrule
    \end{tabular}
    \caption{\small \textbf{Improving OOD detection} on real SMIYC benchmarking using our synthetic data. All results are obtained using the published code and default parameters.}
    \label{tbl:ood_improvement}
\end{SCtable}

\subsection{Improve OOD Detection}
\label{sec:improve_ood_detection}
In this experiment, we investigate if synthetic inpainted data can be used to enhance a deep network's ability to detect OOD objects.
To this end, we adopt the state-of-the-art RbA~\cite{nayal2023rba} approach for OOD detection and train RbA models on our synthetic data.

OOD metrics are computed on RoadAnomaly21 and reported in~\cref{tbl:ood_improvement}.
The RbA models trained on our data significantly outperform the vanilla RbA model.
We reach comparable performance to the RbA variants that leverage the external COCO dataset for augmentation.
Notably, there are no clear differences between using `curated' or `all' sets.
\textit{We conjecture that, unlike benchmarking, training for OOD detection does not demand a high degree of realism from synthetic data.}
This explains why the simple strategy of copy-pasting COCO objects~\cite{nayal2023rba} already proves effective.
All results are consistent across the two addressed backbones. ~\cref{fig:qualres_sec4} illustrates a few qualitative results.

\noindent\textbf{{Discussion.}} {In our work, we take all available levers and study the extent to which synthetic data can be used for evaluation and specific training purposes. We acknowledge the fact that the models used in this work were trained on substantial amounts of data. Of note, we do not claim the efficacy of synthetic data in all aspects, particularly regarding the total amount of training data required. Our findings are limited to highlighting the significant potential of published generative models in the task of reliability assessment. Advances in efficient training of generative models may address concerns regarding data quantity but are beyond the scope of this work.}
	\section{Takeaways}
In this work, we explore the potential of synthetic data in reliability assessment for semantic segmentation networks. We introduce two automatic zero-shot pipelines to generate data in OOD domains and to inpaint OOD objects for virtual reliability assessment.
Our promising results encourage further collective investigations into this research problem, paving the way for synthetic system validation, especially in safety-critical applications.
We summarize here our findings:
\begin{itemize}[label=$\triangleright$]
  \item \textit{Reliability Under Covariate Shifts:} synthetic data can help assess the relative robustness of models in real-life covariate shifts, especially when shifts to the training condition are significant. Synthetic data can well complement real data in system validation, helping reduce the total operational cost. Pretrained models can be calibrated using synthetic data to better estimate prediction confidence in any shifted domains.
  \item \textit{Reliability Against OOD Objects:} synthetic data is useful in both OOD testing and OOD training; however, the demands on synthetic data quality differ in these two cases. In OOD testing, the best result estimations are obtained with the most realistically inpainted data, which may require a certain amount of curation time for qualitative assessment. The curation task is not time-demanding and can be done quickly with a reasonable budget. On the other hand, for OOD training, no curation is actually needed to achieve improvements.
\end{itemize}
\label{sec:conclusion}

	\section*{Acknowledgements}
This work is supported by ELSA - European Lighthouse on Secure and Safe AI funded by the European Union under grant agreement No.~101070617. We thank the authors of RELIS~\cite{de2023reliability} for providing us their pretrained checkpoints.
	\normalsize
\appendix

In this document, we provide technical details for covariate shifts generation and OOD object inpainting (\cref{sec:tech_details}), additional calibration details and results (\cref{sec:calib_details_results}), analysis on direct real-\vs-synthetic data correlation (\cref{sec:fid_score}) and class-wise PCC scores (\cref{sec:class_wise}).~\cref{sec:limitations} discusses the limitations and~\cref{sec:supp_qual} showcases more qualitative examples.
\setlength{\skip\footins}{5pt}
\renewcommand{\thefootnote}{}
\footnotetext{\dag~Work done during an internship at valeo.ai} 
\renewcommand{\thefootnote}{\arabic{footnote}}
\section{Technical Details}
\label{sec:tech_details}
\subsection{Covariate Shifts Training}
We train a ControlNet on top of a frozen Stable Diffusion 1.5 for 2100 steps. The ControlNet used here is a trainable copy of the Stable Diffusion encoder only, as in the original paper~\cite{zhang2023adding}. We use a batch size of 8 with 32 gradient accumulation steps, which makes an effective batch size of 256, and a learning rate of $10^{-5}$.
We use the training set of Cityscapes, and do a random horizontal crop of the images to get square images, and then resize them to 512 $\times$ 512, convenient of Stable Diffusion 1.5.
All other training hyperparameters are the per default settings on the official ControlNet repository.
The objective is to reconstruct the original images of Cityscapes using its semantic masks as input to the ControlNet, and the captions extracted with CLIP-interrogator as input to Stable Diffusion. 

Similarly for SDXL, we train a ControlNet for 27500 steps and use a batch size of 8 with 4 gradient accumulation steps, which makes an effective batch size of 32. The learning rate is $10^{-5}$. The images are square cropped to get 1024 $\times$ 1024 images, convenient for SDXL, and there is no need to resize them.

\subsection{Covariate Shifts Generation}
To generate images with new styles with SD 1.5, we take a semantic mask from the validation set of Cityscapes, crop and resize it as explained in the previous part.
We use nearest neighbor interpolation to keep good values for specific classes. We only use the part of the caption extracted with CLIP-interrogator that corresponds to a BLIP caption.
To this new caption, we add \texttt{[, in $<$domain$>$]} depending on the domain we want to generalize to. Starting from pure noise, we use 25 DDIM steps with a guidance scale of 8. On a RTX 2080, one new image is generated in about 4 seconds.
All other sampling hyperparameters are the per default settings on the official ControlNet repository.

For SDXL, we tune the conditioning strength and the prompt strength to get better images. To get both mask and prompt aligned images, we set the ControlNet conditioning strength to 0.65 and the prompt guidance to 10. The prompt tuning is the same as used for SD 1.5; we also use 25 denoising steps.

\subsection{OOD Objects Generation}
\label{sec:supp:object_generation}
\begin{figure*}
    \centering
    \includegraphics[width=\textwidth]{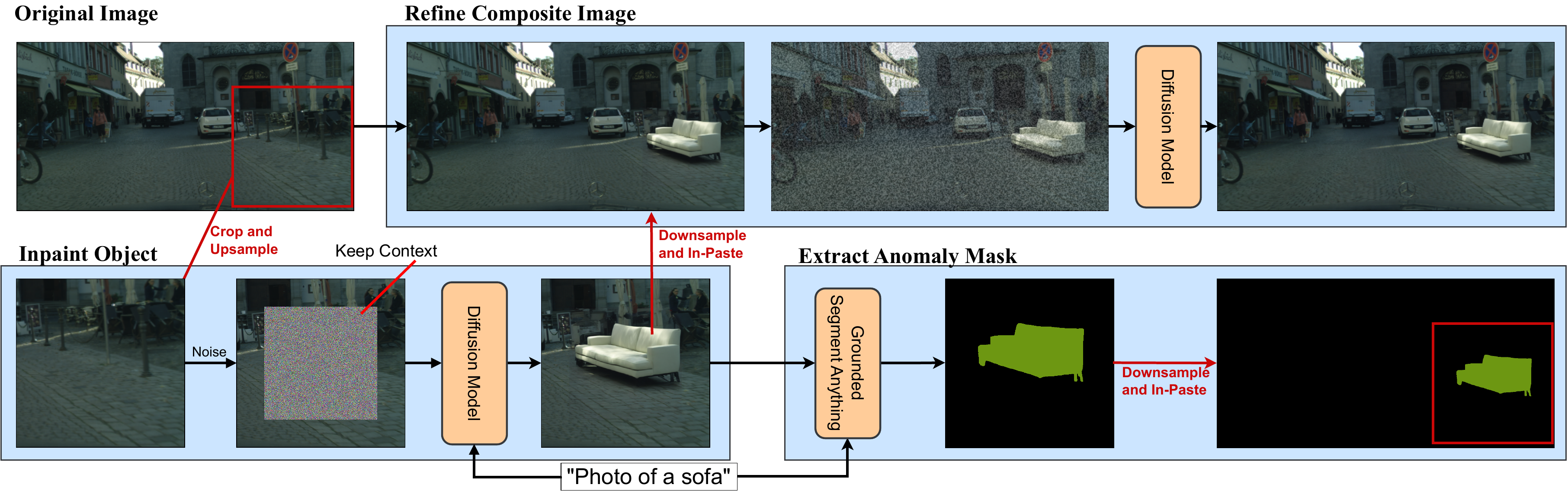}
    \caption{\textbf{OOD object data generation pipeline.}
    We use pretrained Stable Diffusion for inpainting and refining steps, and pretrained Grounded Segment Anything for mask extraction.}
    \label{fig:object_generation}
\end{figure*}

\begin{figure}[ht]
    \centering
    \includegraphics[width=\linewidth]{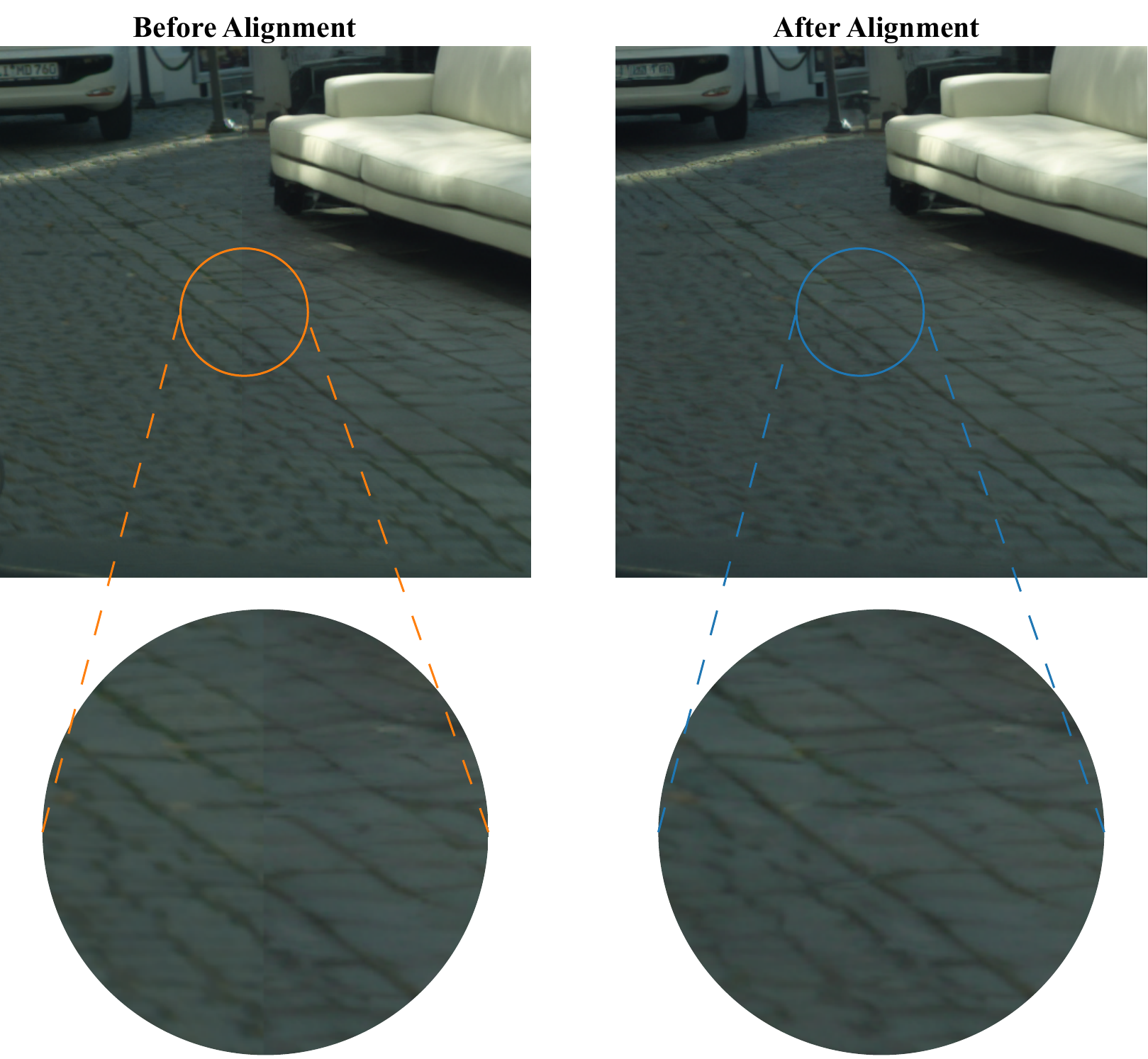}
    \caption{\textbf{Refinement.}
    This example highlights the importance of the refinement step. The left image shows the state before refinement, whereas the right image displays the refined version. Upon zooming into the edge of the inpainting box, a clear distinction between the left and right is evident in the first image. Such difference is eliminated in the second image. Must be viewed in color.
    }
    \label{fig:object_artifacts}
\end{figure}

The OOD object generation pipeline described in Sec. 4.1 is further illustrated in \cref{fig:object_generation}.
It is composed of three steps.
Given a text prompt containing an object, the inpainting step generates a zoomed-in version of the object with the appropriate close-range background given as context. 
The mask extraction step infers the anomaly mask from the zoomed-in generated image and the name of the object. 
Both are in-pasted back in the original complete image or mask.
To reduce some composition artifacts, the composite image is refined with a noise/denoise step.

In details, we first randomly choose a box size for the new object, uniformly sampled between a quarter and half the minimum dimension of the original image.
We also uniformly sample a location for the box in the bottom three quarters of the image.
This box will contain the new object we wish to add, and we refer to it as \textit{inpainted region}.
In addition to the inpainted region, we create a larger box, 1.5$\times$ its height and width, with the inpainted region in its center. 
The contour outside of the inpainted region will serve as \textit{context} for the inpainting process.
We then crop and resize the inpainted region with its context to a resolution of 512 $\times$ 512.
We fully noise the inpainted region, but leave the clean context.
We then denoise the inpainted zone with the prompt \textit{``A photo of an $[$object$]$''}, with a guidance scale of 15.
The full patch is then resized and pasted on the original image, at its original position. 
Some artifact might be still present as shown in \cref{fig:object_artifacts}.
To remedy this, we refine the inpainted zone by noising and denoising it with 0.65 strength, with the default guidance scale of 7.5.
The effect of this refining step is shown in \cref{fig:object_artifacts}.

We list here all 42 objects used in our experiments, which are not present in Cityscapes' classes: arcade machine,
armchair,
baby,
bag,
bathtub,
bench,
billboard,
book,
bottle,
box,
chair,
cheetah,
chimpanzee,
clock,
computer,
desk,
dolphin,
elephant,
flamingo,
giraffe,
gorilla,
graffiti,
hippopotamus,
kangaroo,
koala,
lamp,
lion,
microwave,
mirror,
panda,
penguin,
pillow,
plate,
polar bear,
radiator,
refrigerator,
sofa,
table,
tiger,
toilet,
vase,
and zebra.

\subsection{OOD Detection Training}

For the OOD detection method in Sec. 4.2, we used the codebase of~\cite{nayal2023rba}.
We adapt the code to be able to use our generated data with the binary masks extracted from Grounded-SAM, as explained in~\cref{sec:supp:object_generation}.
As in the original paper~\cite{nayal2023rba}, we fine-tune the mask prediction MLP and classification layer after the transformer decoder.
To obtain all OOD detection results reported in the main paper (Tab. 2), we used the recommended hyperparameters, and train the models for 5000 iterations.

\subsection{Segmentation Models}
We list here all models used in our experiments: 
\small{ANN-R101,
ANN-R50,
APCNet-R101,
APCNet-R50,
BiSeNetV1-R50,
BiSeNetV2-FCN,
CCNet-R101,
CCNet-R50,
ConvNext,
ConvNext-B-In1K,
ConvNext-B-In21K,
DLV3+ResNet101,
DLV3+ResNet18,
DLV3+ResNet50,
GCNet-R101,
GCNet-R50,
ICNet-R101,
ICNet-R18,
ICNet-R50,
MobileNetV3,
PSPNet-R101,
PSPNet-R18,
PSPNet-R50,
SETR-MLA,
SETR-Naive,
SETR-PUP,
SegFormer-B0,
SegFormer-B1,
SegFormer-B2,
SegFormer-B3,
SegFormer-B4,

\noindent SegFormer-B5,
SegFormer-B5-v2,
SegFormer-B5-v3,
Segmenter,
SemFPN-R101,

\noindent SemFPN-R50,
UpperNet-R101,
UpperNet-R18,
UpperNet-R50.}

\normalsize

\section{Calibration Details and Additional Results}
\label{sec:calib_details_results}
We elaborate on our strategy for performing per-class calibration to obtain the synthetic results (\textcolor{mplblue}{\rule{0.5em}{2ex}}) presented in Fig.~6.
Utilizing our synthetic data, a temperature scaling (TS) scalar is learned for each class.
When calibrating models on shifted domains, we choose the corresponding TS scalar based on model predictions. In the case of calibration results with real-shift data (\textcolor{mplorange}{\rule{0.5em}{2ex}}), only one scalar is learned for each model.

\begin{figure*}[h!]
    \centering
  
    \begin{subfigure}[b]{\linewidth}
    \centering
    \includegraphics[width=0.8\textwidth]{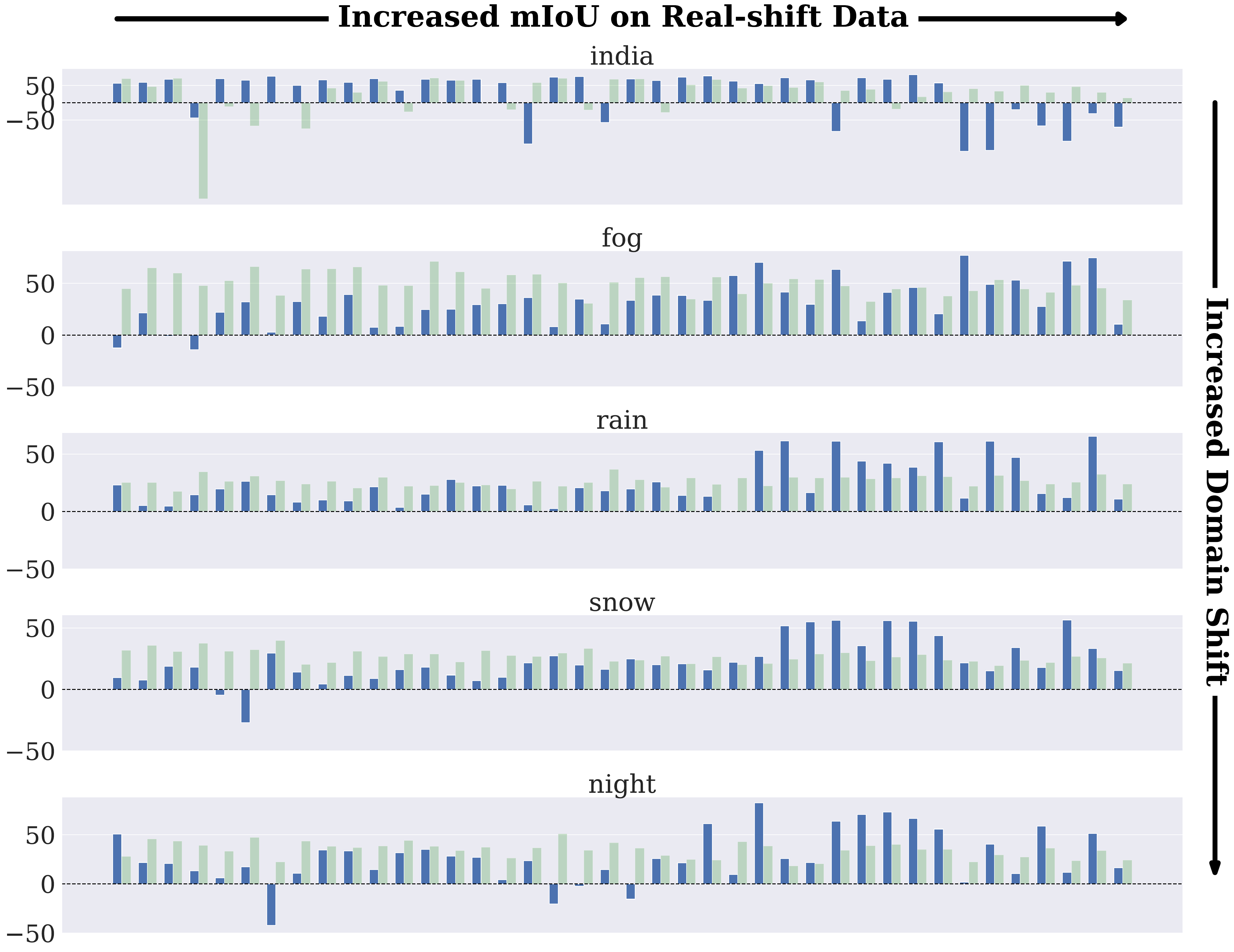}
    \caption{\textbf{Per class Temperature Scaling (\textcolor{mplblue}{\rule{0.5em}{2ex}}) \textit{vs.} One Temperature Scaling (\textcolor{mplgreen}{\rule{0.5em}{2ex}}).} The Figure has the same structure as of Fig. 6 and the bars show relative ECE improvements. We compare the two strategies for performing calibration using synthetic data; both enhance calibration in shifted domains.}
    \label{fig:ece_improvement_suppmat_upper}
    \end{subfigure}
    
    \begin{subfigure}[b]{\linewidth}
    \centering
    \includegraphics[width=0.8\textwidth]{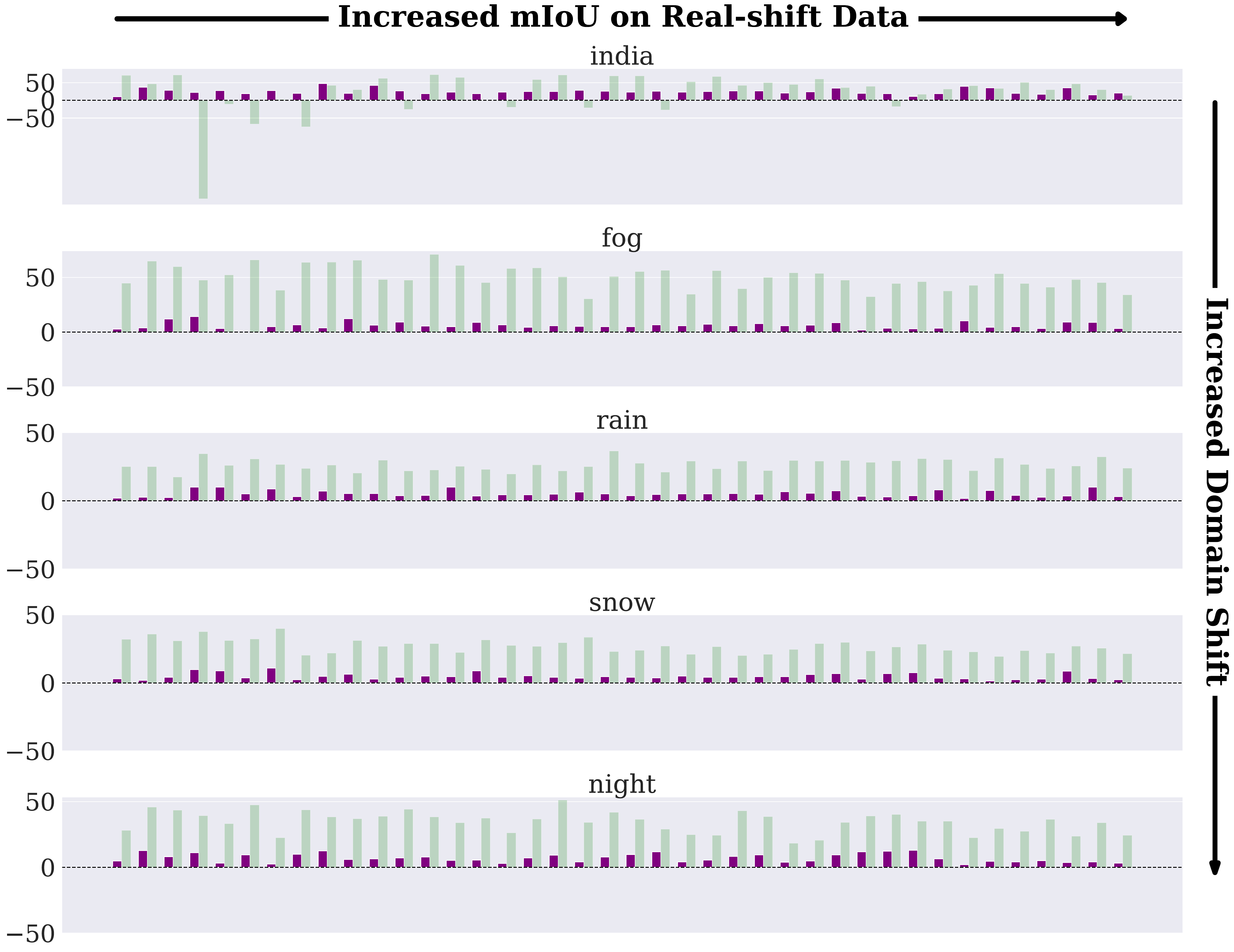}
    \caption{\textbf{Cityscapes (\textcolor{mplpurple}{\rule{0.5em}{2ex}}) \textit{vs.} our synthetic data with one TS (\textcolor{mplgreen}{\rule{0.5em}{2ex}}).} The Figure has the same structure as of Fig. 6 and the bars show relative ECE improvements. Our synthetic data is superior to Cityscapes in improving calibration in shifted domains.}
    \label{fig:ece_improvement_suppmat_lower}
    \end{subfigure}

    \caption{\textbf{Additional Calibration Results.}}
    \label{fig:ece_improvement_suppmat}
\end{figure*}

In~\cref{fig:ece_improvement_suppmat_upper}, we compare per-class TS \textit{vs.} standard TS with one scalar per model, both applied on our synthetic data.
Both strategies enhance calibration results, highlighting the advantage of employing synthetic data for calibration.
Per-class TS demonstrates superiority for more robust models (right part of the plots), while its performance is weaker for less robust ones (left part of the plots).

In~\cref{fig:ece_improvement_suppmat_lower}, we compare Cityscapes \textit{vs.} our synthetic data; in this experiment we adopt the standard TS with one scalar per model.
The results obtained from Cityscapes are clearly inferior to those achieved using our synthetic data, demonstrating the limitations when relying solely on in-domain data for calibration in shifted domains.

\smallskip\noindent\textbf{Generalization.} We ask ChatGPT the generic question like \emph{``give me different cities/weathers that would be representative of the whole world?''}.
We then use all ChatGPT's answers as prompts to generate OOD data, referred to as ``all-domains''. For confidence calibration, using ``all-domains'' shows comparable results as using domain-specific data with manual prompts (see \cref{tbl:fid_scores}), e.g. $100\%$ ECE improvement for \emph{rain}, $78.6\%$ for \emph{india} (cf. Fig.6).
Interestingly, using ``all-domains-but-rain'' (no \emph{rain} prompts), we also obtain similar improvement for \emph{rain}.
Detailed results are reported in~\cref{tbl:all_domains}.
That experiment hints at the generalization potential of our framework.
We list here all the domains we used to form the ``all-domains'' calibration set: Beijing, Cairo, clouds, Dubai, fall, fog, hurricane, India, Istanbul, Johannesburg, lightning, London, Moscow, Mumbai, night, Paris, rain, sandstorm, snow, spring, summer, sun, Sydney, Tokyo, tornado, Toronto, wind, winter. This calibration set is comprised of 64 images per shift.

\begin{table}[h!]
    \centering
    \begin{tabular}{l|ccccc}
        \toprule
        \textbf{Prompt} & India & Fog & Rain & Snow & Night \\
        \midrule
        domain-specific & 72.5 & 92.5 & 100 & 95 & 90 \\
        all domains& 78.6 & 100 & 100 & 100 & 100 \\
        all domains but-rain & - & - & 100 & - & - \\
        \bottomrule
    \end{tabular}
    \caption{\small \textbf{Generalization.} We experiment the effects of prompting and showcase the generalization potentials of our framework. Here ECE improvements are reported (cf. Fig.6).}
    \label{tbl:all_domains}
\end{table}

\section{Direct Data Correlation}
\label{sec:fid_score}
We provide in \cref{tbl:fid_scores} the FID scores($\downarrow$) for direct correlation between synthetic and real data.
Our zero-shot approach outperforms all. Layout differences between Cityscapes and ACDC cause high FIDs -- one critical limitation of FID on structured data.
As our work focuses on performance testing, we prioritize metrics like mIoU and FPR$_{95}$, aligned with recent works.
\begin{table}[h!]
	\centering
	\resizebox{\linewidth}{!}{%
		\begin{tabular}{l|cc|cccc}
			\toprule
			& OOD expertise? & OOD data? & \textbf{Night} & \textbf{Rain} & \textbf{Snow} & \textbf{Fog}\\
			\midrule
			Cityscapes (no aug.) & \textcolor{plt:green}{no} & \textcolor{plt:green}{no} & 236.5 & 188.3 & 194.8 & 184.8 \\
			Hendrycks' aug.~\cite{hendrycks2019benchmarking} & \textcolor{plt:red}{required} & \textcolor{plt:green}{no} & - & - & 210.7 & 191.1 \\
			GAN-based TSIT~\cite{jiang2020tsit} & \textcolor{plt:green}{no} & \textcolor{plt:red}{required} & 254.2 & 223.2 & 225.6 & - \\
			Physics-based Fog Sim.~\cite{sakaridis2018semantic} & \textcolor{plt:red}{required} & \textcolor{plt:green}{no} & - & - & - & 182.8 \\
			Ours w/ SD1.5 & \textcolor{plt:green}{no} & \textcolor{plt:green}{no} & \textbf{180.2} & \textbf{177.5} & \textbf{164.5} & \textbf{163.3} \\
			\bottomrule
		\end{tabular}%
	}%
	\caption{\small \textbf{Direct Data Correlation.} Our pipeline achieves better FID scores while does not require any OOD knowledge.}
	\label{tbl:fid_scores}
\end{table}

Reflected in the FID study, in-domain data (CS) serves as a strong baseline for measuring OOD performance~\cite{de2023reliability,miller2021accuracy}.
Hendrycks' augmentations are unrealistic and inadequate for assessing real OOD performance, as previously revealed in~\cite{taori2020measuring}. Using untouched CS data could be even better; for example, in terms of PCC score for `fog' (cf. Fig.3 and Tab.1), Hendrycks's data obtains 0.70 as compared to 0.78 of CS data and 0.89 of our SDXL data.
Our work, consistent with~\cite{taori2020measuring}, demonstrates that simple augmentations are insufficient for OOD testing, advocating for more realistic testing data.
While showing the zero-shot advantage over standard domain-specific augmentations, we acknowledge that a well crafted augmentation approach like physics-based fog augmentation~\cite{sakaridis2018semantic} can obtain very good results, potentially better than ours if improved. Of note, \emph{we do not aim to obtain best results in all cases; we advocate for a generic zero-shot testing framework as the starting reference in arbitrary domains.}

\section{Class-wise analysis.}
\label{sec:class_wise}
We conduct a class-wise analysis and report the PCC scores per class in~\cref{tbl:class_wise}. Interestingly, we notice that `bicycle' and `bus' have the least correlation for \emph{india} and \emph{night}, respectively, which actually corresponds to the low occurrences of those classes in such conditions. High correlation indicates that either the corresponding classes are easy (`building', `person', `car', or `truck') or those classes are difficult ('sign' or 'pole') and make segmenters struggle either on real or synthetic data.
We conjecture that using synthetic data may provide us with hints about the inherent bias of the pretrained models.

\begin{table}[h!]
    \centering
    \begin{tabular}{l|ccccc}
        \toprule
        Class & Fog & Night & Rain & Snow & India \\
        \midrule
        road & 0.54 & 0.69 & 0.57 & 0.60 & 0.74 \\
        sidewalk & 0.62 & 0.62 & 0.68 & 0.64 & 0.69 \\
        building & 0.67 & \textcolor{plt:green}{0.82} & 0.76 & 0.57 & 0.84 \\
        wall & \textcolor{plt:red}{0.03} & 0.76 & 0.60 & 0.71 & 0.58 \\
        fence & 0.21 & 0.66 & 0.71 & 0.64 & 0.38 \\
        pole & 0.33 & 0.75 & 0.60 & 0.29 & 0.47 \\
        light & 0.36 & 0.53 & 0.63 & 0.70 & 0.30 \\
        sign & 0.33 & \textcolor{plt:green}{0.82} & 0.47 & 0.55 & 0.26 \\
        vegetation & 0.57 & 0.71 & \textcolor{plt:green}{0.77} & 0.52 & 0.00 \\
        terrain & 0.31 & 0.55 & 0.62 & 0.76 & - \\
        sky & 0.70 & 0.44 & \textcolor{plt:red}{0.07} & \textcolor{plt:red}{0.17} & 0.30 \\
        person & 0.56 & 0.72 & 0.52 & 0.69 & \textcolor{plt:green}{0.86} \\
        rider & 0.27 & 0.55 & 0.43 & 0.31 & 0.16 \\
        car & 0.73 & 0.71 & 0.72 & \textcolor{plt:green}{0.78} & 0.61 \\
        truck & \textcolor{plt:green}{0.83} & 0.67 & 0.74 & 0.75 & 0.72 \\
        bus & 0.67 & \textcolor{plt:red}{0.32} & 0.57 & 0.69 & 0.19 \\
        train & 0.35 & 0.65 & 0.35 & 0.41 & - \\
        motocycle & 0.25 & 0.39 & 0.37 & 0.62 & 0.40 \\
        bicycle & 0.49 & 0.64 & 0.73 & 0.72 & \textcolor{plt:red}{-0.16} \\
        \bottomrule
    \end{tabular}
    \caption{\small \textbf{Class-wise analysis.} We provide the PCC scores per class for each shift. The \textcolor{plt:green}{most} and \textcolor{plt:red}{least} scores are colored.}
    \label{tbl:class_wise}
\end{table}

\section{Limitations}
\label{sec:limitations}
We focus our research on the task of semantic segmentation while keeping open the extension possibility to other critical tasks, such as object detection.
Our quantitative assessments are confined to existing publicly available datasets. However, our framework is fully zero-shot and can be applied to any domain of interest. 
On the generative side, our study is restricted to Stable Diffusion and ControlNet due to our resource constraints.
Of note, although improvements in this area should enhance the results, similar insights are expected to be achieved, as primarily shown with SDXL results. Another limitation is our primary focus on autonomous driving (AD) data. That is because the field of AD released diverse datasets to stress-test models across various OOD scenarios. Unlike generalist datasets like MS-COCO, they clearly distinguish between domains, enabling the covariate shift studies in this work.

\smallskip\noindent\textbf{Robustness Assessment.} In order to accommodate the {in-domain} GTs, we make the semantic preservation assumption, similar to the one in domain adaptation.
We realize that this common assumption might raise questions when an element like snow is added; nevertheless, the issue is complex and hinges on the annotation policies.
The ACDC dataset~\cite{sakaridis2021acdc}, for instance, uses clean images to inform the annotators of what is behind the snow, and our approach mimics this.
But they are also very conservative about ambiguity, with an explicit label.
Moving forward, we could take inspiration and try to handle ambiguity.

\section{Qualitative Examples}
\label{sec:supp_qual}
\begin{figure*}[ht!]
	\setlength{\tabcolsep}{0.002\linewidth}
    \setlength{\arrayrulewidth}{2pt}
	\centering
        \scriptsize
	\begin{tabular}{ccccc|cccc}
        &Syn. Img& Sem.FPN & MobileV3 & SegF-B5 & Syn. Img& Sem.FPN & MobileV3 & SegF-B5 \\
        \rotatebox{90}{Sun glare}&\includegraphics[width=0.11\linewidth]{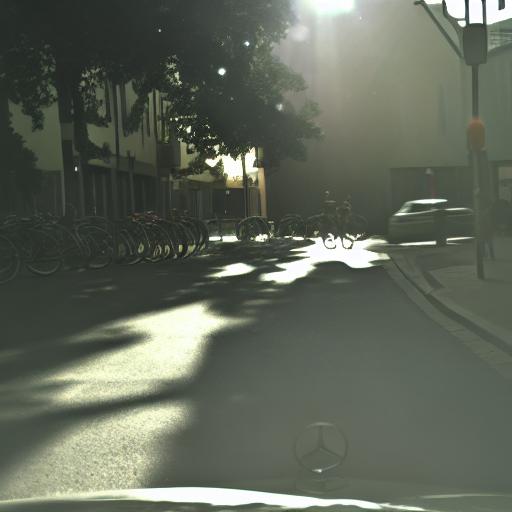} & \includegraphics[width=0.11\linewidth]{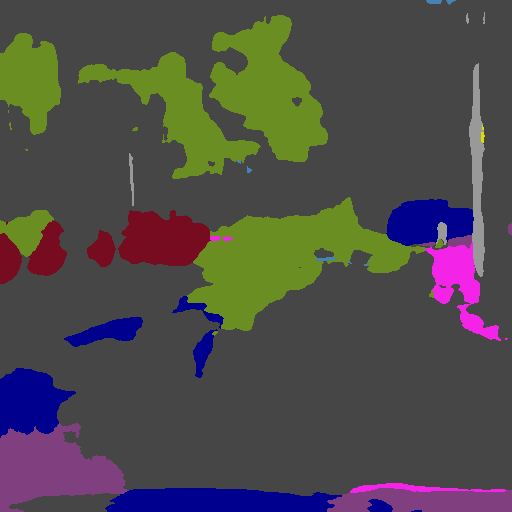} & \includegraphics[width=0.11\linewidth]{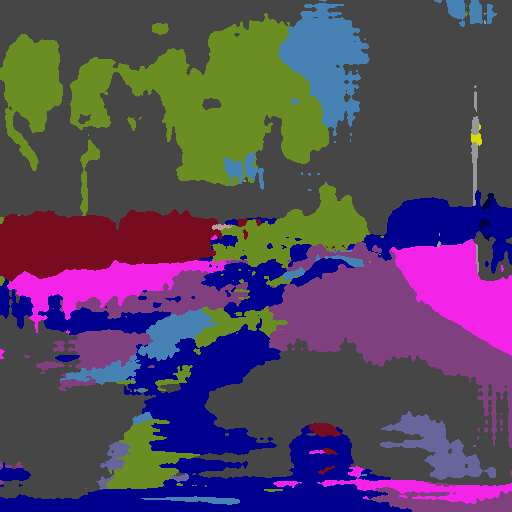} & \includegraphics[width=0.11\linewidth]{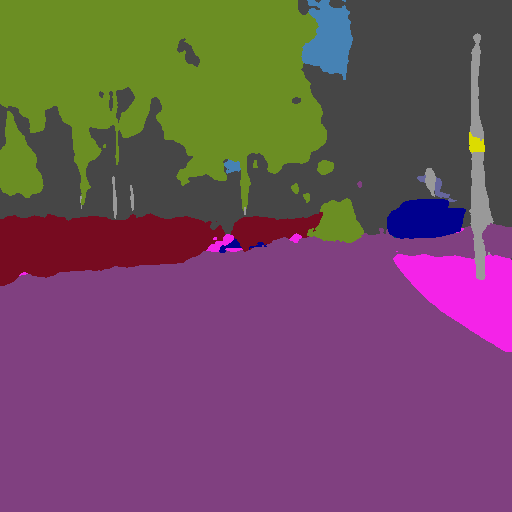} & \includegraphics[width=0.11\linewidth]{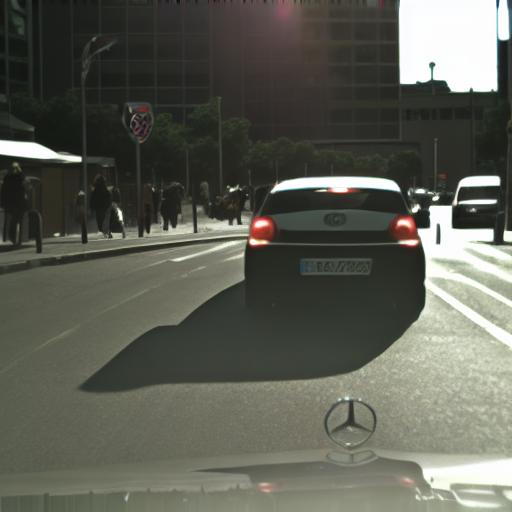} & \includegraphics[width=0.11\linewidth]{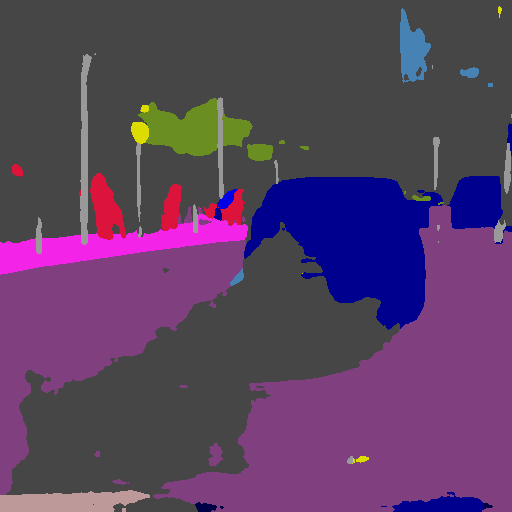} & \includegraphics[width=0.11\linewidth]{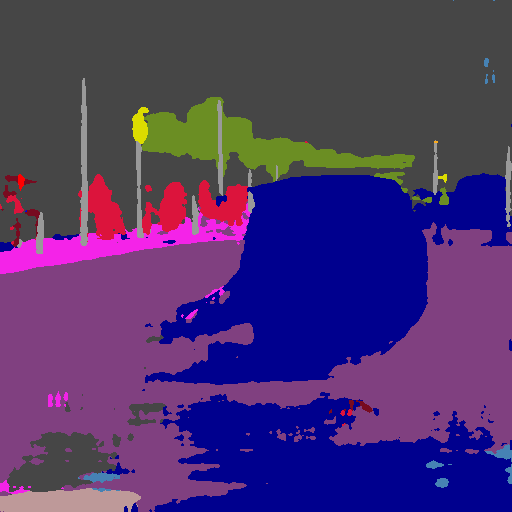} & \includegraphics[width=0.11\linewidth]{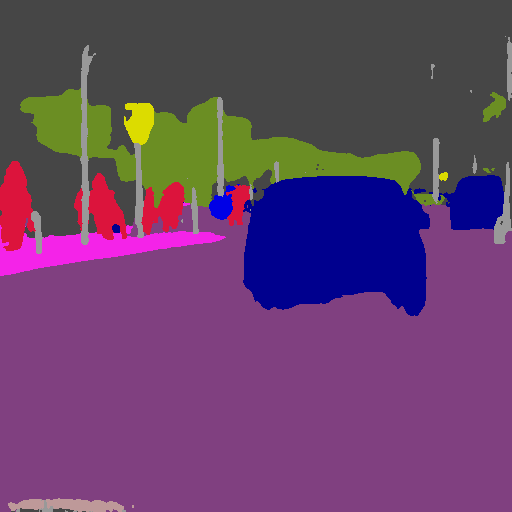} \\
        \rotatebox{90}{Flood}&\includegraphics[width=0.11\linewidth]{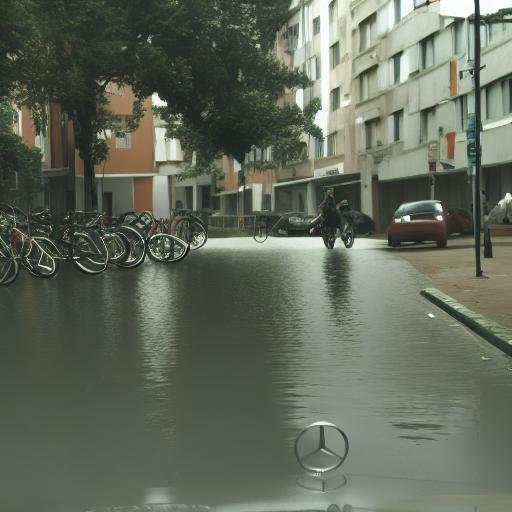} & \includegraphics[width=0.11\linewidth]{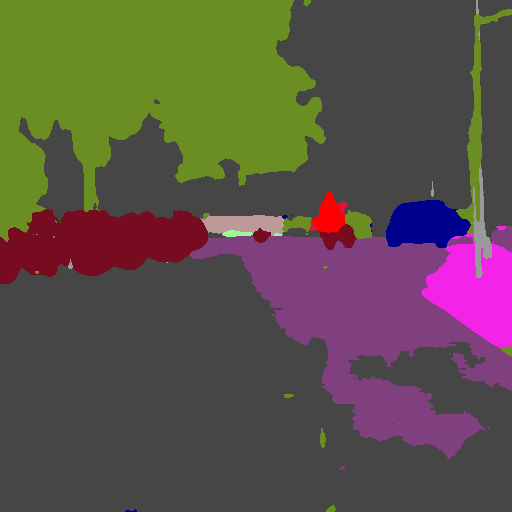} & \includegraphics[width=0.11\linewidth]{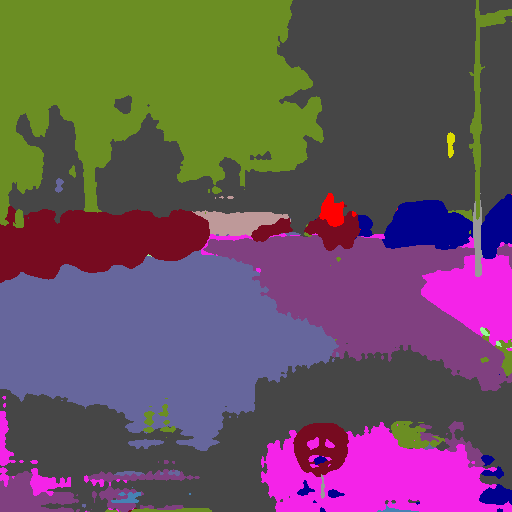} & \includegraphics[width=0.11\linewidth]{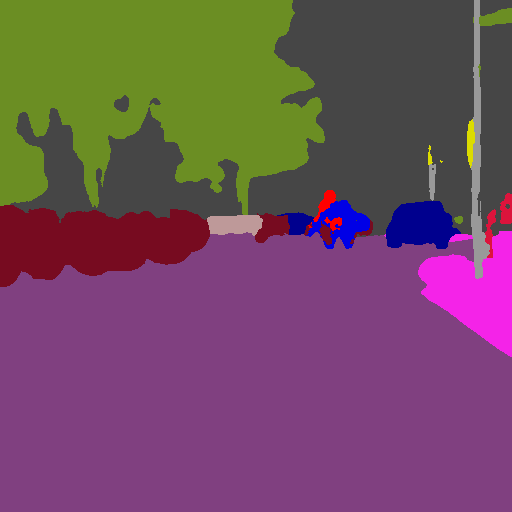}  & \includegraphics[width=0.11\linewidth]{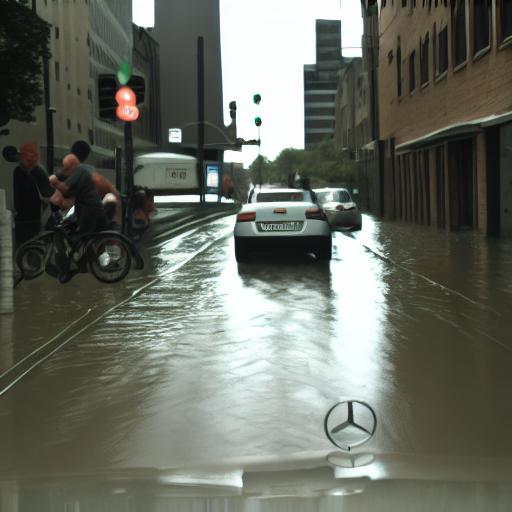} & \includegraphics[width=0.11\linewidth]{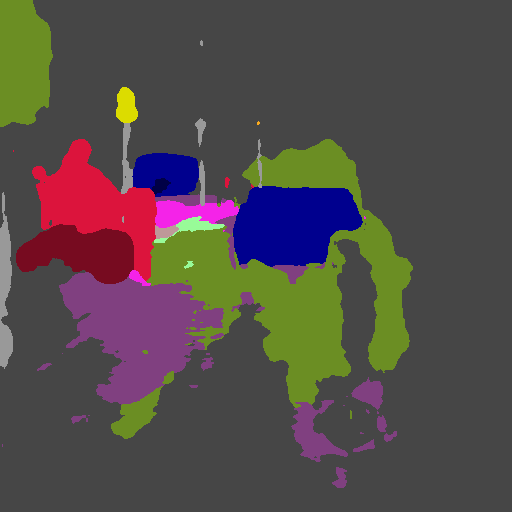} & \includegraphics[width=0.11\linewidth]{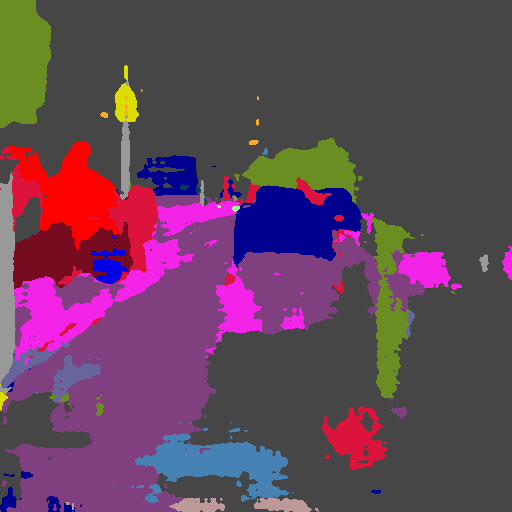} & \includegraphics[width=0.11\linewidth]{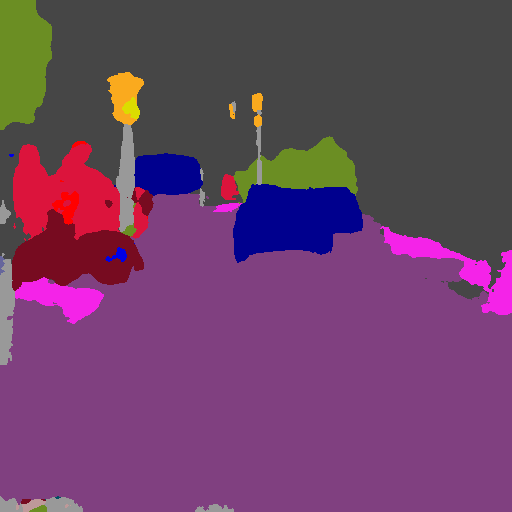} \\
        \rotatebox{90}{Autumn}&\includegraphics[width=0.11\linewidth]{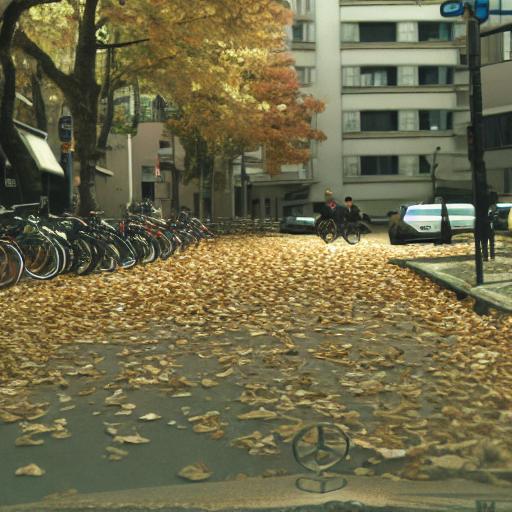} & \includegraphics[width=0.11\linewidth]{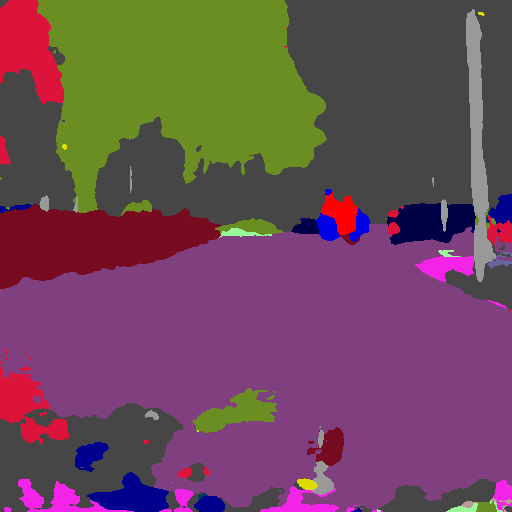} & \includegraphics[width=0.11\linewidth]{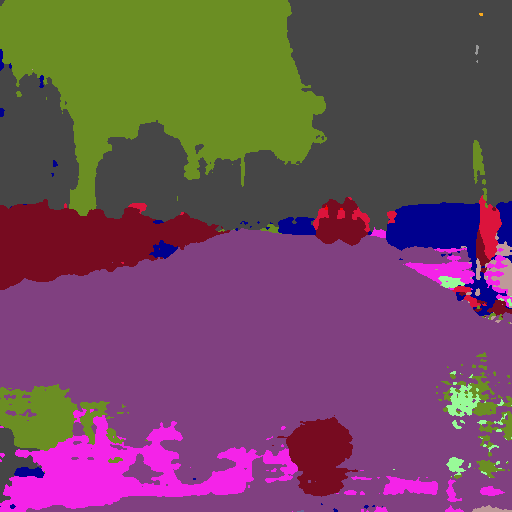} & \includegraphics[width=0.11\linewidth]{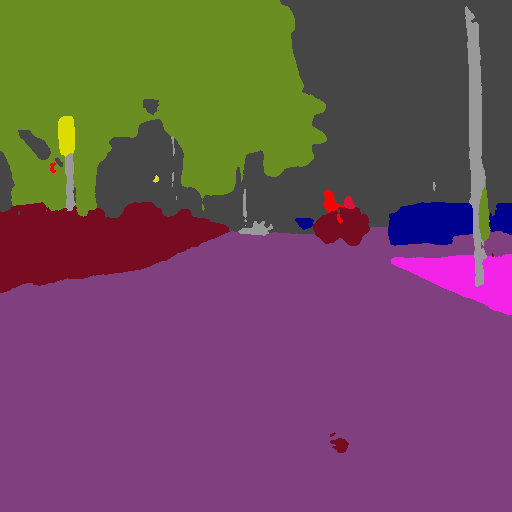} & \includegraphics[width=0.11\linewidth]{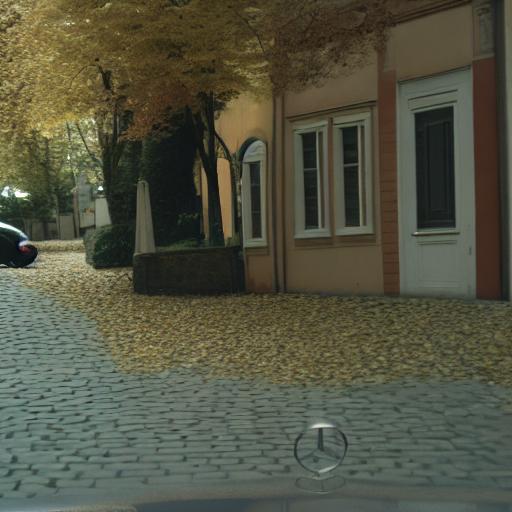} & \includegraphics[width=0.11\linewidth]{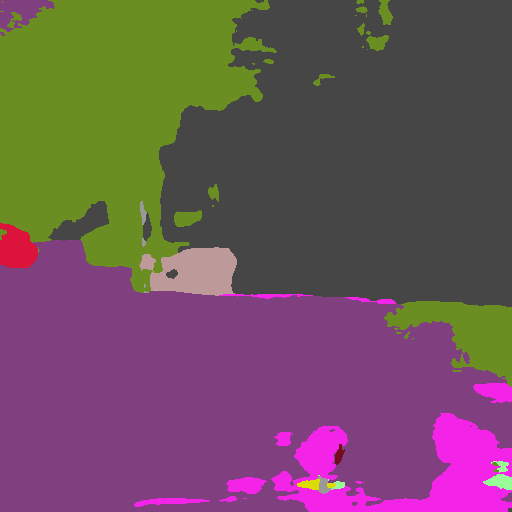} & \includegraphics[width=0.11\linewidth]{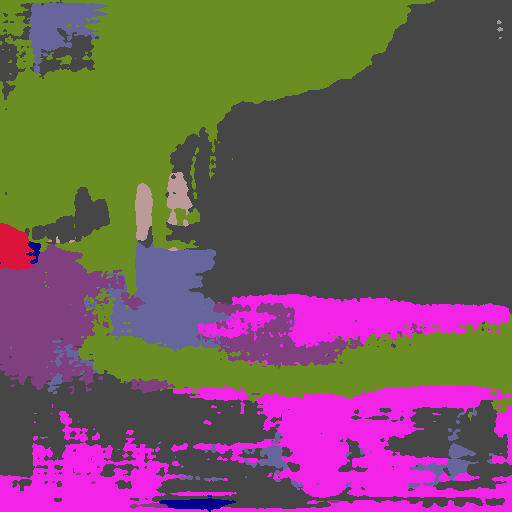} & \includegraphics[width=0.11\linewidth]{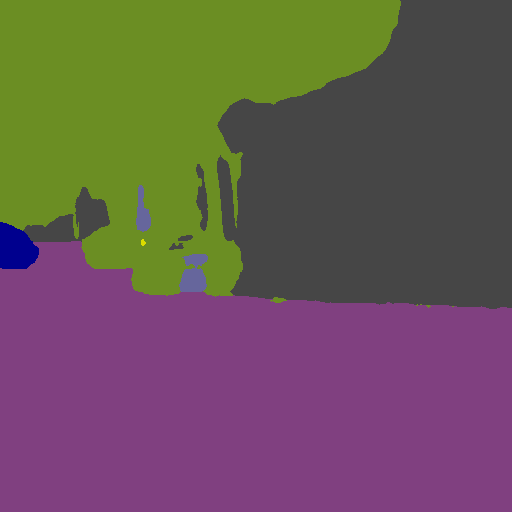} \\
        \rotatebox{90}{Fire}&\includegraphics[width=0.11\linewidth]{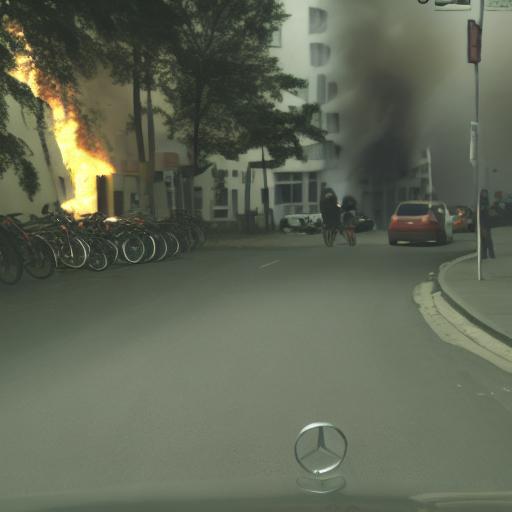} & \includegraphics[width=0.11\linewidth]{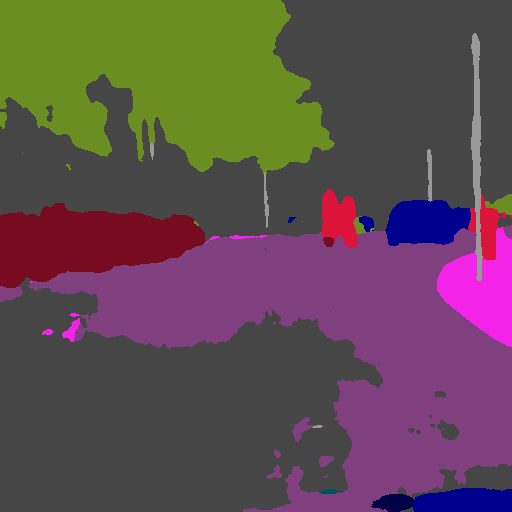} & \includegraphics[width=0.11\linewidth]{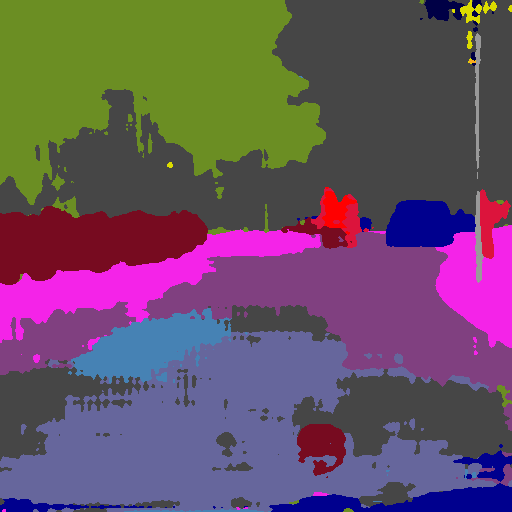} & \includegraphics[width=0.11\linewidth]{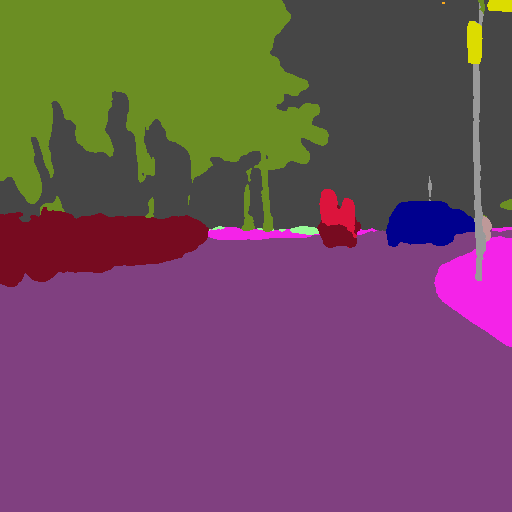} & \includegraphics[width=0.11\linewidth]{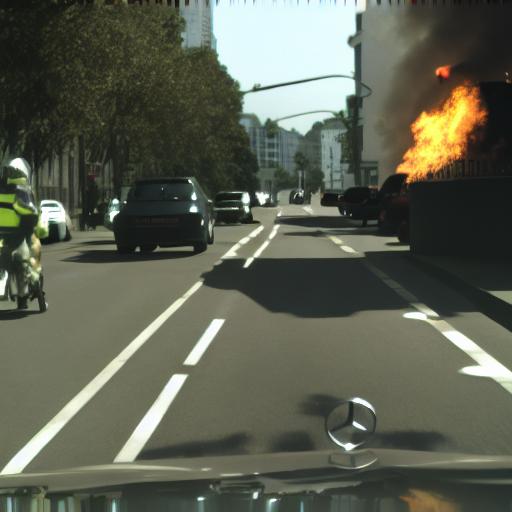} & \includegraphics[width=0.11\linewidth]{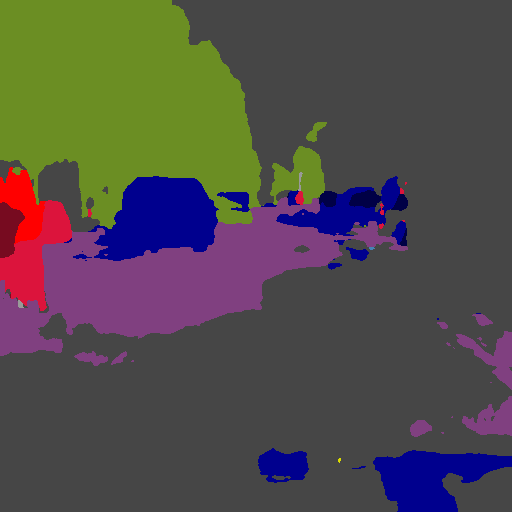} & \includegraphics[width=0.11\linewidth]{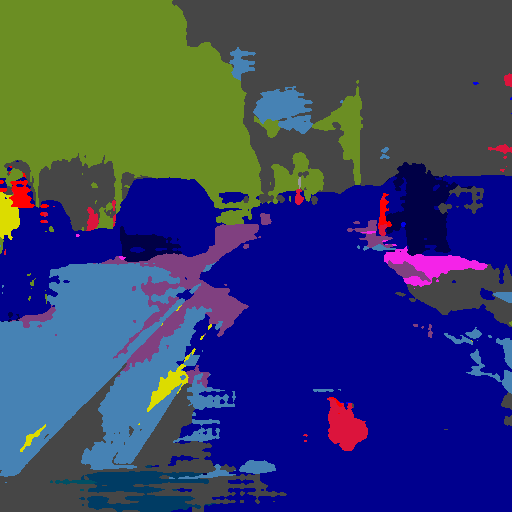} & \includegraphics[width=0.11\linewidth]{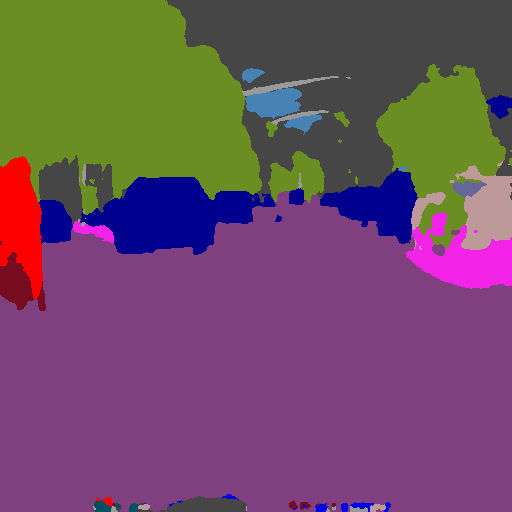} \\
        \rotatebox{90}{Fog}&\includegraphics[width=0.11\linewidth]{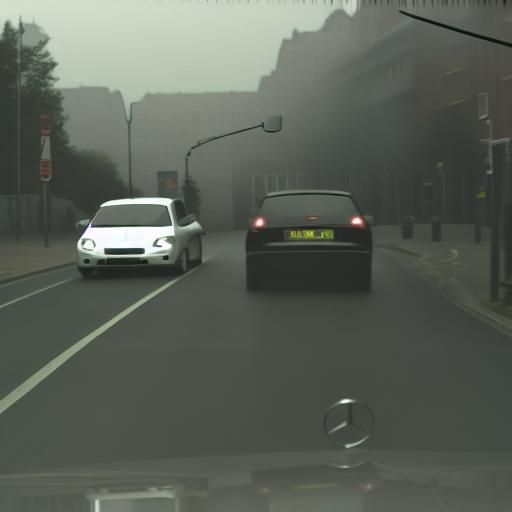} & \includegraphics[width=0.11\linewidth]{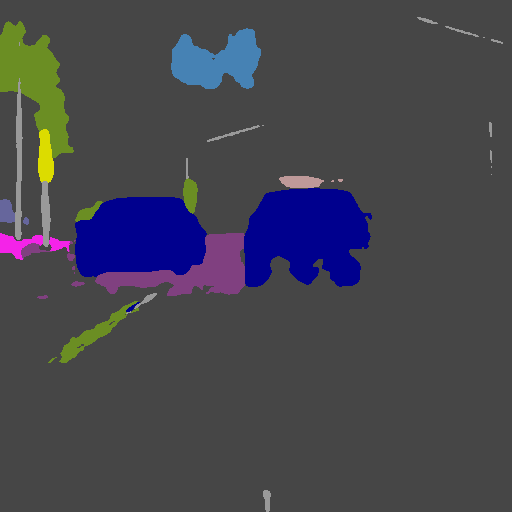} & \includegraphics[width=0.11\linewidth]{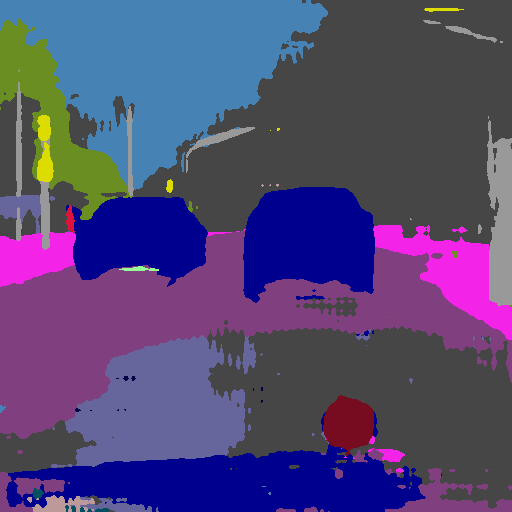} & \includegraphics[width=0.11\linewidth]{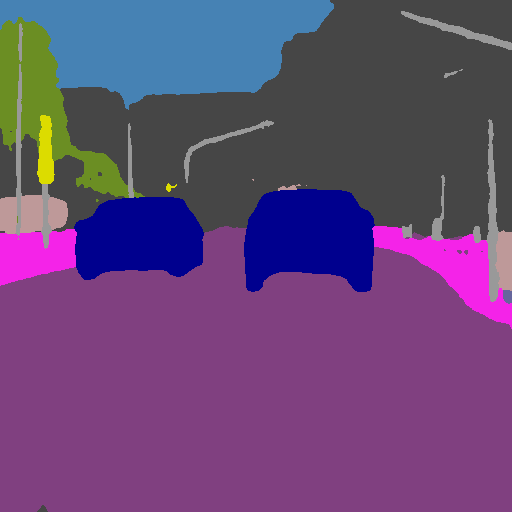} & \includegraphics[width=0.11\linewidth]{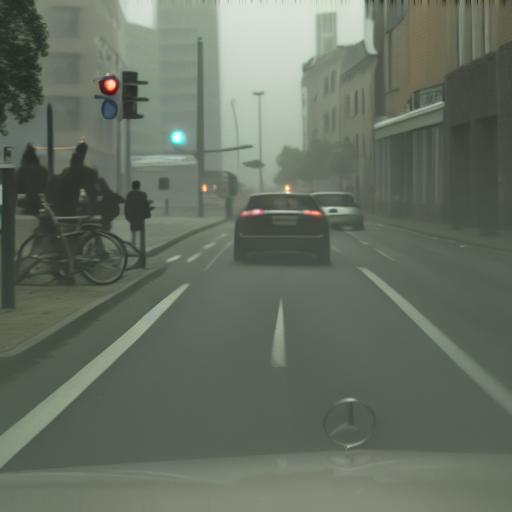} & \includegraphics[width=0.11\linewidth]{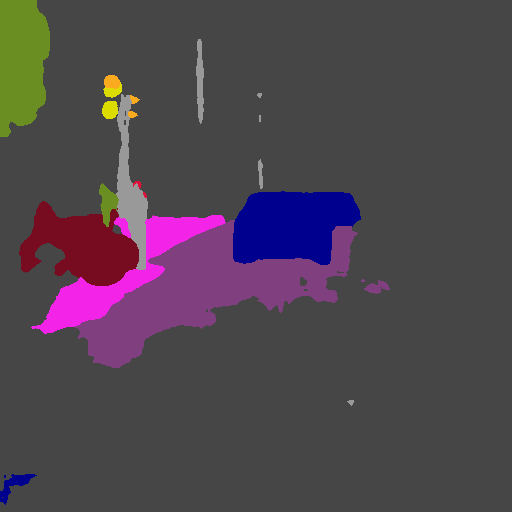} & \includegraphics[width=0.11\linewidth]{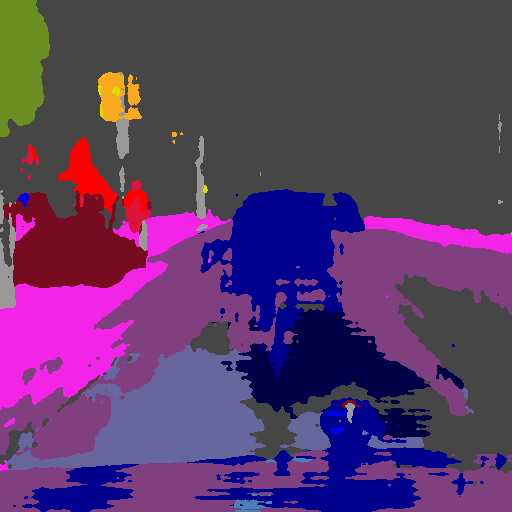} & \includegraphics[width=0.11\linewidth]{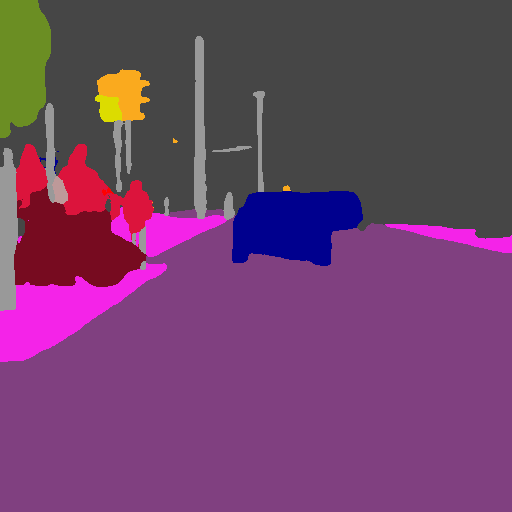} \\
        \rotatebox{90}{Rain}&\includegraphics[width=0.11\linewidth]{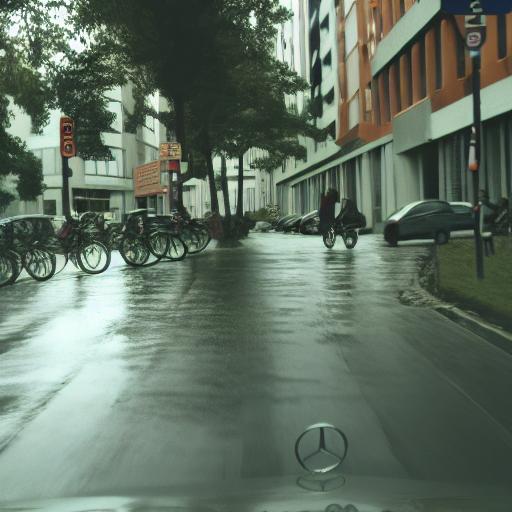} & \includegraphics[width=0.11\linewidth]{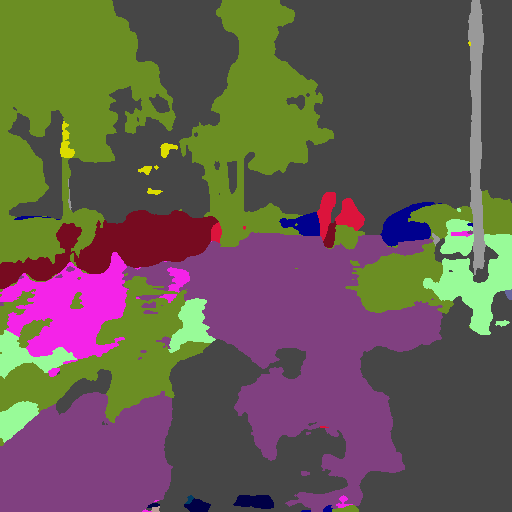} & \includegraphics[width=0.11\linewidth]{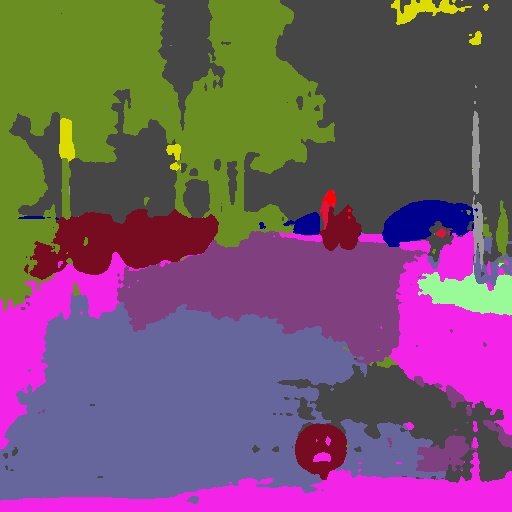} & \includegraphics[width=0.11\linewidth]{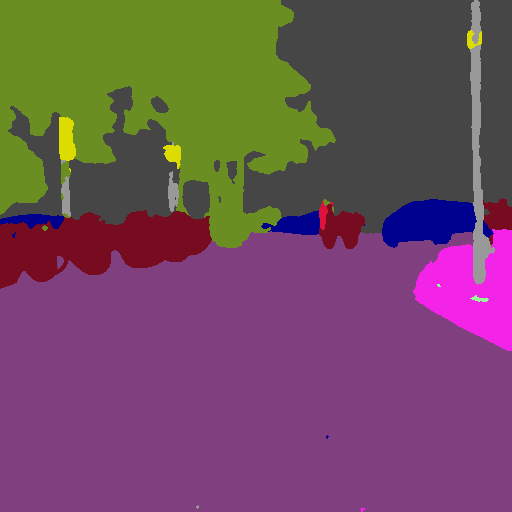} & \includegraphics[width=0.11\linewidth]{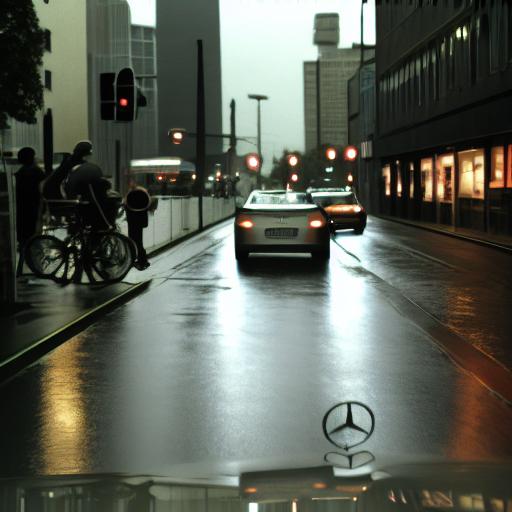} & \includegraphics[width=0.11\linewidth]{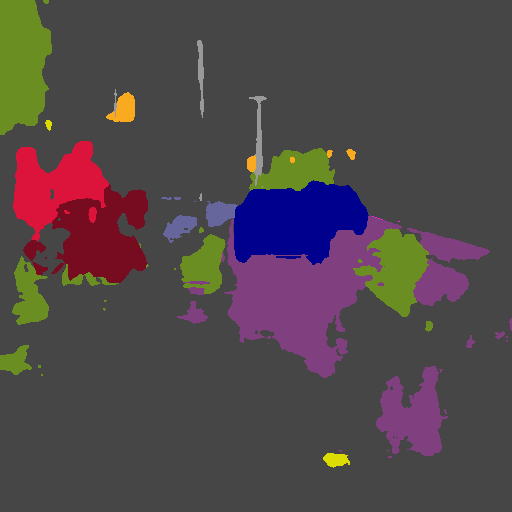} & \includegraphics[width=0.11\linewidth]{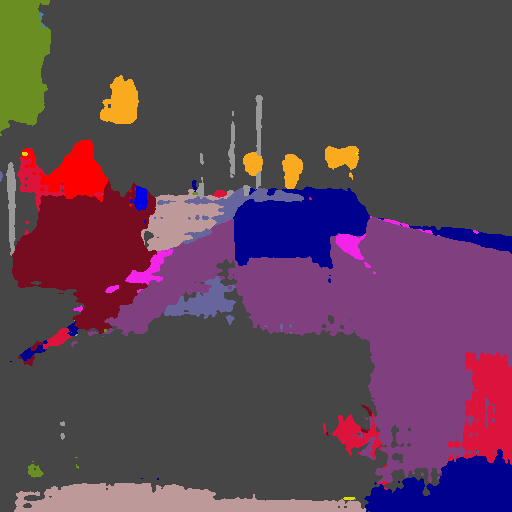} & \includegraphics[width=0.11\linewidth]{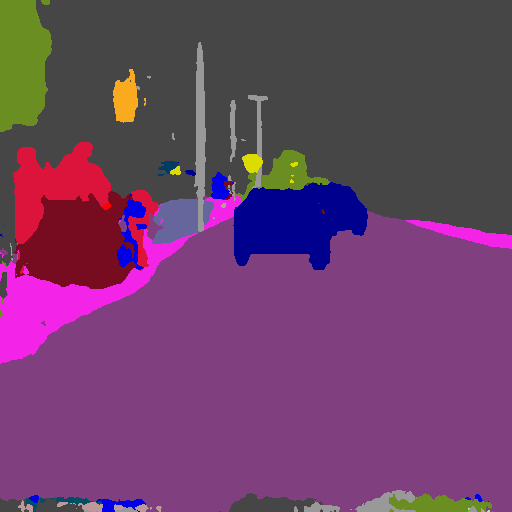} \\
        \rotatebox{90}{Snow}&\includegraphics[width=0.11\linewidth]{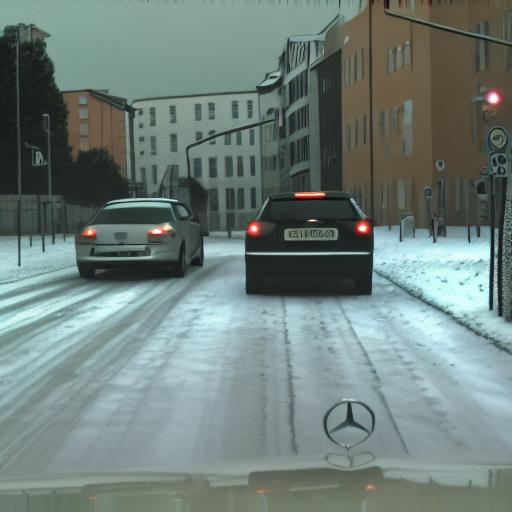} & \includegraphics[width=0.11\linewidth]{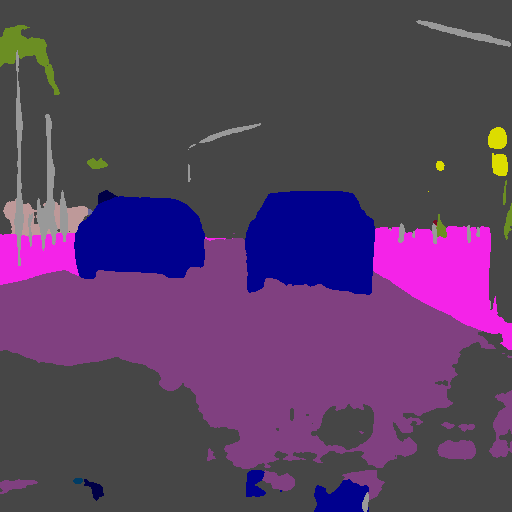} & \includegraphics[width=0.11\linewidth]{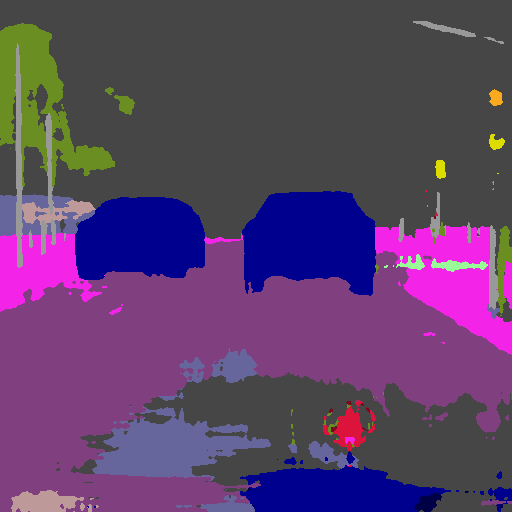} & \includegraphics[width=0.11\linewidth]{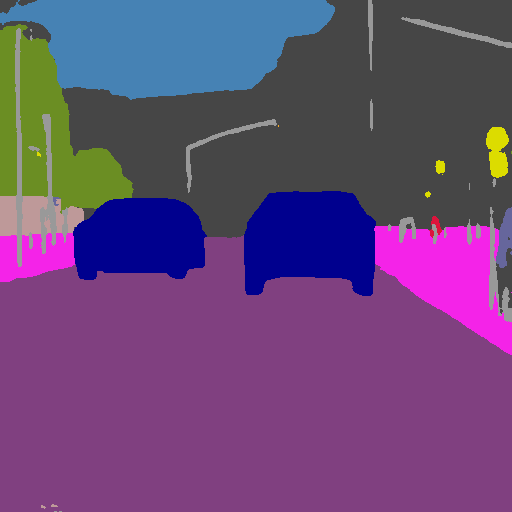} & \includegraphics[width=0.11\linewidth]{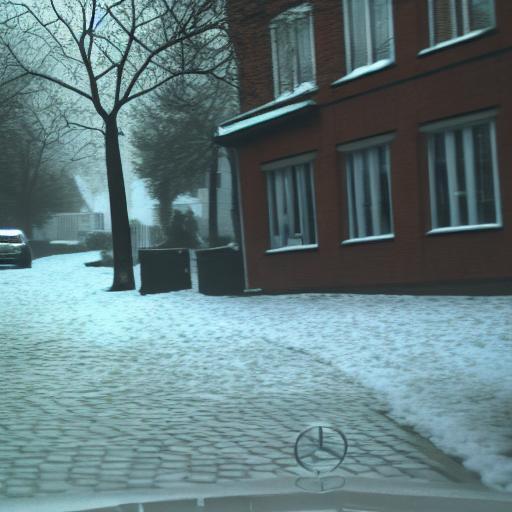} & \includegraphics[width=0.11\linewidth]{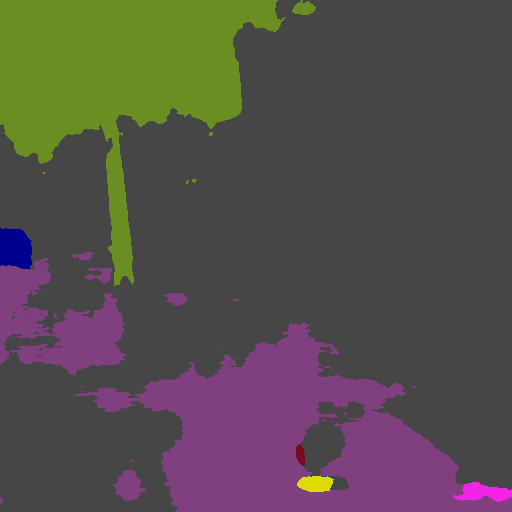} & \includegraphics[width=0.11\linewidth]{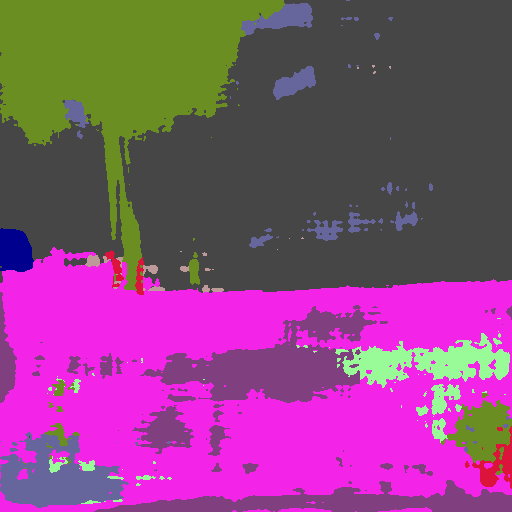} & \includegraphics[width=0.11\linewidth]{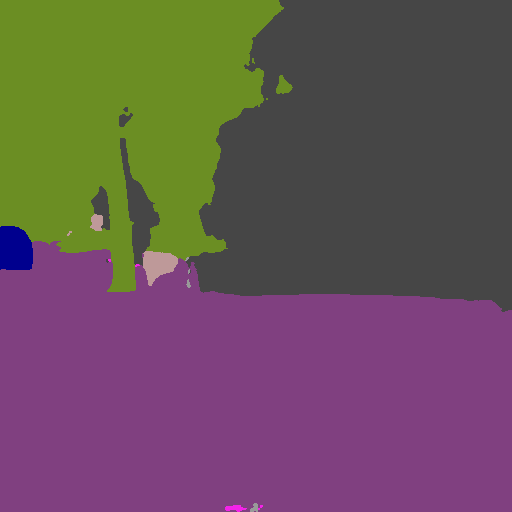} \\
        \rotatebox{90}{Night}&\includegraphics[width=0.11\linewidth]{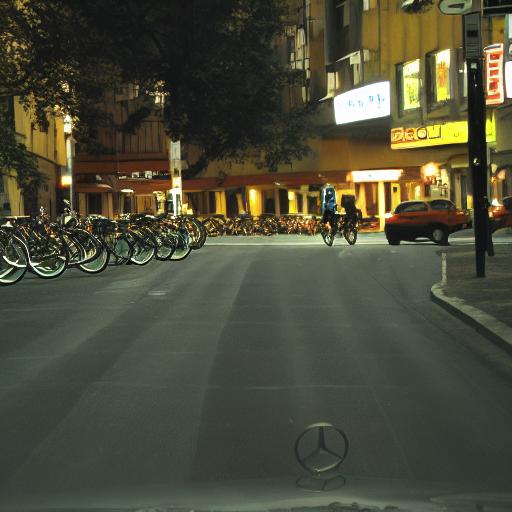} & \includegraphics[width=0.11\linewidth]{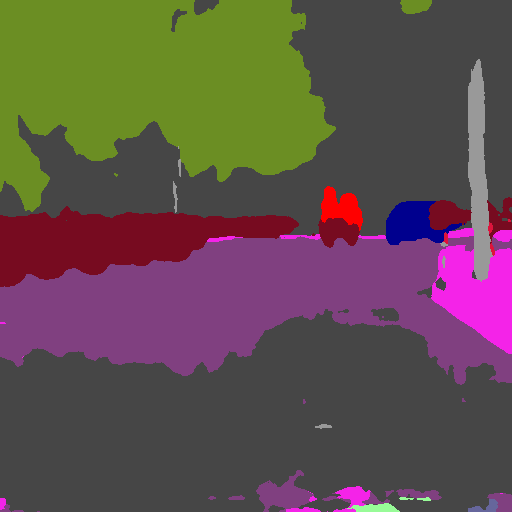} & \includegraphics[width=0.11\linewidth]{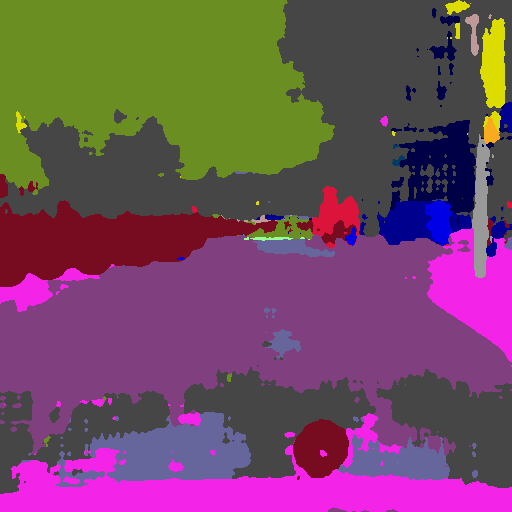} & \includegraphics[width=0.11\linewidth]{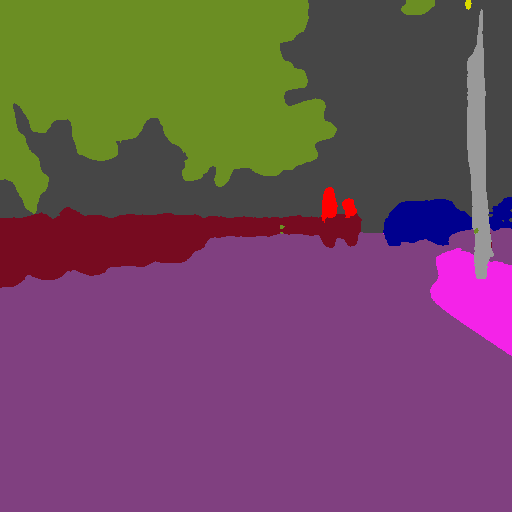} & \includegraphics[width=0.11\linewidth]{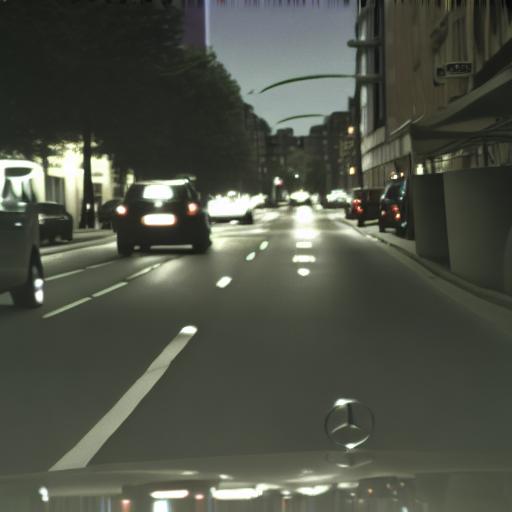} & \includegraphics[width=0.11\linewidth]{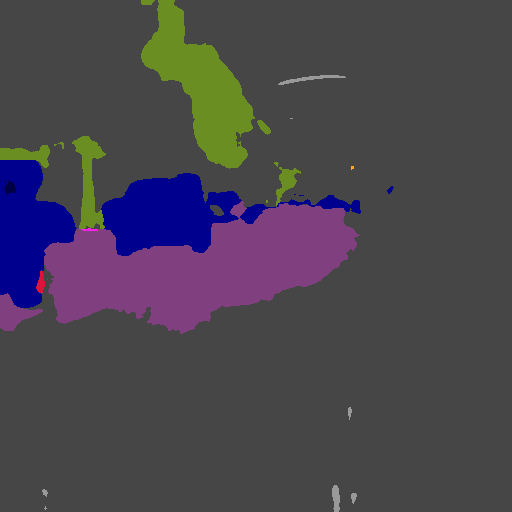} & \includegraphics[width=0.11\linewidth]{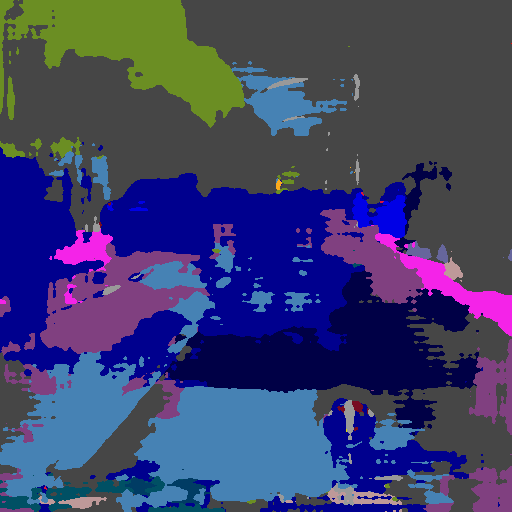} & \includegraphics[width=0.11\linewidth]{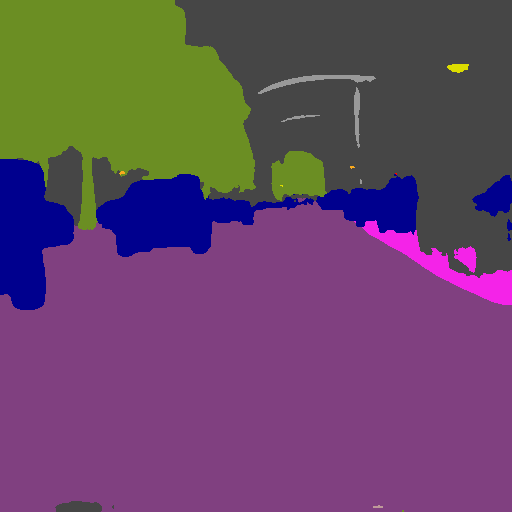} \\
        
	\end{tabular}
    \smallskip
    \caption{\textbf{Qualitative results.} Examples of rare conditions generated for testing and predictions from different models. Results of the strong model like SegFormer-B5 is visibly better than the Semantic-FPN and MobileNetV3.}
    \label{fig:qualres_sec3_suppmat}
\end{figure*}

\begin{figure*}[ht!]
	\setlength{\tabcolsep}{0.002\linewidth}
    \setlength{\fboxsep}{0pt}
    \setlength{\fboxrule}{1.5pt}
	\centering
    \small
	\begin{tabular}{ccccc}
        & \multicolumn{4}{c}{
        \begin{tikzpicture}
            \draw[->,>=latex, line width=2.5pt] (-6.6,0) -- (2,0) node[midway,fill=white] {\textbf{Better Models in AUROC on Synthetic}};
            \end{tikzpicture}
        } \\
        Images & Semantic-FPN & MobileNetV3 & SegFormer-B5 & SETR-PUP \\
        \includegraphics[width=0.18\textwidth]{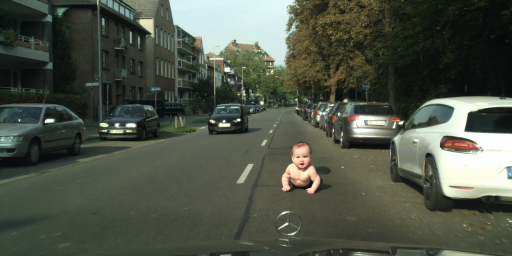}
        & \includegraphics[width=0.18\textwidth]{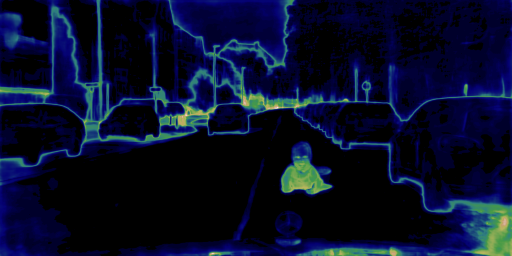}
        & \includegraphics[width=0.18\textwidth]{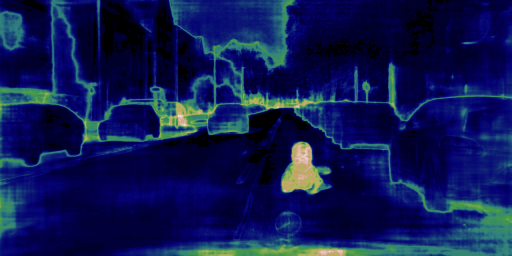}
        & \includegraphics[width=0.18\textwidth]{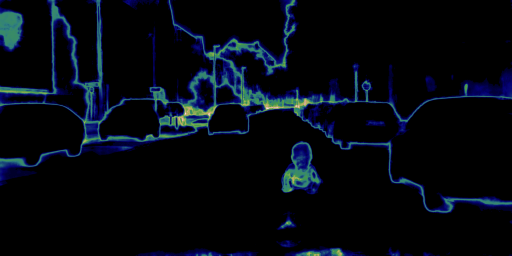}
        & \includegraphics[width=0.18\textwidth]{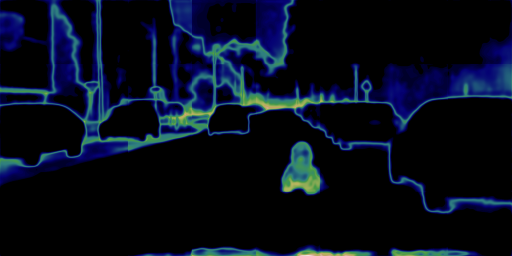} \\
        \includegraphics[width=0.18\textwidth]{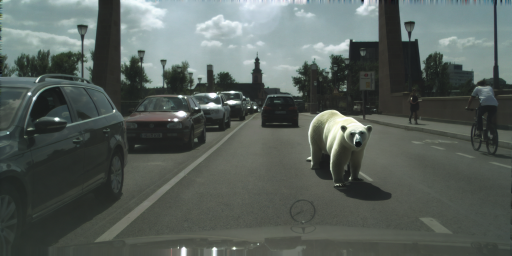}
        & \includegraphics[width=0.18\textwidth]{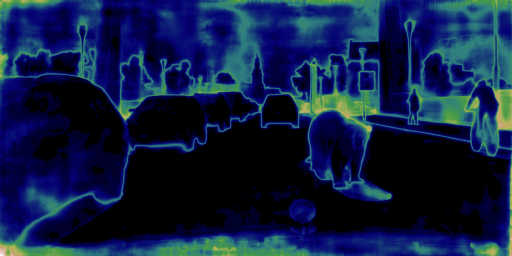}
        & \includegraphics[width=0.18\textwidth]{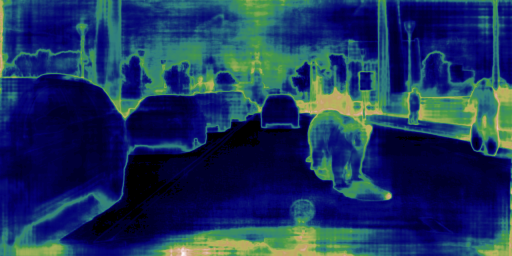}
        & \includegraphics[width=0.18\textwidth]{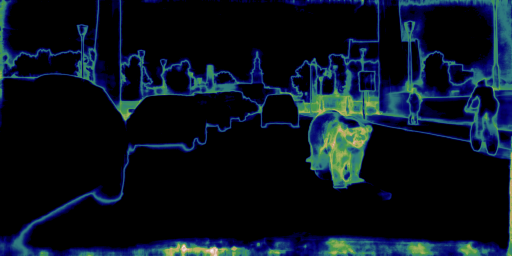}
        & \includegraphics[width=0.18\textwidth]{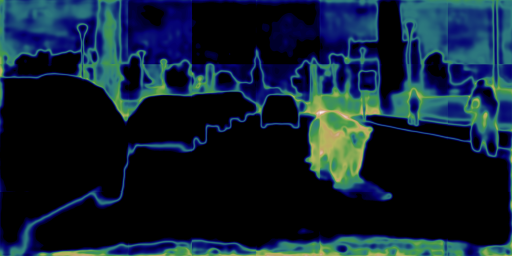} \\
        \includegraphics[width=0.18\textwidth]{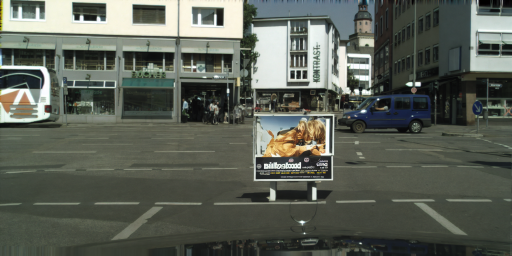}
        & \includegraphics[width=0.18\textwidth]{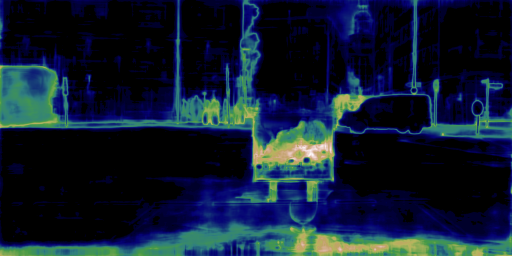}
        & \includegraphics[width=0.18\textwidth]{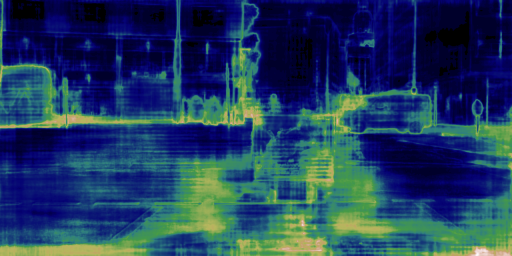}
        & \includegraphics[width=0.18\textwidth]{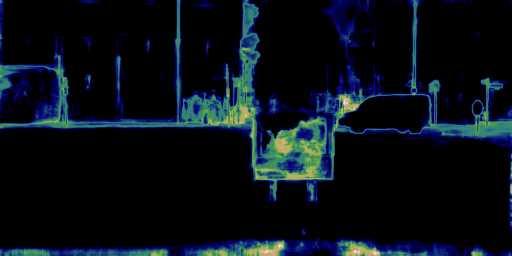}
        & \includegraphics[width=0.18\textwidth]{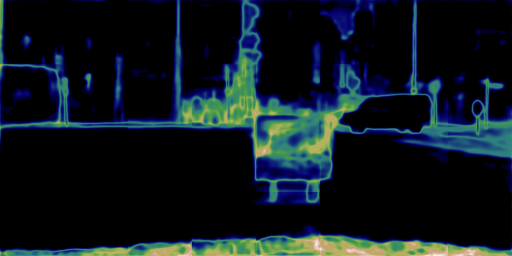} \\
        \includegraphics[width=0.18\textwidth]{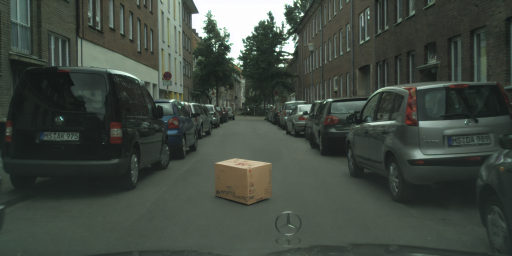}
        & \includegraphics[width=0.18\textwidth]{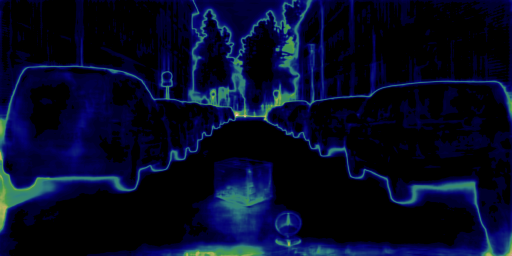}
        & \includegraphics[width=0.18\textwidth]{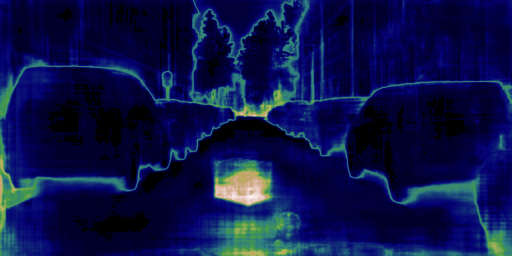}
        & \includegraphics[width=0.18\textwidth]{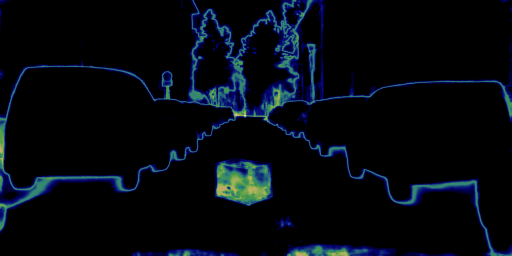}
        & \includegraphics[width=0.18\textwidth]{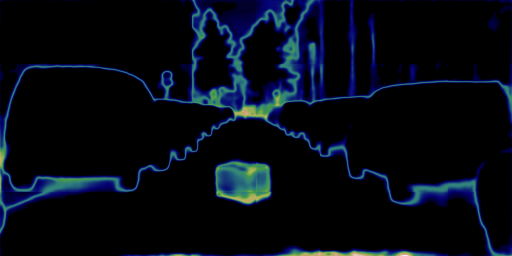} \\
        \includegraphics[width=0.18\textwidth]{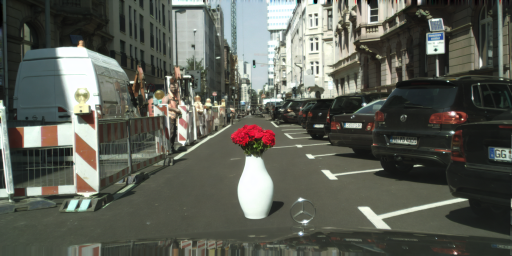}
        & \includegraphics[width=0.18\textwidth]{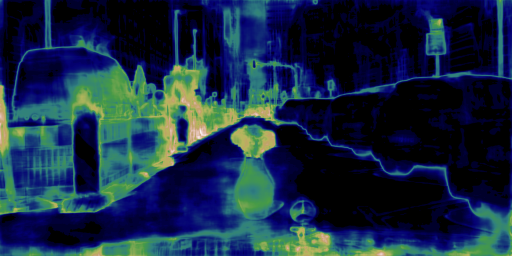}
        & \includegraphics[width=0.18\textwidth]{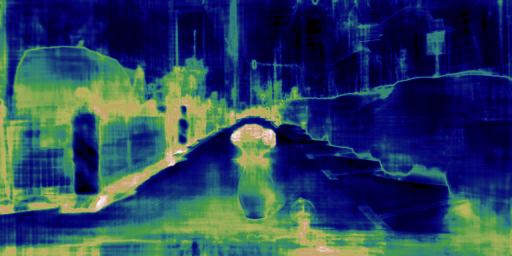}
        & \includegraphics[width=0.18\textwidth]{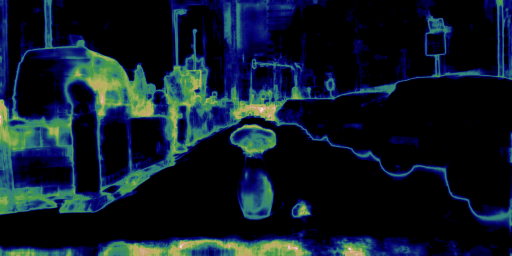}
        & \includegraphics[width=0.18\textwidth]{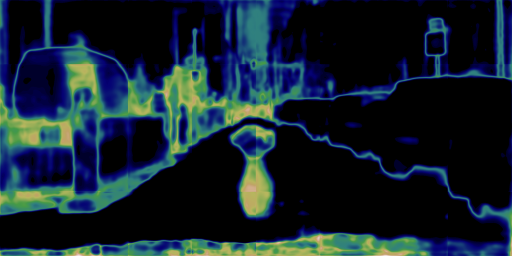} \\
        \includegraphics[width=0.18\textwidth]{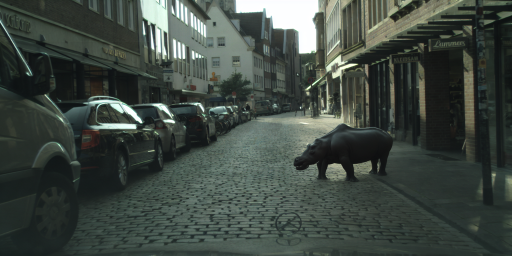}
        & \includegraphics[width=0.18\textwidth]{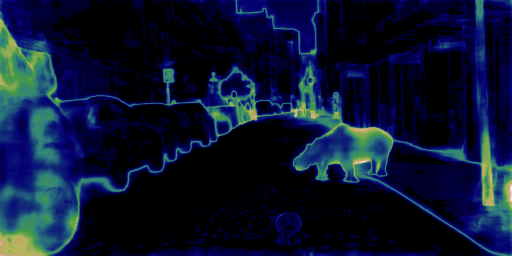}
        & \includegraphics[width=0.18\textwidth]{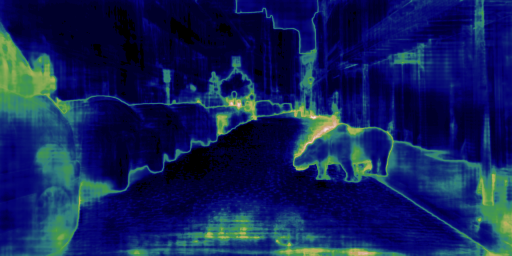}
        & \includegraphics[width=0.18\textwidth]{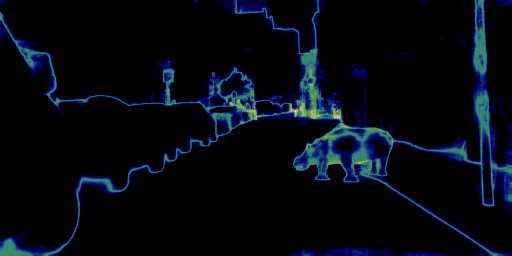}
        & \includegraphics[width=0.18\textwidth]{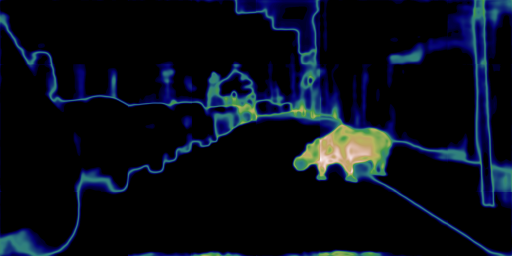} \\
        \includegraphics[width=0.18\textwidth]{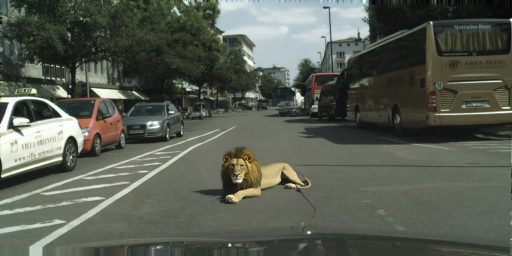}
        & \includegraphics[width=0.18\textwidth]{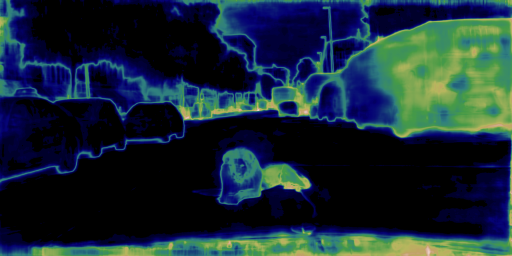}
        & \includegraphics[width=0.18\textwidth]{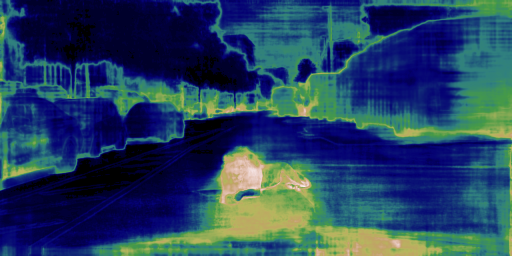}
        & \includegraphics[width=0.18\textwidth]{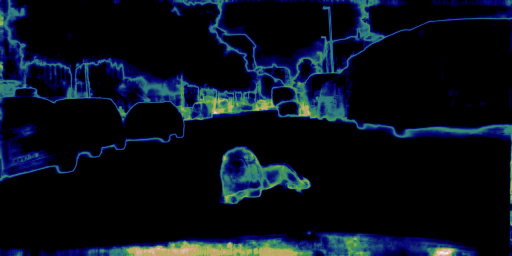}
        & \includegraphics[width=0.18\textwidth]{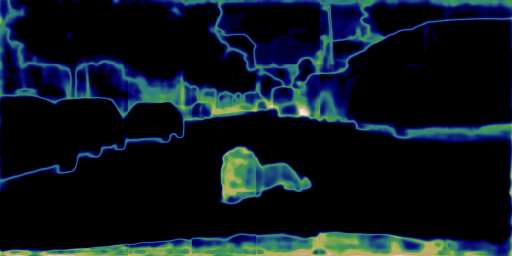} \\
        \includegraphics[width=0.18\textwidth]{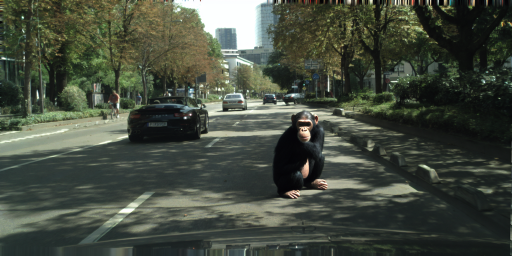}
        & \includegraphics[width=0.18\textwidth]{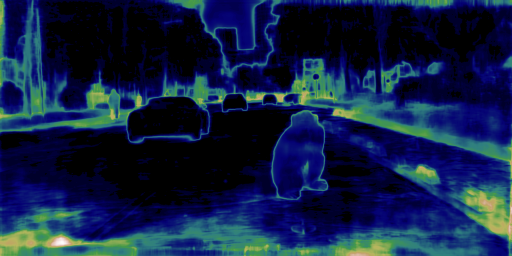}
        & \includegraphics[width=0.18\textwidth]{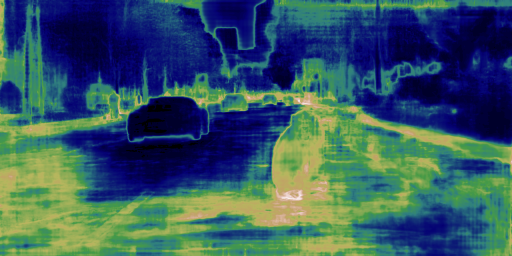}
        & \includegraphics[width=0.18\textwidth]{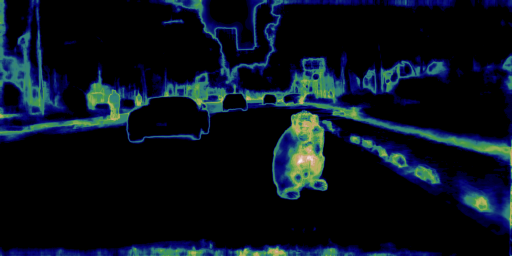}
        & \includegraphics[width=0.18\textwidth]{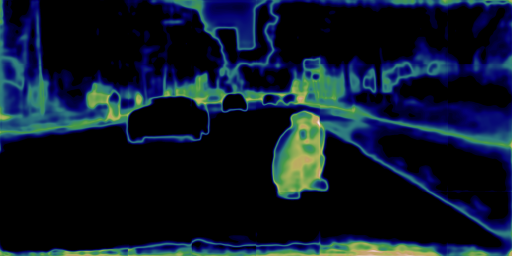} \\
        \includegraphics[width=0.18\textwidth]{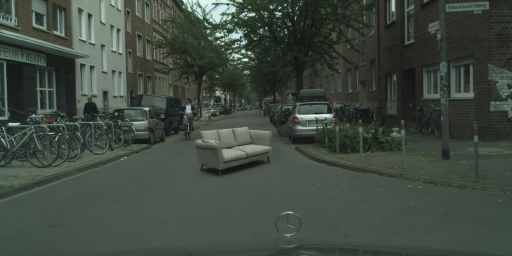}
        & \includegraphics[width=0.18\textwidth]{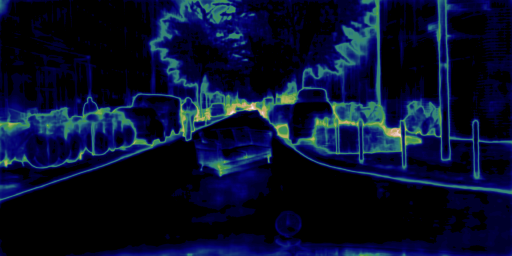}
        & \includegraphics[width=0.18\textwidth]{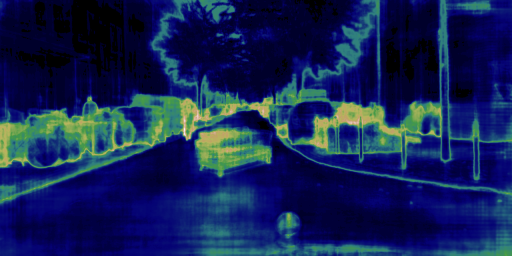}
        & \includegraphics[width=0.18\textwidth]{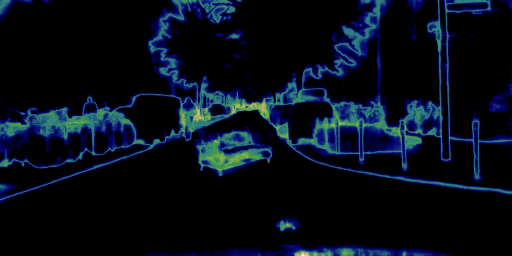}
        & \includegraphics[width=0.18\textwidth]{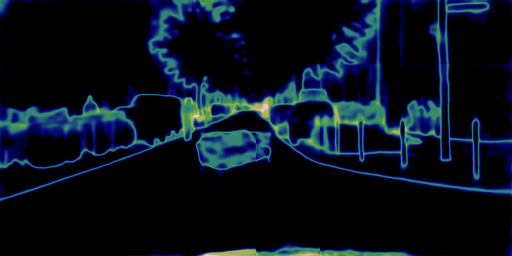} \\
        \includegraphics[width=0.18\textwidth]{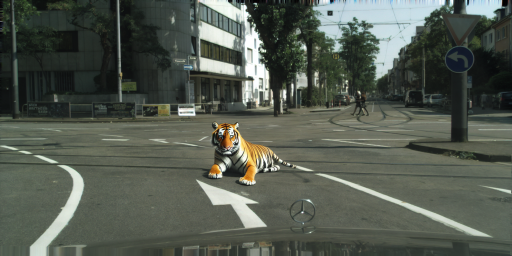}
        & \includegraphics[width=0.18\textwidth]{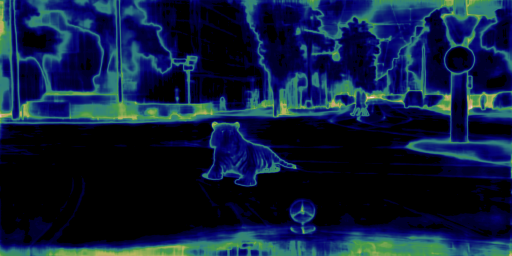}
        & \includegraphics[width=0.18\textwidth]{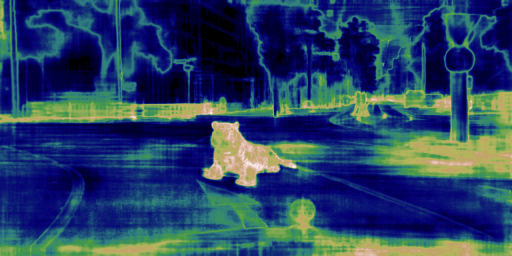}
        & \includegraphics[width=0.18\textwidth]{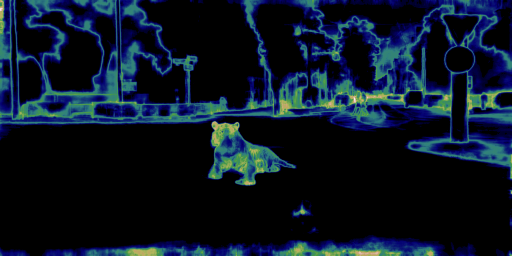}
        & \includegraphics[width=0.18\textwidth]{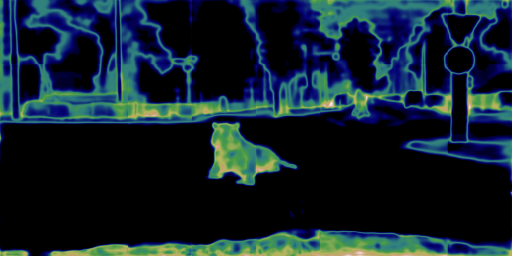} \\
        \includegraphics[width=0.18\textwidth]{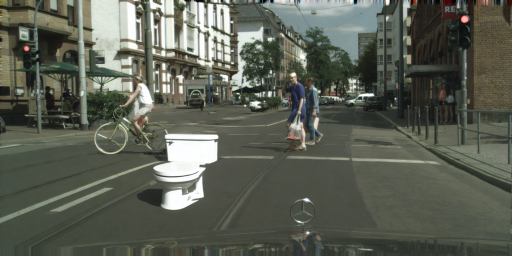}
        & \includegraphics[width=0.18\textwidth]{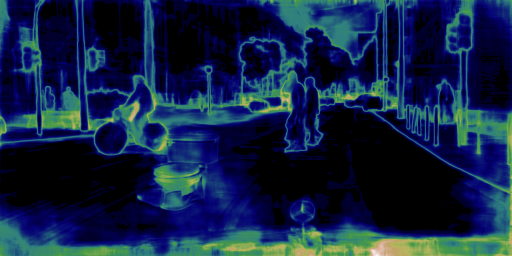}
        & \includegraphics[width=0.18\textwidth]{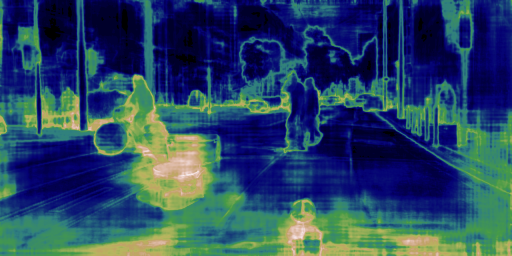}
        & \includegraphics[width=0.18\textwidth]{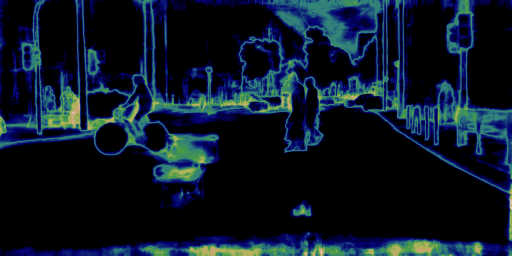}
        & \includegraphics[width=0.18\textwidth]{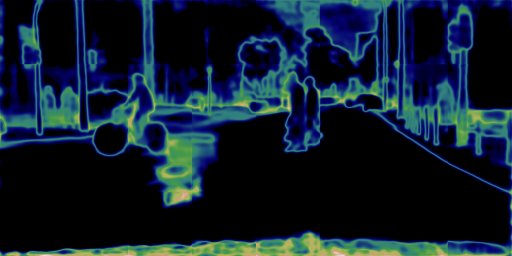} \\
        \includegraphics[width=0.18\textwidth]{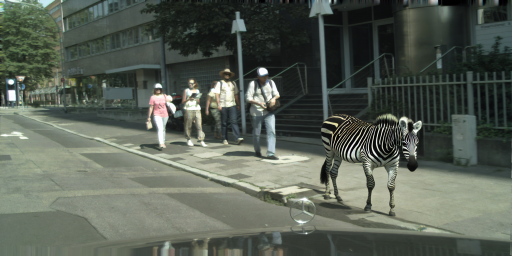}
        & \includegraphics[width=0.18\textwidth]{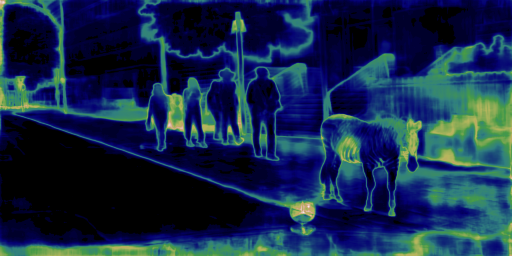}
        & \includegraphics[width=0.18\textwidth]{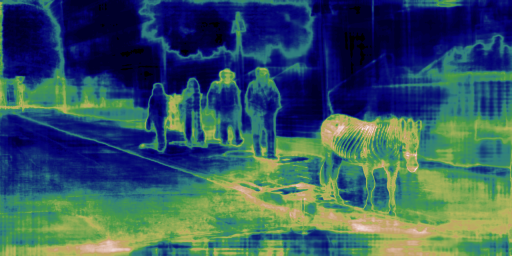}
        & \includegraphics[width=0.18\textwidth]{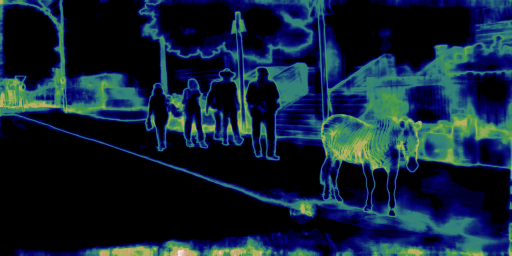}
        & \includegraphics[width=0.18\textwidth]{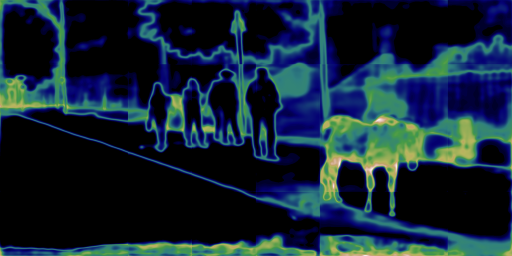} \\
        \end{tabular}
    \smallskip
    \caption{\textbf{Qualitative results.} Confidence maps are visualized for the four exemplified models on synthetic inpainted data. Hotter colors correspond to higher OOD likelihood. Ideally, results should exhibit hot colors in OOD areas and cold colors everywhere else.}
    \label{fig:supp_sec4_quali}
\end{figure*}
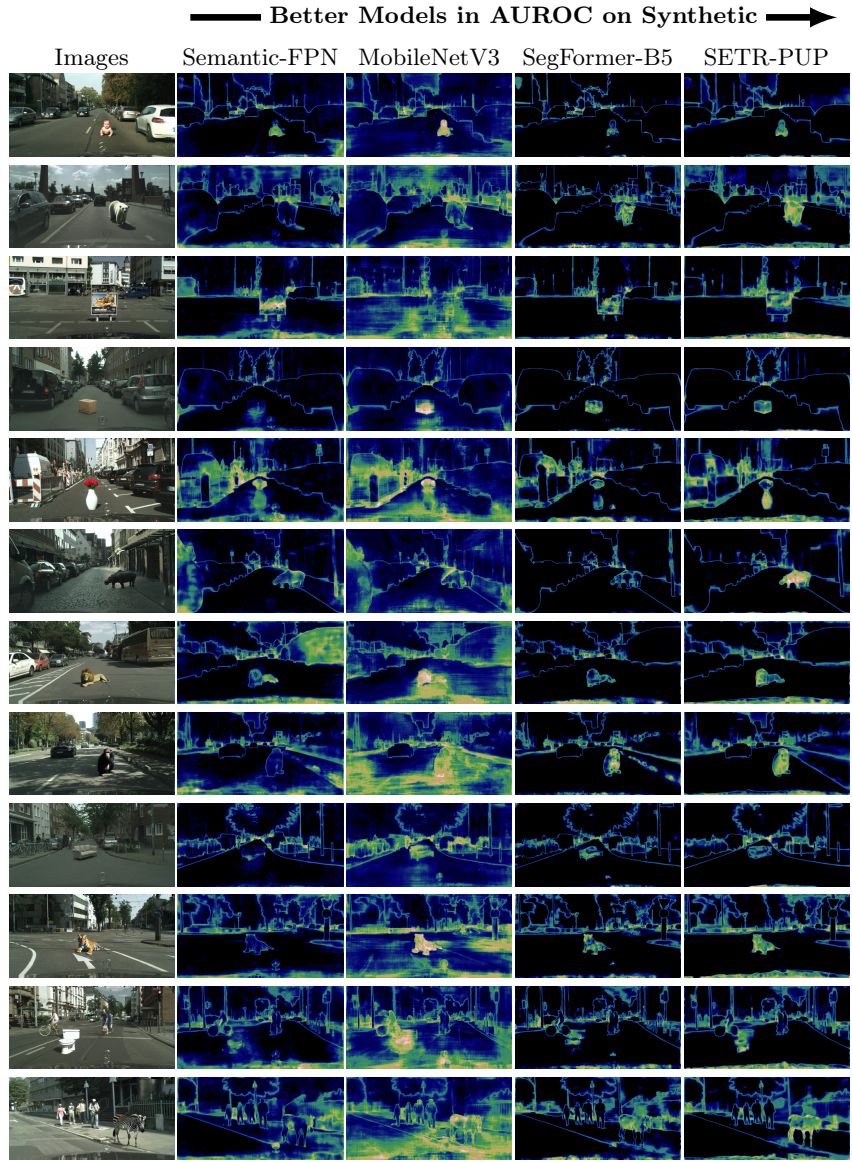

We show more qualitative examples for synthetic covariate shifts in \cref{fig:qualres_sec3_suppmat} and synthetic OOD objects in \cref{fig:supp_sec4_quali}.

	%
	%
	\bibliographystyle{splncs04}
	\bibliography{main}

\begin{thebibliography}{10}
\providecommand{\url}[1]{\texttt{#1}}
\providecommand{\urlprefix}{URL }
\providecommand{\doi}[1]{https://doi.org/#1}

\bibitem{clipinterrogator}
clip-interrogator. \url{https://github.com/pharmapsychotic/clip-interrogator}
  (2023)

\bibitem{besnier2020dataset}
Besnier, V., Jain, H., Bursuc, A., Cord, M., P{\'e}rez, P.: This dataset does
  not exist: training models from generated images. In: ICASSP (2020)

\bibitem{blum2021fishyscapes}
Blum, H., Sarlin, P.E., Nieto, J., Siegwart, R., Cadena, C.: The fishyscapes
  benchmark: Measuring blind spots in semantic segmentation (2021)

\bibitem{chan2021segmentmeifyoucan}
Chan, R., Lis, K., Uhlemeyer, S., Blum, H., Honari, S., Siegwart, R., Fua, P.,
  Salzmann, M., Rottmann, M.: Segmentmeifyoucan: A benchmark for anomaly
  segmentation. In: NeurIPS (2021)

\bibitem{geirhos2021partial}
Geirhos, R., Narayanappa, K., Mitzkus, B., Thieringer, T., Bethge, M.,
  Wichmann, F.A., Brendel, W.: Partial success in closing the gap between human
  and machine vision. NeurIPS  (2021)

\bibitem{guo2017calibration}
Guo, C., Pleiss, G., Sun, Y., Weinberger, K.Q.: On calibration of modern neural
  networks. In: ICML (2017)

\bibitem{hariat2024learning}
Hariat, M., Laurent, O., Kazmierczak, R., Zhang, S., Bursuc, A., Yao, A.,
  Franchi, G.: Learning to generate training datasets for robust semantic
  segmentation. In: WACV (2024)

\bibitem{he2022synthetic}
He, R., Sun, S., Yu, X., Xue, C., Zhang, W., Torr, P., Bai, S., Qi, X.: Is
  synthetic data from generative models ready for image recognition?  (2023)

\bibitem{hendrycks2019scaling}
Hendrycks, D., Basart, S., Mazeika, M., Zou, A., Kwon, J., Mostajabi, M.,
  Steinhardt, J., Song, D.: Scaling out-of-distribution detection for
  real-world settings. In: ICML (2022)

\bibitem{hendrycks2021many}
Hendrycks, D., Basart, S., Mu, N., Kadavath, S., Wang, F., Dorundo, E., Desai,
  R., Zhu, T., Parajuli, S., Guo, M., et~al.: The many faces of robustness: A
  critical analysis of out-of-distribution generalization. In: ICCV (2021)

\bibitem{hendrycks2019benchmarking}
Hendrycks, D., Dietterich, T.: Benchmarking neural network robustness to common
  corruptions and perturbations. In: ICLR (2019)

\bibitem{hendrycks2017baseline}
Hendrycks, D., Gimpel, K.: A baseline for detecting misclassified and
  out-of-distribution examples in neural networks. In: ICLR (2017)

\bibitem{jiang2020tsit}
Jiang, L., Zhang, C., Huang, M., Liu, C., Shi, J., Loy, C.C.: Tsit: A simple
  and versatile framework for image-to-image translation. In: ECCV (2020)

\bibitem{de2023reliability}
de~Jorge, P., Volpi, R., Torr, P.H., Rogez, G.: Reliability in semantic
  segmentation: Are we on the right track? In: CVPR (2023)

\bibitem{kirillov2023segany}
Kirillov, A., Mintun, E., Ravi, N., Mao, H., Rolland, C., Gustafson, L., Xiao,
  T., Whitehead, S., Berg, A.C., Lo, W.Y., Doll{\'a}r, P., Girshick, R.:
  Segment anything. arXiv:2304.02643  (2023)

\bibitem{koh2021wilds}
Koh, P.W., Sagawa, S., Marklund, H., Xie, S.M., Zhang, M., Balsubramani, A.,
  Hu, W., Yasunaga, M., Phillips, R.L., Gao, I., et~al.: Wilds: A benchmark of
  in-the-wild distribution shifts. In: ICLR (2021)

\bibitem{le2021semantic}
Le~Moing, G., Vu, T.H., Jain, H., P{\'e}rez, P., Cord, M.: Semantic palette:
  Guiding scene generation with class proportions. In: CVPR (2021)

\bibitem{li2022bigdatasetgan}
Li, D., Ling, H., Kim, S.W., Kreis, K., Fidler, S., Torralba, A.:
  Bigdatasetgan: Synthesizing imagenet with pixel-wise annotations. In: CVPR
  (2022)

\bibitem{li2023imagenet}
Li, X., Chen, Y., Zhu, Y., Wang, S., Zhang, R., Xue, H.: Imagenet-e:
  Benchmarking neural network robustness via attribute editing. In: CVPR (2023)

\bibitem{liu2023grounding}
Liu, S., Zeng, Z., Ren, T., Li, F., Zhang, H., Yang, J., Li, C., Yang, J., Su,
  H., Zhu, J., et~al.: Grounding dino: Marrying dino with grounded pre-training
  for open-set object detection. arXiv preprint arXiv:2303.05499  (2023)

\bibitem{LugmayrDRYTG22}
Lugmayr, A., Danelljan, M., Romero, A., Yu, F., Timofte, R., Gool, L.V.:
  Repaint: Inpainting using denoising diffusion probabilistic models. In: CVPR
  (2022)

\bibitem{marathe2023wedge}
Marathe, A., Ramanan, D., Walambe, R., Kotecha, K.: Wedge: A multi-weather
  autonomous driving dataset built from generative vision-language models. In:
  CVPRW (2023)

\bibitem{miller2021accuracy}
Miller, J.P., Taori, R., Raghunathan, A., Sagawa, S., Koh, P.W., Shankar, V.,
  Liang, P., Carmon, Y., Schmidt, L.: Accuracy on the line: on the strong
  correlation between out-of-distribution and in-distribution generalization.
  In: ICLR (2021)

\bibitem{naeini2015obtaining}
Naeini, M.P., Cooper, G., Hauskrecht, M.: Obtaining well calibrated
  probabilities using bayesian binning. In: AAAI (2015)

\bibitem{nayal2023rba}
Nayal, N., Yavuz, M., Henriques, J.F., G{\"u}ney, F.: Rba: Segmenting unknown
  regions rejected by all. In: ICCV (2023)

\bibitem{ovadia2019can}
Ovadia, Y., Fertig, E., Ren, J., Nado, Z., Sculley, D., Nowozin, S., Dillon,
  J., Lakshminarayanan, B., Snoek, J.: Can you trust your model's uncertainty?
  evaluating predictive uncertainty under dataset shift. NeurIPS  (2019)

\bibitem{pinggera2016lost}
Pinggera, P., Ramos, S., Gehrig, S., Franke, U., Rother, C., Mester, R.: Lost
  and found: detecting small road hazards for self-driving vehicles. In: IROS
  (2016)

\bibitem{podell2023sdxl}
Podell, D., English, Z., Lacey, K., Blattmann, A., Dockhorn, T., M{\"u}ller,
  J., Penna, J., Rombach, R.: Sdxl: Improving latent diffusion models for
  high-resolution image synthesis. arXiv preprint arXiv:2307.01952  (2023)

\bibitem{prabhu2023lance}
Prabhu, V., Yenamandra, S., Chattopadhyay, P., Hoffman, J.: Lance:
  Stress-testing visual models by generating language-guided counterfactual
  images. In: NeurIPS (2023)

\bibitem{recht2019imagenet}
Recht, B., Roelofs, R., Schmidt, L., Shankar, V.: Do imagenet classifiers
  generalize to imagenet? In: ICLR (2019)

\bibitem{rombach2022high}
Rombach, R., Blattmann, A., Lorenz, D., Esser, P., Ommer, B.: High-resolution
  image synthesis with latent diffusion models. In: CVPR (2022)

\bibitem{sakaridis2018semantic}
Sakaridis, C., Dai, D., Van~Gool, L.: Semantic foggy scene understanding with
  synthetic data. IJCV  (2018)

\bibitem{sakaridis2021acdc}
Sakaridis, C., Dai, D., Van~Gool, L.: Acdc: The adverse conditions dataset with
  correspondences for semantic driving scene understanding. In: ICCV (2021)

\bibitem{sariyildiz2023fake}
Sariyildiz, M.B., Alahari, K., Larlus, D., Kalantidis, Y.: Fake it till you
  make it: Learning transferable representations from synthetic imagenet
  clones. In: CVPR (2023)

\bibitem{singh2023benchmarking}
Singh, A., Sarangmath, K., Chattopadhyay, P., Hoffman, J.: Benchmarking
  low-shot robustness to natural distribution shifts. In: ICCV (2023)

\bibitem{taori2020measuring}
Taori, R., Dave, A., Shankar, V., Carlini, N., Recht, B., Schmidt, L.:
  Measuring robustness to natural distribution shifts in image classification.
  NeurIPS  (2020)

\bibitem{teney2022id}
Teney, D., Lin, Y., Oh, S.J., Abbasnejad, E.: Id and ood performance are
  sometimes inversely correlated on real-world datasets. arXiv  (2022)

\bibitem{tran2022plex}
Tran, D., Liu, J., Dusenberry, M.W., Phan, D., Collier, M., Ren, J., Han, K.,
  Wang, Z., Mariet, Z., Hu, H., et~al.: Plex: Towards reliability using
  pretrained large model extensions. arXiv  (2022)

\bibitem{varma2019idd}
Varma, G., Subramanian, A., Namboodiri, A., Chandraker, M., Jawahar, C.: Idd: A
  dataset for exploring problems of autonomous navigation in unconstrained
  environments. In: WACV (2019)

\bibitem{wu2023diffumask}
Wu, W., Zhao, Y., Shou, M.Z., Zhou, H., Shen, C.: Diffumask: Synthesizing
  images with pixel-level annotations for semantic segmentation using diffusion
  models. In: ICCV (2023)

\bibitem{yang2022openood}
Yang, J., Wang, P., Zou, D., Zhou, Z., Ding, K., Peng, W., Wang, H., Chen, G.,
  Li, B., Sun, Y., et~al.: Openood: Benchmarking generalized
  out-of-distribution detection. NeurIPS  (2022)

\bibitem{zhang2023adding}
Zhang, L., Rao, A., Agrawala, M.: Adding conditional control to text-to-image
  diffusion models. In: ICCV (2023)

\bibitem{zhang2021datasetgan}
Zhang, Y., Ling, H., Gao, J., Yin, K., Lafleche, J.F., Barriuso, A., Torralba,
  A., Fidler, S.: Datasetgan: Efficient labeled data factory with minimal human
  effort. In: CVPR (2021)

\end{thebibliography}
\end{document}